\documentclass[twoside]{article}
\usepackage[arxiv]{optional}

\usepackage[accepted]{aistats2019}
\usepackage{lmodern}

\usepackage{natbib} %

\usepackage[utf8]{inputenc} %
\usepackage[T1]{fontenc}    %
\usepackage[colorlinks,citecolor=blue,urlcolor=black,linkcolor=blue,pdfborder={0 0 0}]{hyperref}       %
\usepackage{url}            %
\usepackage{booktabs}       %
\usepackage{amsfonts}       %
\usepackage{nicefrac}       %
\usepackage{microtype}      %
\usepackage{float}

\usepackage{threeparttable}
\usepackage{subcaption}
\usepackage{graphicx}
\usepackage{amsmath, amsthm, bm, bbm, amssymb}
\usepackage{mathtools}
\usepackage{natbib}
\usepackage[capitalize]{cleveref}  
\usepackage{xr}
\usepackage{array}

\crefname{nlem}{Lemma}{Lemmas}
\crefname{nprop}{Proposition}{Propositions}
\crefname{ncor}{Corollary}{Corollaries}
\crefname{nthm}{Theorem}{Theorems}
\crefname{nnon}{Conjecture}{Conjectures}
\crefname{assumption}{Assumption}{Assumptions}

\RequirePackage[usenames,dvipsnames]{color}

\usepackage{jhh-misc2}
\usepackage{mathrsfs}

\newcommand{\abscont}{\ll}  % absolutely continuous 

\newcommand{\textder}[2]{\dee #1/\dee #2}

\newcommand{\mainmeas}{\eta}
\newcommand{\estmeas}{\tilde{\mainmeas}}
\newcommand{\mainmean}{\mu}
\newcommand{\estmean}{\tilde{\mainmean}}
\newcommand{\mainstd}{s}
\newcommand{\eststd}{\tilde{\mainstd}}

\newcommand{\covspace}{\mcX}
\newcommand{\data}{\bD}
\newcommand{\x}{\boldsymbol x}
\newcommand{\xn}[1]{\x_{#1}}
\newcommand{\xall}{\boldsymbol X}
\newcommand{\yn}[1]{y_{#1}}
\newcommand{\yall}{\boldsymbol y}
\newcommand{\indpts}{\boldsymbol\tX}
\newcommand{\indpt}[1]{\boldsymbol \tx_{#1}}
\newcommand{\loglik}[1]{\mcL_{#1}}
\newcommand{\indptloglik}[1]{\tilde\mcL_{#1}}

\newcommand{\auxpoints}{\hat\xall}
\newcommand{\auxy}{\hat\yall}
\newcommand{\auxdata}{\hat\data}

\newcommand{\GPmean}{\mu}
\newcommand{\auxmean}{\hat\GPmean}
\newcommand{\postmean}{\GPmean_{\data}}
\newcommand{\postkernel}{\GPkernel_{\data}}

\newcommand{\basedist}{\lambda}
\newcommand{\priordist}{\targetdist_{0}}

\newcommand{\Wiener}[1]{W_{#1}}
\newcommand{\dWt}[1]{\dee \Wiener{#1}}

\newcommand{\drift}{b}
\newcommand{\adrift}{\tilde\drift}

\newcommand{\Xt}[1]{X_{#1}}
\newcommand{\dXt}[1]{\dee \Xt{#1}}
\newcommand{\Yt}[1]{Y_{#1}}
\newcommand{\dYt}[1]{\dee \Yt{#1}}
\newcommand{\dt}{\dee t}

\newcommand{\probspace}{\mcP}

\newcommand{\Fderiv}{\mcD}  % Frechet derivative
\newcommand{\targetdist}{\pi}
\newcommand{\approxdist}{\tilde\targetdist}
\newcommand{\auxdist}{\nu}

\newcommand{\GPkernel}{k}
\newcommand{\sqGPkernel}{\GPkernel'}
\newcommand{\auxkernel}{\hat\GPkernel}
\newcommand{\RKHSkernel}{r}

\newcommand{\rkhs}{\mathbb H}
\newcommand{\rkhsarg}[1]{\rkhs_{#1}}

\newcommand{\normarg}[2]{\norm{#2}_{#1}}
\newcommand{\staticnormarg}[2]{\staticnorm{#2}_{#1}}

\newcommand{\kernelop}[1]{\mcC_{#1}}

\newcommand{\innerarg}[3]{\inner{#2}{#3}_{#1}} % Inner product

\newcommand{\rinner}[2]{\innerarg{\RKHSkernel}{#1}{#2}}

\newcommand{\basis}[1]{e_{#1}}
\newcommand{\keval}[1]{\lambda_{#1}}  % eigenvalues for RKHS kernel
\newcommand{\Lparg}[2]{L^{#1}(#2)}
\newcommand{\Lpnormarg}[3]{\norm{#3}_{\Lparg{#1}{#2}}}

\newcommand{\couplings}[2]{\Gamma(#1, #2)}
\newcommand{\coupling}{\gamma}

\newcommand{\pwassSimple}[3]{\mcW_{#1}(#2, #3)}

\def\pf{pF\xspace}
\newcommand{\pfd}[3]{d_{\mathrm{\pf},#1}(#2 || #3)}
\def\pfdname{\pf divergence\xspace}
\newcommand{\fd}[3]{d_{\mathrm{F},#1}(#2, #3)}
\newcommand{\pfdtc}{\pf-{DTC}\xspace}

\def\norm#1{\left\|{#1}\right\|} % A norm with 1 argument
\newcommand{\infnorm}[1]{\norm{#1}_{\infty}} % Linfty norm
\def\staticnorm#1{\|{#1}\|} % A static norm that does not resize with input
\newcommand{\statictwonorm}[1]{\staticnorm{#1}_2} % L2 norm
\newcommand{\inner}[2]{\langle{#1},{#2}\rangle} % Inner product
\def\Holder{H\"older\xspace}
\def\Ito{It\^o\xspace}
\def\Nystrom{Nystr\"om\xspace}

\def\Matern{Mat\'ern\xspace}
\def\Frechet{Fr\'echet\xspace}

\boldshortcuts

\opt{conf}{\suppresscomments}
\opt{conf}{\externaldocument{scalable-GPs-supplementary-material-AISTATS-2019}}

\graphicspath{ {figures/} }

\begin{document}

\runningtitle{Scalable Gaussian Process Inference with Finite-data Mean and Variance Guarantees}%
\runningauthor{Jonathan H.~Huggins, Trevor Campbell, Miko{\l}aj Kasprzak, Tamara Broderick}

\twocolumn[

\aistatstitle{Scalable Gaussian Process Inference with \\ Finite-data Mean and Variance Guarantees}

\aistatsauthor{Jonathan H.~Huggins \And Trevor Campbell \And ~~~~~~Miko{\l}aj Kasprzak \And Tamara Broderick}
\aistatsaddress{Harvard University \And University of British Columbia \And ~~~~~~University of Luxembourg \And MIT}
]

\begin{abstract}
Gaussian processes (GPs) offer a flexible class of priors for nonparametric Bayesian regression, but 
popular GP posterior inference methods are typically prohibitively slow or lack desirable finite-data guarantees on quality.
We develop a scalable approach to approximate GP regression, with
finite-data guarantees on the accuracy of our pointwise posterior mean and variance estimates. 
Our main contribution is a novel objective for approximate inference 
in the nonparametric setting: the \emph{preconditioned Fisher (\pf) divergence}.
We show that unlike the Kullback--Leibler divergence (used in variational inference),
the \pfdname bounds the 2-Wasserstein distance, which in turn provides tight bounds on the pointwise
error of mean and variance estimates.
We demonstrate that, for sparse GP likelihood approximations, we can minimize the \pfdname efficiently.
Our experiments show that optimizing the \pfdname has the same computational requirements 
as variational sparse GPs while providing comparable empirical performance---in addition to our novel 
finite-data quality guarantees.
\end{abstract}

\section{Introduction} \label{sec:intro}

Gaussian processes (GPs) offer a versatile class of models for functions. 
In particular, GPs are able to capture complex, highly non-linear relationships in data. 
Unfortunately, even in the setting of GP regression with Gaussian noise, 
exact GP inference---that is, computing the posterior mean and covariance functions 
of the GP given $N$ observations---incurs a prohibitive $O(N^{3})$ running time and $O(N^{2})$ memory cost~\citep{Rasmussen:2006}. By contrast, modern, large-scale data sets require algorithms with time and memory requirements at most linear in $N$. A natural question arises: can a reliably good GP \emph{approximation} be found with at most linear cost? To understand what counts as a ``good'' approximation, we observe that practitioners tend to report pointwise estimates and uncertainties for the GP regression curve, especially pointwise posterior means and standard deviations~\citep{Sacks:1989,Kaufman:2010,Kaufman:2011,Gramacy:2012,Rasmussen:2002,Rasmussen:2006,Snoek:2012,Osborne:2012}. To the best of our knowledge, it is an open problem whether non-trivial, non-asymptotic theoretical bounds exist on the quality of these estimates when running linear-time approximations to GP regression. 
We believe we provide the first approach that simultaneously satisfies the following desiderata: (1) linear running time and memory cost in $N$,
(2) theoretical bounds on the quality of pointwise estimates of the posterior mean, standard deviation, and variance for finite data, 
and (3) a practical mechanism for provably decreasing these bounds to zero.

Previous approaches to scaling GP inference primarily fall into two categories: sparse GP 
methods~\citep{Smola:2000,Seeger:2003,QuinoneroCandela:2005,Snelson:2005,Titsias:2009,Bauer:2016,Hensman:2013,deGMatthews:2016,Bui:2017a} 
and structured kernel matrix approximations~\citep{Wilson:2015,Ding:2017,Katzfuss:2017b,Gardner:2018,Izmailov:2018}. 
The sparse GP approach introduces a set of $M$ \emph{inducing points}, which are chosen either greedily or at random and can then be optimized. 
The most widely used and top-performing sparse GP methods are variational and originate with the work of \citet{Titsias:2009}.
These methods optimize the inducing point locations using a variational evidence lower bound~\citep{Titsias:2009,Bauer:2016,Hensman:2013,deGMatthews:2016,Bui:2017a}
and have $O(M^{2}N)$ time and $O(MN)$ space complexity. 
Structured kernel approximations replace the $N \times N$ kernel matrix with an approximation that can be represented 
more compactly and operated on more efficiently. The time and space requirements of these methods 
vary and usually depend on the dimensionality of the input space. 
Neither approach is guaranteed to provide accurate estimates of the posterior mean or variance. Typically, 
these methods are validated empirically on small datasets~\citep{Bauer:2016}.

In fact, after introducing GP regression in more detail in \cref{sec:background}, we note in \cref{sec:wasserstein-and-fisher-distances}
 that a posterior approximation
can be close in Kullback-Leibler (KL) divergence to the exact posterior and still exhibit bad posterior mean and variance approximations.
Thus, rather than pursue variational methods with KL divergence, we instead 
observe that closeness in \emph{2-Wasserstein distance} implies closeness of means and covariances---as well as many 
other posterior functionals (e.g.\ expectations of any function with a bounded gradient).
For GPs, it is not possible to efficiently compute the 2-Wasserstein distance to the exact posterior,
so instead we develop a theoretically-motivated proxy.

In particular, we build from a related
metric called the \emph{Fisher distance}. 
The squared Fisher distance between two distributions is defined as the expected squared norm of the 
difference between their score functions.
This distance has been successfully applied in the finite-dimensional setting both to design practical algorithms~\citep{Campbell:2017} and for theoretical analysis~\citep{Huggins:2017a,Ogden:2017}.
In \cref{sec:fisher-divergence}, we adapt the Fisher distance to the infinite-dimensional GP setting and call our new measure of fit the \emph{preconditioned Fisher (\pf) divergence}. 
We carefully design the \pfdname to work with likelihoods that are parameterized by functions---as opposed to, e.g., the finite vector parameters that are more familiar from the parametric setting. 
We show that the pointwise mean and variance approximation errors are bounded by a constant times the \pfdname, and thus minimizing 
the \pfdname results in small mean and variance approximation error.
In \cref{sec:methods} we demonstrate that the \pfdname can be computed and efficiently minimized for a sparse GP likelihood approximation. 
Our experiments in \cref{sec:experiments} on simulated and real data show that our method (1) yields comparable empirical performance with the
existing state-of-the-art variational methods while
(2) using as much or less computation.
Thus, using the \pfdname as an objective for approximate GP inference is a promising alternative to the variational approach,
offering state-of-the-art real-world performance while crucially providing finite-data guarantees on the accuracy of the posterior mean and variance estimates.

\section{Sparse Gaussian process regression} \label{sec:background}

A Gaussian process on covariate space $\covspace$ is determined by a mean function ${\GPmean : \covspace \to \reals}$ 
and a positive-definite covariance, or kernel, function ${\GPkernel : \covspace \times \covspace \to \reals}$. 
Take any two sets of locations $\xall = (\xn{1}, \dots, \xn{N}) \in \covspace^{N}$ and 
$\xall' = (\xn{1}', \dots, \xn{N'}') \in \covspace^{N'}$ and any function $f : \covspace \to \reals$. Let $f_{\xall} \defined (f(\xn{1}),\dots, f(\xn{N}))$, and let
$\GPkernel_{\xall\xall'}$ be the $N \times N'$ matrix where $[\GPkernel_{\xall\xall'}]_{nn'} = \GPkernel(\xn{n}, \xn{n'}')$. 
If $f \dist \distGP(\GPmean, \GPkernel)$ is a GP-distributed function, 
 $f$ is marginally Gaussian at any set of locations: $f_{\xall} \dist \distNorm(\GPmean_{\xall}, \GPkernel_{\xall\xall})$.

In GP regression~\citep{Rasmussen:2006}, we model noisy observed output $\yn{n}$ at input location $\xn{n}$:
\(
\yn{n} \given f, \xn{n} &\distind \distNorm(f(\xn{n}), \sigma^{2}), \quad n = 1,\dots, N \\
f &\dist \distGP(0, \GPkernel),
\)
where $\sigma^{2} > 0$ is the observation noise variance and
we make a standard assumption that the mean function is identically zero.
We collect the response variables as $\yall \defined (\yn{1}\dots, \yn{N}) \in \reals^{N}$.
The posterior distribution given the data $\data = (\xall, \yall)$ is the Gaussian process
$\targetdist = \distGP(\postmean, \postkernel)$, 
where $\postmean(\x) \defined \GPkernel_{\x \xall}[\GPkernel_{\xall\xall} + \sigma^{2}I]^{-1}\yall$ 
and $\postkernel(\x, \x') \defined \GPkernel(\x, \x') - \GPkernel_{\x\xall}[\GPkernel_{\xall\xall} + \sigma^{2}I]^{-1}\GPkernel_{\xall\x'}$ 
are, respectively, the posterior mean and kernel functions.
The posterior mean and kernel functions are usually the quantities of interest 
since they fully define the GP posterior distribution. They can then be used for prediction, in Bayesian optimization, in the design of computer experiments, 
and for other tasks~\citep{Sacks:1989,Kaufman:2010,Kaufman:2011,Gramacy:2012,Rasmussen:2002,Rasmussen:2006,Snoek:2012,Osborne:2012}.
Computing the posterior exactly requires $O(N^{3})$ time due to the cost of inversion of the $N \times N$ matrix $\GPkernel_{\xall\xall} + \sigma^{2}I$.
Sparse Gaussian process approximations make GP inference more scalable by
introducing a set of $M$ \emph{inducing points} $\indpts = (\indpt{1},\dots,\indpt{m})$
and evaluating the function $f$ only at those locations~\citep{Seeger:2003,QuinoneroCandela:2005,Snelson:2005,Titsias:2009}.
The key idea is to replace the exact log-likelihood $\loglik{}(f) \defined \log p_{\distNorm}(\yall \given f_{\xall}, \sigma^{2}I)$
with an approximate log-likelihood that makes use of the observation
that if $f \dist \distGP(0, \GPkernel)$, then $\EE[f_{\xall} \given f_{\indpts}] = \GPkernel_{\xall\indpts}\GPkernel_{\indpts\indpts}^{-1}f_{\indpts}$.
For example, the deterministic training conditional~(DTC) approximation~\citep{Seeger:2003,QuinoneroCandela:2005} 
is $\indptloglik{\text{DTC}}(f) \defined \log p_{\distNorm}(\yall \given \GPkernel_{\xall\indpts}\GPkernel_{\indpts\indpts}^{-1}f_{\indpts}, \sigma^{2}I)$.

\section{Wasserstein and Fisher distances} \label{sec:wasserstein-and-fisher-distances}

Scalable GP inference methods based on optimizing a variational lower bound for the model evidence 
have so far provided the best empirical performance when compared to alternative methods~\citep{Titsias:2009,Bauer:2016,Hensman:2013,Hensman:2015}.
While variational inference is an elegant approach, there are no finite-data guarantees on the accuracy of the approximate mean and covariance functions 
produced by variational methods.
The issue is that variational methods minimize the KL divergence 
between the sparse posterior approximation and the exact posterior~\citep{Titsias:2009,Bauer:2016,Hensman:2013}, 
but a small KL divergence does not necessarily imply small error in the mean and covariance estimates. 
For clarity, we consider the case of finite-dimensional Gaussians. 
\bnprop \label{prop:KL-divergence-problems}
$\forall \delta > 0$ there exist distributions $\mainmeas = \distNorm(\mainmean, \mainstd^{2})$ 
and $\estmeas = \distNorm(\estmean, \eststd^{2})$ on $\reals$ such that $\kl{\estmeas}{\mainmeas} = \delta$,
$(\mainmean - \estmean)^{2} = \eststd^{2}\{\exp(2\delta) - 1\}$, and $\eststd^{2} = \exp(-2\delta)\mainstd^{2}$.
\enprop
See proof in \cref{app:KL-prob-proof}.
\cref{prop:KL-divergence-problems} shows that, for example, if $\kl{\estmeas}{\mainmeas} = 5$ then
the mean estimate may be off by more than $148\eststd$. 
Since $\eststd$ provides a natural unit of uncertainty about the parameter value, 
we see that a moderate Kullback--Leibler divergence can correspond to a very large error in the mean estimate. 

An alternative to KL divergences that does imply closeness of means and covariances is the 2-Wasserstein distance.
For the remainder of the paper let $\mainmeas$, $\estmeas$, and $\nu$ denote probability measures 
on a measurable space $(\rkhs, \mcB)$, where $(\rkhs, \innerarg{\rkhs}{\cdot}{\cdot})$ is a Hilbert space. 
We assume these distributions are absolutely continuous with respect to a base measure $\basedist$. 
We will typically take $\mainmeas$ to be the GP regression posterior $\targetdist$,
$\estmeas$ to be an approximation $\approxdist$ to the posterior, and $\basedist$ to be the prior $\priordist$.
Let $\couplings{\mainmeas}{\estmeas}$ denote the collection of 
distributions $\coupling$ on $\rkhs \times \rkhs$ such that $\coupling$ has marginal distributions $\mainmeas$ and $\estmeas$:
that is, $\mainmeas = \coupling(\cdot, \rkhs)$ and $\estmeas = \coupling(\rkhs, \cdot)$. 
The $2$-Wasserstein distance between $\mainmeas$ and $\estmeas$ is given by 
\(
\pwassSimple{2}{\mainmeas}{\estmeas}
\defined \inf_{\coupling \in \couplings{\mainmeas}{\estmeas}}\bigg\{\int \staticnormarg{\rkhs}{\theta - \tilde{\theta}}^{2} \coupling(\dee \theta, \dee \tilde{\theta})\bigg\}^{1/2}.
\)
Closeness in 2-Wasserstein implies closeness of means, standard deviations, and many
other expectations:
\bnprop \label{prop:mean-var-Wasserstein-error}
Assume that $\rkhs = \reals$, that $\mainmeas$ has mean $\mainmean$ and variance $\mainstd^{2}$, 
and that $\estmeas$ has mean $\estmean$ and variance $\eststd^{2}$. 
If $\pwassSimple{2}{\mainmeas}{\estmeas} \le \veps$, then 
$|\mainmean - \estmean| \le \veps$, $|\mainstd - \eststd| \le 2 \veps$, and, 
for any function $\phi : \reals \to \reals$ such that $|\textder{\phi}{\theta}| \le L < \infty$, 
$|\EE_{\theta \dist \mainmeas}[\phi(\theta)] - \EE_{\ttheta \dist v}[\phi(\ttheta)] \le L\eps$. 
\enprop
Similar results hold for distributions in $\reals^{d}$ and for Gaussian processes 
(detailed later in \cref{thm:pfd-controls-mean-and-variance,app:proof-of-pfd-controls-mean-and-variance}). 
Unfortunately it is not feasible to directly minimize the 2-Wasserstein distance between 
the exact posterior $\targetdist$ and some approximation $\approxdist$ because it would require computing the 
posterior, including its normalizing constant. %
Therefore, we introduce an alternative divergence measure we call the \emph{$\nu$-Fisher distance}.
Crucially, the $\nu$-Fisher distance satisfies two properties:
 (1) it does not depend on the normalizing constant of the target distribution, so it can be computed in practice,
and (2) it provides an upper bound on the 2-Wasserstein distance, so it provides accuracy guarantees on means and standard deviations.
Let $\Lparg{2}{\nu}$ denote the space of functions that are square-integrable with respect to $\nu$: 
$\phi \in \Lparg{2}{\nu}  \implies  \Lpnormarg{2}{\nu}{\phi} \defined (\int \phi(\theta)^{2}\nu(\dee \theta))^{1/2} < \infty$. 
\bnumdefn
When $\rkhs = \reals^{d}$, the \textbf{$\nu$-Fisher distance} between two distributions is defined as 
the square root of the expected squared Euclidean distance between the score functions of the two densities:
\(
\fd{\nu}{\mainmeas}{\estmeas}
	\defined \Lpnormarg{2}{\nu}{\Big\|\grad\log\der{\estmeas}{\theta} - \grad\log\der{\mainmeas}{\theta}\Big\|_{2}}
\)
\enumdefn

The $\nu$-Fisher distance implies the following 2-Wasserstein bound, which for simplicity we state for the most relevant case
of when $\mainmeas$ and $\estmeas$ are (finite-dimensional) Gaussians. 
For a matrix $A$, let $\sigma_{\mathrm{min}}(A)$ denote the smallest absolute eigenvalue of $A$
and for a function $\phi$, let $\infnorm{\phi}$ denote the usual sup-norm. 
\bnprop \label{prop:finite-dim-Wasserstein-error}
Assume $\rkhs = \reals^{d}$, $\mainmeas = \distNorm(\mainmean, \Sigma)$, $\estmeas = \distNorm(\estmean, \tilde{\Sigma})$. 
Then for any probability measure $\nu$ such that $\mainmeas \abscont \nu$, 
$
\pwassSimple{2}{\mainmeas}{\estmeas} 
\le  \sigma_{\mathrm{min}}(\tilde{\Sigma})^{-1}\infnorm{\dee \mainmeas / \dee \nu}^{1/2}\fd{\nu}{\mainmeas}{\estmeas}. %
$
\enprop
The special case when $\nu = \mainmeas$ or $\estmeas$ is known as (e.g.) the \emph{Fisher divergence} and has appeared in many applications. 
It serves as the objective in score matching, a model estimation technique~\citep{Sriperumbudur:2017,Hyvarinen:2005}.
\citet{Johnson:2004} and \citet{Ley:2013} bounded certain integral probability measures 
(e.g.\ the total variation and Kolmogorov distances) in terms of the Fisher divergence while
\citet{Huggins:2017a} bounded the 1-Wasserstein distance in terms of the Fisher divergence.\footnote{Regularity conditions were required in both cases. 
	}
In the Bayesian setting, \citet{Campbell:2017,Campbell:2018} minimized the $\nu$-Fisher distance to construct 
a high-quality posterior approximation based on a coreset. 
\citet{Campbell:2017,Campbell:2018} found their coreset method performed very well empirically, including accurate posterior mean and variance estimates.
\citet{Ogden:2017} and \citet{Dalalyan:2017b} developed related guarantees---the former using a score function approach similar in spirit to the $\nu$-Fisher distance.
Finally, \citet{Huggins:2017b} used the results of \citet{Huggins:2017a} to prove finite-data Wasserstein guarantees for a
scalable approximate inference algorithm for generalized linear models.

\section{Infinite-dimensional spaces} \label{sec:fisher-divergence}

Together, \cref{prop:mean-var-Wasserstein-error,prop:finite-dim-Wasserstein-error} 
show that small $\nu$-Fisher distance between two finite-dimensional Gaussians implies small differences in their means and standard deviations. 
Based on these theoretical foundations and the empirical successes of \citet{Campbell:2017,Campbell:2018},
the $\nu$-Fisher distance appears to be an attractive objective for optimizing the sparse GP likelihood approximation. 
However, there are a number of subtleties that must be addressed before using this distance, or a similar distance, 
for GP regression---in particular, due to the infinite-dimensional parameter space in this case. 
In the GP setting the Hilbert space $\rkhs$ is a function space, so
instead of gradients of the log-likelihood, we must use a functional derivative $\Fderiv$,
taken with respect to the Hilbert norm $\normarg{\rkhs}{\cdot}$.
That is, for a function $F : \rkhs \to \reals$, $\Fderiv F : \rkhs \to \rkhs$ is the linear operator that 
satisfies $\lim_{h \to 0}\normarg{\rkhs}{F(f + h) - F(f) - (\Fderiv F)(h)}/\normarg{\rkhs}{h} = 0$ for all $f \in \rkhs$. 
The major obstacle to overcome is that the na\"{i}ve extension of the $\nu$-Fisher distance 
can differ from the 2-Wasserstein distance by an arbitrarily large multiplicative factor.
If we tried to directly translate the $\nu$-Fisher distance to the GP setting we would get
\(
\fd{\nu}{\mainmeas}{\estmeas} 
\defined \Lpnormarg{2}{\nu}{\Big\|\Fderiv\log\der{\estmeas}{\basedist} - \Fderiv\log\der{\mainmeas}{\basedist}\Big\|_{\rkhs}}.%
\)
Notice that the derivatives are now functional derivatives and that the Radon-Nikodym derivatives are no longer with respect to 
Lebesgue measure, but rather with respect to the base measure $\basedist$. 

To gain intuition for what can go wrong in the infinite-dimensional setting, 
recall from \cref{prop:finite-dim-Wasserstein-error} that in the finite-dimensional Gaussian case,
the constant in the bound depends on $\sigma_{\mathrm{min}}(\tilde{\Sigma})^{-1}$, the reciprocal of smallest eigenvalue of the 
covariance of $\estmeas$. 
In infinite dimensions, we replace $\tilde{\Sigma}$ with the \emph{covariance operator} $\kernelop{\estmeas}$ associated with 
$\estmeas$~\citep[Ch.~1.4]{Ibragimov:1978}, the formal definition of which we defer. 
The eigenvalues of $\kernelop{\estmeas}$ get arbitrarily close to zero, 
so $\sigma_{\mathrm{min}}(\kernelop{\estmeas})^{-1} = \infty$ and the bound is vacuous.  
The next example illustrates the issue for GP regression.

\bexa \label{exa:fisher-distance-broken}
Consider a 1-dimensional Gaussian process on $\covspace = \reals$ with squared exponential kernel $\GPkernel(x, x') = e^{-(x-x')^{2}/2}$.
Let $\mainmeas$ (respectively $\estmeas$) denote the GP regression posterior distribution with $N=1$, $x_{1} = 0$, and $\yn{1} = t$ (resp.~$\yn{1} = \tilde{t}\,$). 
Then $\fd{\nu}{\mainmeas}{\estmeas}  = c(1 + \sigma^{-2})\pwassSimple{2}{\mainmeas}{\estmeas}$, where $c$ is a constant 
that depends on the choice of $\rkhs$ and the factor of $(1 + \sigma^{-2})$ can be made  arbitrarily large by taking $\sigma^{2} \to 0$. 
See \cref{app:fisher-distance-broken} for a detailed derivation. 
\eexa

\cref{exa:fisher-distance-broken} shows that the Fisher divergence bound on the Wasserstein distance could be arbitrarily large even when the Wasserstein distance itself is finite.
To understand how to fix the problem, we can take inspiration from the finite-dimensional setting; suppose $\tilde{\Sigma}$ had a very small but non-zero eigenvalue. We could perform a change of basis: 
$\theta \mapsto \tilde{\Sigma}^{1/2}\theta$.
In the new basis, $\estmeas$ would have covariance equal to the identity matrix $I$, and hence the minimum eigenvalue would be one. 
Working with the score function, this change of variable corresponds to replacing
$\grad \log \der{\estmeas}{\theta}(\theta) = -\tilde{\Sigma}^{-1}(\theta - \estmean)$ with 
$\tilde{\Sigma}\grad \log \pi(\theta) = -(\theta - \estmean)$, where the final expression is equal to the gradient of the log density of a Gaussian with covariance $I$. 
Formally, this procedure carries over to the infinite-dimensional case:
we replace the functional derivative $\Fderiv$  with $\kernelop{\estmeas}\Fderiv$.

To precisely define the covariance operator, we must describe the Hilbert space $\rkhs$ in more detail. 
Specifically, we will take $\rkhs$ to be a \emph{reproducing kernel Hilbert space} (RKHS)~\citep[Ch.~4]{Steinwart:2008}.
For a positive definite \emph{reproducing kernel} $\RKHSkernel : \covspace \times \covspace \to \reals$,
the corresponding RKHS $\rkhsarg{\RKHSkernel}$ comes equipped with an inner product $\rinner{\cdot}{\cdot}$.
The function $\RKHSkernel_{\x} \defined \RKHSkernel(\x, \cdot)$ is the evaluation function at $\x$ 
(that is, the inner product and reproducing kernel satisfy the reproducing property):
for any $f \in \rkhsarg{\RKHSkernel}$ and $\x \in \covspace$, $\rinner{\RKHSkernel_{\x}}{f} = f(\x)$.
We take $\RKHSkernel$ to be fixed but for clarity make the $\RKHSkernel$-dependence explicit. 
Now we can define the covariance operator $\kernelop{\estmeas}$ associated with a distribution $\estmeas$ as the self-adjoint
operator that satisfies the identity  $\rinner{\RKHSkernel_{\x}}{\kernelop{\estmeas}\RKHSkernel_{\x'}} = \cov(f(\x), f(\x'))$,
where $f \dist \estmeas$.

\bnrmk
We emphasize the very different roles played by the kernels $\GPkernel$ and $\RKHSkernel$. 
Whereas $\GPkernel$ determines the prior covariance of the Gaussian process used for regression,
$\RKHSkernel$ induces a reproducing kernel Hilbert space $\rkhsarg{\RKHSkernel}$. 
Recall that $\mainmeas$ and $\estmeas$ should be thought of as exact and approximating GP regression
posterior distributions, which depend on the choice of $\GPkernel$. 
On the other hand, for a sample $f \dist \mainmeas$ or $\estmeas$, by construction $f \in \rkhsarg{\RKHSkernel}$.
But, as we discuss further below, $f \notin \rkhsarg{\GPkernel}$ almost surely. %
\enrmk

We can give our definition of the modified $\nu$-Fisher distance,
which we call the \emph{$\nu$-preconditioned Fisher ($\nu$-\pf) divergence}.\footnote{We use ``divergence'' instead of ``distance'' because 
the presence of the covariance operator $\kernelop{\estmeas}$ means that the \pfdname is not symmetric in its arguments.}
\bnumdefn
The \textbf{$\nu$-preconditioned Fisher divergence} is defined by 
\(
\pfd{\nu}{\estmeas}{\mainmeas}
 \defined \Lpnormarg{2}{\nu}{\Big\|\kernelop{\estmeas}\Fderiv\log\der{\estmeas}{\basedist} - \kernelop{\estmeas}\Fderiv\log\der{\mainmeas}{\basedist}\Big\|_{\RKHSkernel}} .
\)
\enumdefn
In the setting of \cref{exa:fisher-distance-broken}, we can show that 
$\pfd{\nu}{\estmeas}{\mainmeas} = \pwassSimple{2}{\mainmeas}{\estmeas}$~(see \cref{app:fisher-distance-broken}). 
For a distribution $\eta$, let $\mu_{\eta}(\x) \defined \EE_{f \dist \eta}[f(\x)]$ and
$\GPkernel_{\eta}(\x, \x') \defined \EE_{f \dist \eta}[(f(\x) - \mu_{\eta}(\x))(f(\x') - \mu_{\eta}(\x'))]$
denote the mean and covariance functions associated with $\eta$. 
More generally, we have the following powerful result:
\bnthm \label{thm:pfd-controls-mean-and-variance}
Let  $\mainmeas$, $\estmeas$, and $\nu$ denote probability measures 
on the measurable space $(\rkhsarg{\RKHSkernel}, \mcB)$, all absolutely continuous with respect to a base measure $\basedist$. 
Assume $\mainmeas$, $\estmeas$, and $\basedist$ are Gaussian measures.
Let $\veps \defined \infnorm{\dee\mainmeas/\dee\nu}^{1/2}\pfd{\nu}{\estmeas}{\mainmeas}$ and $\underline{\GPkernel}(\x, \x) \defined \GPkernel_{\estmeas}(\x, \x)\wedge\GPkernel_{\mainmeas}(\x, \x)$.
If $\veps < \infty$, 
then $\pwassSimple{2}{\mainmeas}{\estmeas} \le \veps$
and $\forall \x \in \covspace$,
\(
|\mu_{\estmeas}(\x) - \mu_{\mainmeas}(\x)| 
	&\le \RKHSkernel(\x, \x)^{1/2}\veps  \\ %
|\GPkernel_{\estmeas}(\x, \x)^{1/2} - \GPkernel_{\mainmeas}(\x, \x)^{1/2}|  
	&\le \sqrt{6}\RKHSkernel(\x, \x)^{1/2}\veps  \\
|\GPkernel_{\estmeas}(\x, \x) - \GPkernel_{\mainmeas}(\x, \x)|
	&\le 3\,\RKHSkernel(\x, \x)^{1/2}\underline{\GPkernel}(\x, \x)^{1/2}\veps  \\
	&\phantom{\le\ } + 6\,\RKHSkernel(\x, \x)\veps^{2},
\)
and, for any $\phi : \rkhs \to \reals$ such that $|\Fderiv \phi| \le L < \infty$,
\(
|\EE_{\tf \dist \estmeas}[\phi(\tf)] - \EE_{f \dist \mainmeas}[\phi(f)]| \le L\eps.
\)
\enthm

The proof of \cref{thm:pfd-controls-mean-and-variance} (found in \cref{app:proof-of-pfd-controls-mean-and-variance}) does not require $\mainmeas$ and $\estmeas$ to be Gaussian
measures, but for simplicity we have stated it for that case. 
As long as $\auxdist$ is an over-approximation of $\mainmeas$ then we can use the $\auxdist$-\pfdname to control
the 2-Wasserstein distance. 
By ``over-approximation'' we mean that $\auxdist$ has heavier tails than $\mainmeas$ so 
$\infnorm{\textder{\mainmeas}{\auxdist}}$ is bounded.
A bound on the 2-Wasserstein distance between $\mainmeas$ and $\estmeas$
implies control on the difference in their mean and covariance functions.
The factors of $\RKHSkernel(\x, \x)^{1/2}$ account for the structure of the functions sampled from the GP. 

In our application of \cref{thm:pfd-controls-mean-and-variance} we will take $\mainmeas = \targetdist$, the exact posterior;
$\estmeas = \approxdist$, the approximate posterior; and $\basedist = \priordist$, the prior.
Therefore $\log\der{\targetdist}{\priordist}(f) = \loglik{}(f)$, the exact log-likelihood,
and $\log\der{\approxdist}{\priordist}(f) = \indptloglik{}(f)$, the approximate log-likelihood.
In order to make use of the \pfdname algorithmically, we must choose three free parameters: (1) the reproducing kernel $\RKHSkernel$,
(2) the auxiliary distribution $\auxdist$, and (3) a family of log-likelihood approximations $\indptloglik{}$.
We address these choices in \cref{sec:RKHS-kernel-choice}, \cref{sec:auxdist-choice}, 
and \cref{sec:methods}, respectively.

\bnrmk
A direct computation of the constant $\veps$ in \cref{thm:pfd-controls-mean-and-variance} is difficult in practice. 
However, it could be approximated by taking $\nu = \mainmeas$ and using an importance sampling approach to estimate
$\pfd{\mainmeas}{\estmeas}{\mainmeas}$. 
\enrmk

\subsection{Choosing the reproducing kernel}
\label{sec:RKHS-kernel-choice}

The seemingly natural choice for the reproducing kernel $\RKHSkernel$ is $\GPkernel$.
However, $\rkhsarg{\GPkernel}$ is not suitable because if $\rkhsarg{\GPkernel}$  
is infinite-dimensional (as is almost always the case in practice), then for $f \dist \distGP(0, \GPkernel)$, 
$\Pr\{f \in \rkhsarg{\GPkernel} \} = 0$~(see \citet{Lukic:2001} and \citet[Ch.~6.1]{Rasmussen:2006}). 
Therefore we must choose some $\RKHSkernel \ne \GPkernel$ for which $\Pr\{f \in \rkhsarg{\RKHSkernel} \} = 1$.
The following result (proved in \cref{sec:proof-of-use-k-as-r}) greatly simplifies the issue:
\bnprop \label{prop:use-k-as-r}
Suppose $\rkhsarg{\GPkernel}$ has an orthonormal basis $(\basis{j})_{j\geq 1}$ such that
for any $\x \in \mcX$, $\left|\basis{j}(\x)\right| = o(j^{-1})$ as $j\to\infty$.
Then for any finite set of points $\bS$ and $\epsilon >0$, there exists a kernel $\RKHSkernel$
that satisfies $\Pr\{f \in \rkhsarg{\RKHSkernel} \} = 1$ and $\max_{\x,\x' \in \bS}|\GPkernel(\x, \x') - \RKHSkernel(\x, \x')| <\epsilon$.
\enprop
The decay condition $\left|\basis{j}(\x)\right| = o(j^{-1})$ is satisfied for Gaussian kernels~\citep{Steinwart:2008}
and we suspect for \Matern kernels as well. 
Since we can take $\epsilon$ arbitrarily small (e.g.~smaller than floating point error), as a practical matter we can choose $\RKHSkernel = \GPkernel$.
Furthermore, since in the GP regression setting we only need to evaluate the posterior approximation at a finite number of locations (say, the training 
points $\xall$ and some additional test locations $\xall^{*}$),
the restriction that $\bS$ be a finite set is not problematic. 
Hence, we use $\RKHSkernel = \GPkernel$ in our experiments.
In the kernel regression literature, our current approach is most closely related to fixed-design error bounds~\citep{Cortes:2010,Alaoui:2015};
but choosing  $\RKHSkernel \ne \GPkernel$ might allow more powerful generalization guarantees~\citep{Caponnetto:2007,Rudi:2015}.

\begin{figure*}[tb]
\begin{center}
\begin{subfigure}[b]{.32\textwidth} 
\includegraphics[height=80pt]{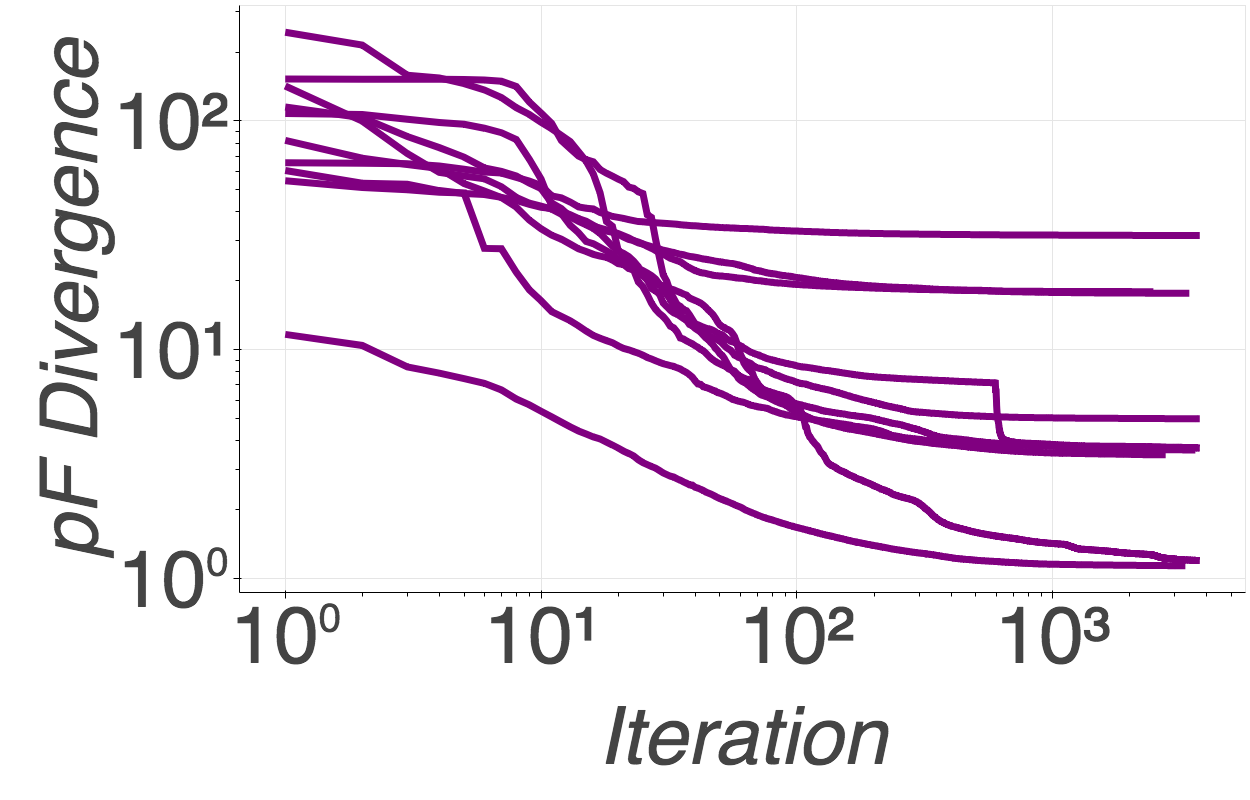} 
\caption{}
\label{fig:pfdtc-objective}
\end{subfigure}
\begin{subfigure}[b]{.32\textwidth} 
\includegraphics[height=80pt]{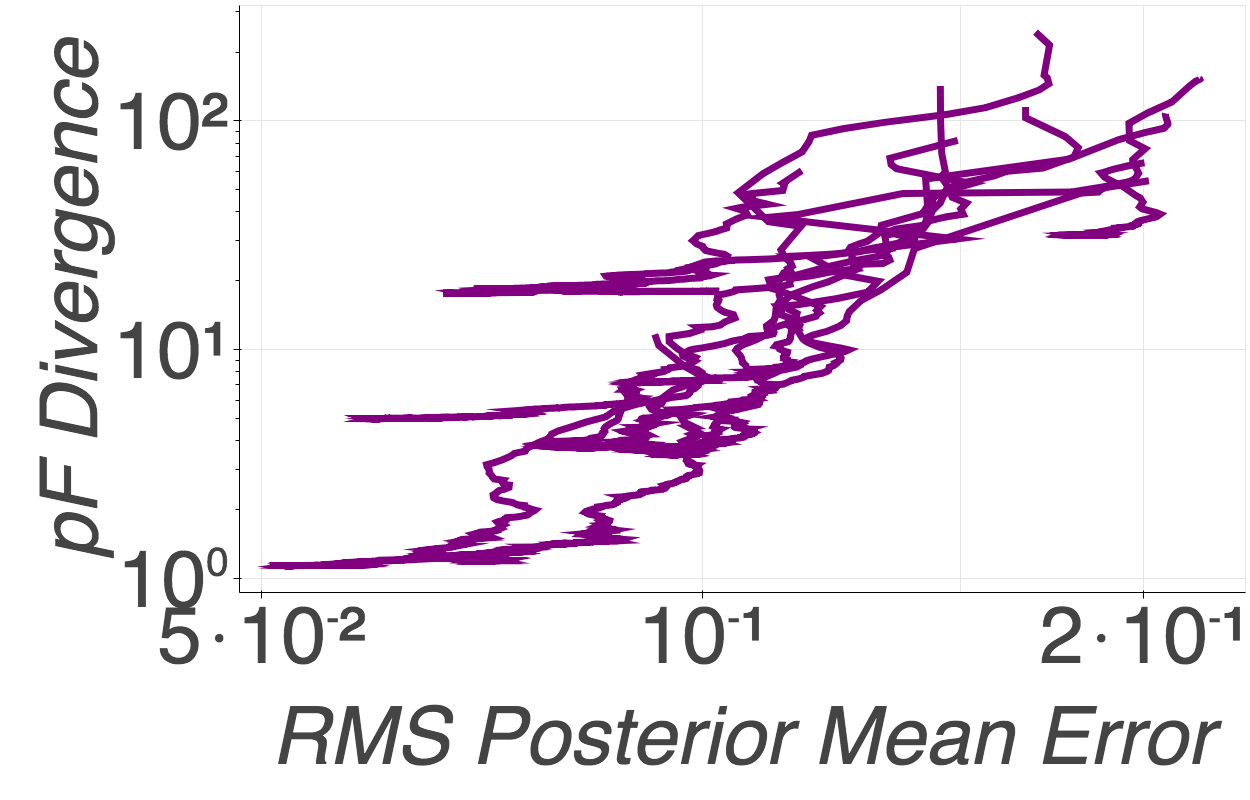} 
\caption{}
\label{fig:meantrace}
\end{subfigure} 
\begin{subfigure}[b]{.32\textwidth} 
\includegraphics[height=80pt]{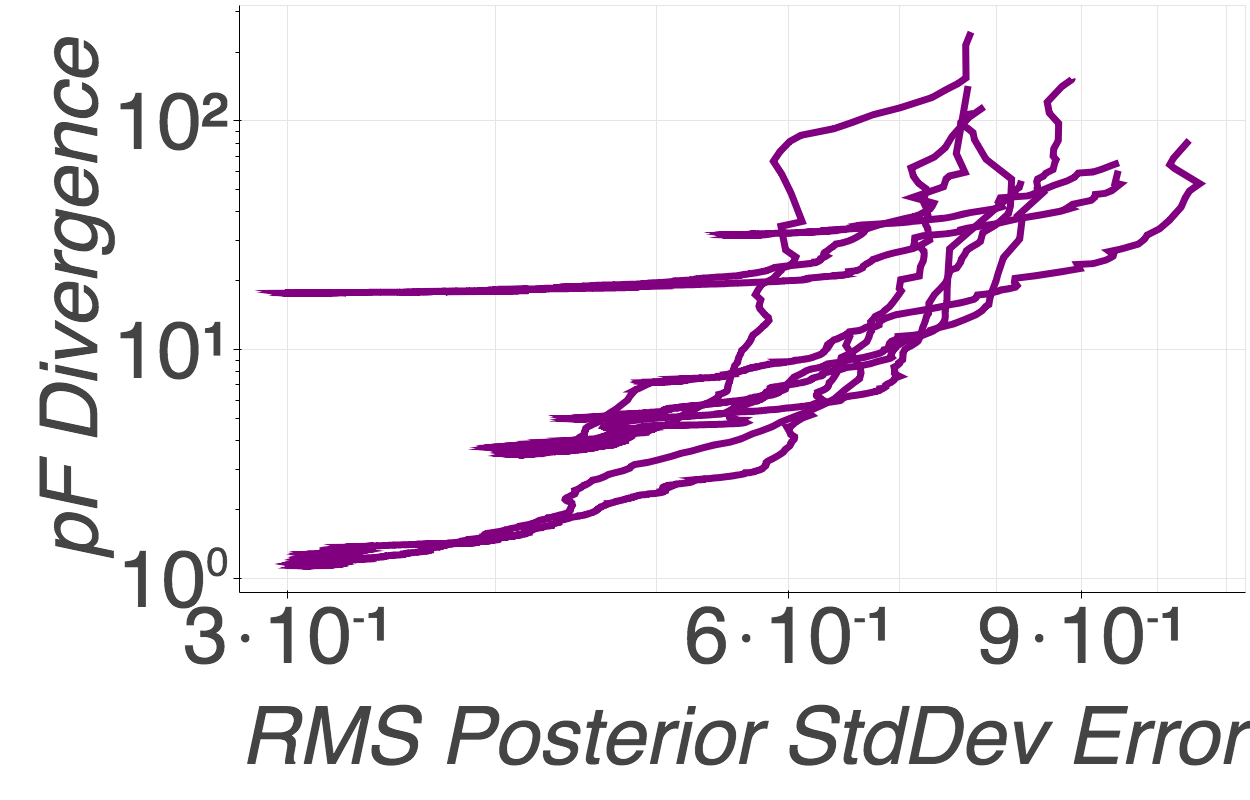} 
\caption{} 
\label{fig:stdtrace}
\end{subfigure}  
\end{center}
\caption{\textbf{(a)} Value of the \pfdtc objective on 10 runs for the airfoil data. 
\textbf{(b,c)} Relationship between the \pfdtc objective and the mean/standard deviation error on
10 runs for the airfoil dataset. 
}
\label{fig:pf-DTC}
\end{figure*}

\begin{figure*}[tb]
\begin{center}
\begin{subfigure}[b]{.32\textwidth} 
\includegraphics[trim={20 41 0 0},clip,height=72pt]{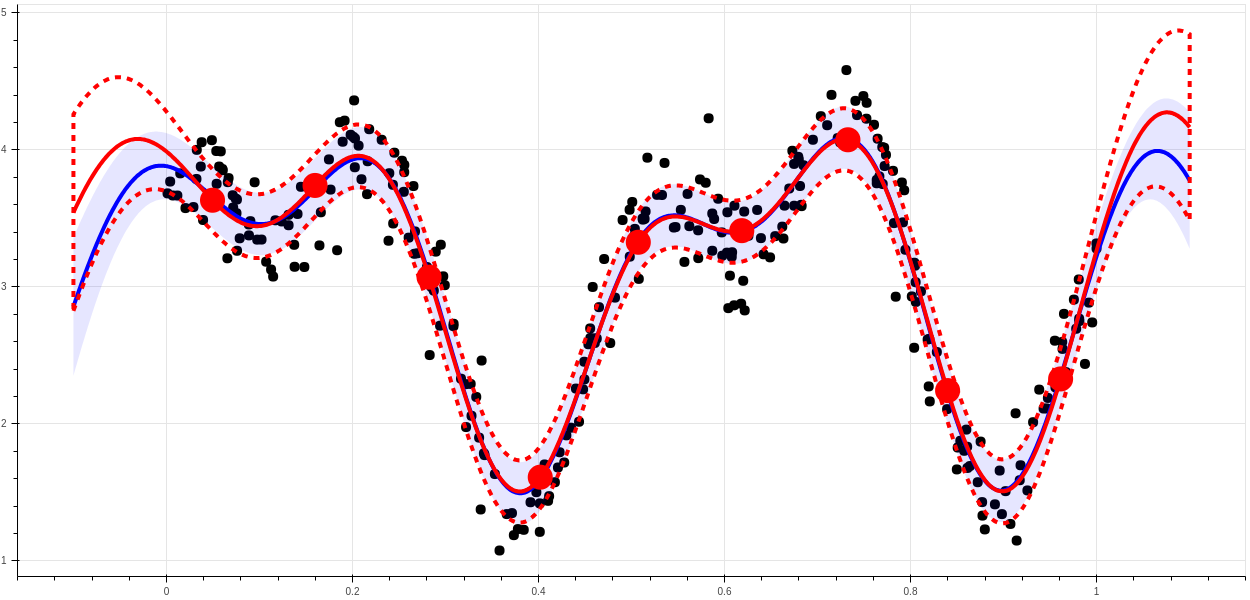} 
\caption{\pfdtc}
\label{fig:pfisher}
\end{subfigure} 
\begin{subfigure}[b]{.32\textwidth} 
\includegraphics[trim={20 41 0 0},clip,height=72pt]{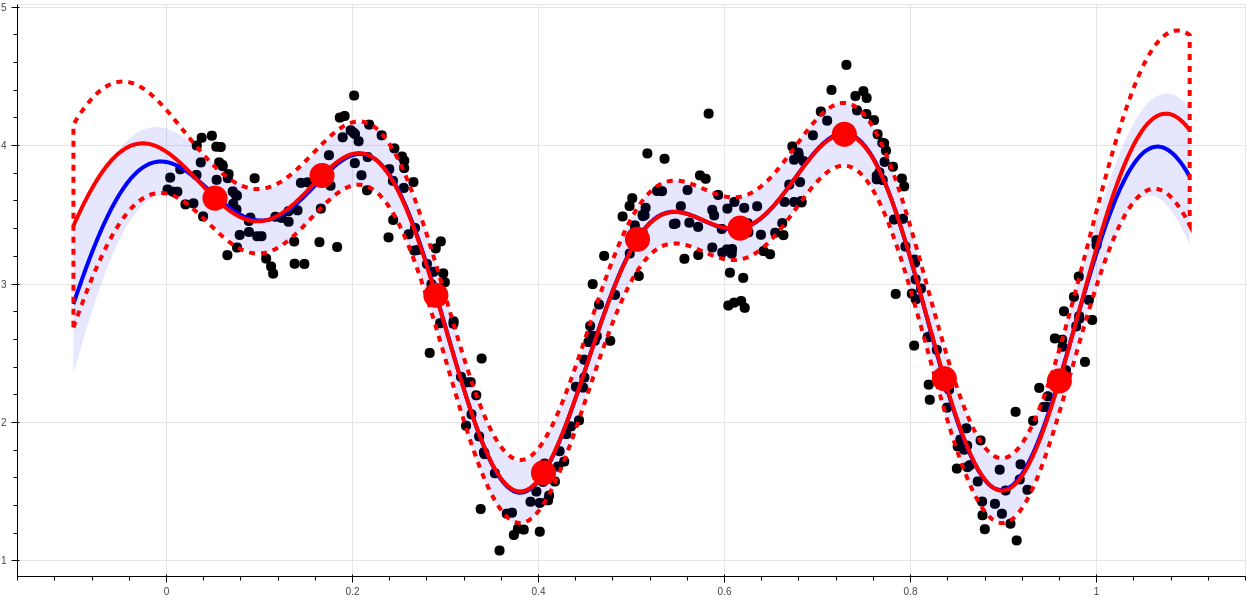} 
\caption{VFE}
\label{fig:variational}
\end{subfigure}
\begin{subfigure}[b]{.32\textwidth} 
\includegraphics[trim={20 41 0 0},clip,height=72pt]{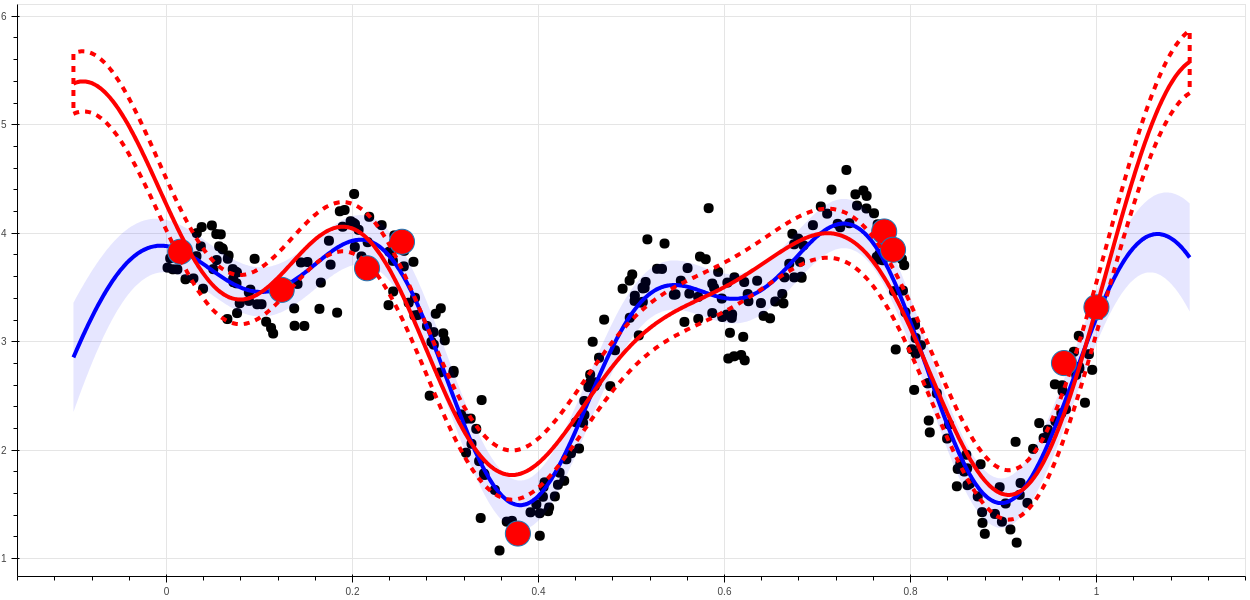} 
\caption{SoR}
\label{fig:sor}
\end{subfigure} 
\end{center}
\caption{Comparison of \pfdtc, VFE, and SoR on the synthetic example with $M=9$ inducing points. 
Data is black, exact posterior mean and standard deviation are blue,
and approximate fits and inducing points are red. 
}
\label{fig:synthetic-fit}
\end{figure*}

\subsection{Choosing the auxiliary distribution}
\label{sec:auxiliary-distribution-choice}

To apply \cref{thm:pfd-controls-mean-and-variance}, the auxiliary distribution $\auxdist$ must satisfy 
$\infnorm{\dee\targetdist/\dee\auxdist} < \infty$.
This finiteness condition is easy to achieve by letting $\auxdist = \distGP(\auxmean, \auxkernel)$ and constructing $\auxmean$ and $\auxkernel$ 
via a simple subset of datapoints approximation~\citep[Ch.~8.3.3]{Rasmussen:2006}. 
Let $\auxpoints \subset \xall$ be a random subset of the input locations of size $M'$, and take $\auxdist$ to be the GP posterior 
conditional on observing $\auxy$, the subset of $\yall$ corresponding to $\auxpoints$.
Therefore the auxiliary mean and covariance functions are $\auxmean(\x) = \GPkernel_{\x \auxpoints}[\GPkernel_{\auxpoints\auxpoints} + \sigma^{2}I]^{-1}\auxy$ 
and $\auxkernel(\x, \x') = \GPkernel(\x, \x') - \GPkernel_{\x\auxpoints}[\GPkernel_{\auxpoints\auxpoints} + \sigma^{2}I]^{-1}\GPkernel_{\auxpoints \x'}$.
Let $\auxdata = (\auxpoints, \auxy)$ denote the auxiliary data, and
let $\mcZ_{\data}$ and $\loglik{\data}$ denote the marginal likelihood and log-likelihood of the data $\data$
(with $\mcZ_{\auxdata}$ and $\loglik{\auxdata}$ defined analogously). 
We can conclude that
\(
\der{\targetdist}{\auxdist}(f) 
&= \frac{\mcZ_{\auxdata}e^{\loglik{\data}(f) - \loglik{\auxdata}(f)}}{\mcZ_{\data}} \\
&= \frac{\mcZ_{\auxdata}e^{-\sum_{\xn{n} \in \xall \setminus \auxpoints}(f(\xn{n}) - \yn{n})^{2}/(2\sigma^{2})}}{(2\pi\sigma^{2})^{\frac{N-M'}{2}}\mcZ_{\data}} \\
&\le \frac{\mcZ_{\auxdata}}{(2\pi\sigma^{2})^{\frac{N-M'}{2}}\mcZ_{\data}} < \infty. 
\)
Hence \cref{thm:pfd-controls-mean-and-variance} applies when the auxiliary distribution is a subset of datapoints approximation.

\label{sec:auxdist-choice}

\section{Preconditioned Fisher DTC} \label{sec:methods}

In this section we describe how to optimize the DTC inducing point approximation described in \cref{sec:background}
using the \pfdname.
We call the resulting inference algorithm \emph{preconditioned Fisher DTC} (\pfdtc). 
Recall the DTC log-likelihood for the $n$th observation: 
\(
\indptloglik{n}(f) \defined -\frac{1}{2\sigma^{2}}(\GPkernel_{\xn{n}\indpts}\GPkernel_{\indpts\indpts}^{-1}f_{\indpts} - \yn{n})^{2} - \frac{1}{2}\log(2\pi \sigma^{2}).
\)
We consider the case where $\RKHSkernel = \GPkernel$ at the locations of interest, which will simplify formulas~(cf.~\cref{sec:RKHS-kernel-choice}). 
Let $Q_{\xall\xall} \defined \GPkernel_{\xall\indpts}\GPkernel_{\indpts\indpts}^{-1}\GPkernel_{\indpts \xall}$,
$\bar Q_{\xall\indpts} \defined \GPkernel_{\xall\indpts}\GPkernel_{\indpts\indpts}^{-1}$, and
$S_{\xall\xall} \defined Q_{\xall\xall}(I - (Q_{\xall\xall} + \sigma^{2}I)^{-1}Q_{\xall\xall})^{2}$. 
Our main result in this section provides a formula for computing the \pfdname in the sparse GP setting
and guarantees that, up to a constant independent of the inducing points,
$\pfd{\auxdist}{\approxdist}{\targetdist}$ can be computed efficiently.
\bnprop \label{prop:pf-for-DTC}
For the DTC log-likelihood approximation, if $\auxdist = \distGP(\auxmean, \auxkernel)$, then when $\RKHSkernel = \GPkernel$,
\(
\lefteqn{\pfd{\auxdist}{\approxdist}{\targetdist}} \\
&= \underbrace{\tr((\auxkernel_{\xall\xall} + (\auxmean_{\xall} - \yall)(\auxmean_{\xall} - \yall)^{\top})(\GPkernel_{\xall\xall} - Q_{\xall\xall}))}_\mathrm{(I)} \\
&\phantom{=} + \underbrace{\tr(\auxkernel_{\xall\xall}S_{\xall\xall}) + \tr(\auxkernel_{\indpts\indpts}\bar Q_{\xall\indpts}^{\top}S_{\xall\xall}\bar Q_{\xall\indpts})}_\mathrm{(IIa)} \\
&\phantom{=} - \underbrace{2\tr(\auxkernel_{\indpts\xall}S_{\xall\xall}\bar Q_{\xall\indpts})}_\mathrm{(IIb)} \\
&\phantom{=} + \underbrace{(\auxmean_{\xall} - \bar Q_{\xall\indpts} \auxmean_{\indpts} )^{\top} S_{\xall\xall}(\auxmean_{\xall} - \bar Q_{\xall\indpts} \auxmean_{\indpts})}_\mathrm{(III)}.
\)
Furthermore, as long as vector multiplication by $\auxkernel_{\xall\xall}$ takes $O(NM)$ time,
$\pfd{\auxdist}{\approxdist}{\targetdist} - C(\xall)$ can be computed in $O(NM^{2})$ time and $O(NM)$ space,
where $C(\xall)$ is a function that does not depend on $\indpts$. 
\enprop
We can interpret the three terms in the expression for $\pfd{\auxdist}{\approxdist}{\targetdist}$ quite naturally.
Term (I) measures how well the kernel matrix $\GPkernel_{\xall\xall}$ is approximated by the \Nystrom approximation $Q_{\xall\xall}$.
To interpret term (II) = (IIa) + (IIb), recall that the exact posterior kernel is 
$\postkernel(\x, \x') = \GPkernel(\x, \x') - \GPkernel_{\x\xall}[\GPkernel_{\xall\xall} + \sigma^{2}I]^{-1}\GPkernel_{\xall\x'}$. 
Observing the $(Q_{\xall\xall} + \sigma^{2}I)^{-1}$ terms in $S_{\xall\xall}$, we can see that term (II) measures how well 
the correction term $\GPkernel_{\x\xall}[\GPkernel_{\xall\xall} + \sigma^{2}I]^{-1}\GPkernel_{\xall\x'}$ is estimated. 
Finally, term (III) measures the quality of $\bar Q_{\xall\indpts} \auxmean_{\indpts}$, which is the inducing point approximation 
to the auxiliary mean function $\auxmean_{\xall}$. 
If we had to compute the terms in $\pfd{\auxdist}{\approxdist}{\targetdist}$ involving $\GPkernel_{\xall\xall}$, then the time and space complexity would be,
respectively, $O(MN^{2})$ and $O(N^{2})$. 
However, the terms that include $\GPkernel_{\xall\xall}$ do not depend on the inducing points, so we absorb them into $C(\xall)$.

\section{Experiments} \label{sec:experiments}

Our theory shows that \pfdtc can provide guarantees on the quality of 
the approximate GP posterior means and variances that practitioners typically report (\cref{sec:intro}).
We now check empirically that our method yields competitive estimates of
exact GP means and variances in practice.
We compare \pfdtc to to variational DTC (VFE)~\citep{Titsias:2009},  
subset of regressors (SoR)~\citep{Smola:2000}, and exact GP inference with a random subsample of the data (subsample). To compare to the exact GP as ground truth, all experiments use the same kernel (squared exponential kernel with separate length scales
for each dimension) and identical hyperparameters.
To that end, unless noted otherwise, we fix the kernel hyperparameters and the observation noise $\sigma^{2}$,
all of which we learn with a pilot run of VFE using 200 fixed randomly chosen inducing points.
We found that if we optimized the hyperparameters for \pfdtc and VFE after inducing
point optimization, they remained relatively stable. 
For \pfdtc we take $\RKHSkernel = \GPkernel$, as justified by \cref{prop:use-k-as-r}. 
Although constructing $\auxdist$ from a subset of data is currently the most justified choice theoretically~(see \cref{sec:auxiliary-distribution-choice}), 
preliminary experiments showed better empirical performance using SoR with a small random subset of the data. 

We used one synthetic and five real datasets with dimensionality ranging from 1 to 11. 
In order to run the exact GP on all datasets we subsampled the original airline delays dataset used
by \citet{Hensman:2013} down to 10,000 observations.
The remaining non-synthetic datasets were obtained from the UCI Machine Learning Repository. 
We held out the maximum of 1,000 observations or 20\% of each dataset for testing and 
used the remainder for training (training set sizes were between 1,000 and 8,000).
See \cref{tbl:datasets} in \cref{app:experiments} for full details.
We repeated all experiments 10 times. 

\textbf{Behavior of the {\pfdname}.}
The \pfdtc objective is non-convex, so we first check that we can effectively optimize it.
As seen in \cref{fig:pfdtc-objective}, for the airfoil dataset, optimizing the locations of the 200 inducing points
substantially reduces the size of the \pfdname, though clearly we are finding only local optima. 
The bounds in \cref{thm:pfd-controls-mean-and-variance} suggest a linear relationship between the \pfdname and 
the errors in estimates of the posterior mean $\postmean(\x)$ 
and standard deviation $\postkernel(\x, \x)^{1/2}$.
\cref{fig:meantrace,fig:stdtrace} show an approximately linear relationship
does in fact hold in both cases.

\opt{NOTUSED}{
\begin{figure}
\begin{center}
\begin{subfigure}[b]{.3\textwidth} 
    \includegraphics[height=105pt]{synth2-legend} 
    \includegraphics[height=105pt]{synth3}
    \caption{synthetic}
    \label{fig:synthetic}
\end{subfigure}  
\begin{subfigure}[b]{.3\textwidth} 
    \includegraphics[height=105pt]{delays2} 
    \includegraphics[height=105pt]{delays3}
    \caption{delays10k}
    \label{fig:delays}
\end{subfigure} 
\begin{subfigure}[b]{.3\textwidth} 
    \includegraphics[height=105pt]{abalone2} 
    \includegraphics[height=105pt]{abalone3}
    \caption{abalone}
    \label{fig:abalone}
\end{subfigure}  \\
\begin{subfigure}[b]{.3\textwidth} 
    \includegraphics[height=105pt]{airfoil2} 
    \includegraphics[height=105pt]{airfoil3}
    \caption{airfoil}
    \label{fig:airfoil}
\end{subfigure} 
\begin{subfigure}[b]{.3\textwidth} 
    \includegraphics[height=105pt]{ccpp2} 
    \includegraphics[height=105pt]{ccpp3}
    \caption{CCPP}
    \label{fig:CCPP}
\end{subfigure} 
\begin{subfigure}[b]{.3\textwidth} 
    \includegraphics[height=105pt]{wine2} 
    \includegraphics[height=105pt]{wine3}
    \caption{wine}
    \label{fig:wine}
\end{subfigure} 
\end{center}
\caption{Root mean squared error of the approximate posterior mean (left) and standard deviation (right) 
at the held-out test locations.}
\label{fig:mean-and-stdev-error}
\end{figure}

\begin{figure}
\begin{center}
\begin{subfigure}[b]{.3\textwidth} 
    \includegraphics[trim={0 0 0 0},clip,height=100pt]{synth1}
    \caption{synthetic}
    \label{fig:prediction-synthetic}
\end{subfigure} 
\begin{subfigure}[b]{.3\textwidth} 
    \includegraphics[trim={0 0 0 0},clip,height=100pt]{delays1} 
    \caption{delays10k}
    \label{fig:prediction-delays} 
\end{subfigure}  
\begin{subfigure}[b]{.3\textwidth} 
    \includegraphics[trim={0 0 0 0},clip,height=100pt]{abalone1}
    \caption{abalone}
    \label{fig:prediction-abalone}
\end{subfigure} \\
\begin{subfigure}[b]{.3\textwidth} 
    \includegraphics[trim={0 0 0 0},clip,height=100pt]{airfoil1} 
    \caption{airfoil}
    \label{fig:prediction-airfoil}
\end{subfigure} 
\begin{subfigure}[b]{.3\textwidth} 
    \includegraphics[trim={0 0 0 0},clip,height=100pt]{ccpp1}
    \caption{CCPP}
    \label{fig:prediction-ccpp}
\end{subfigure} 
\begin{subfigure}[b]{.3\textwidth} 
    \includegraphics[trim={0 0 0 0},clip,height=100pt]{wine1-legend} 
    \caption{wine}
    \label{fig:prediction-wine}
\end{subfigure} 
\end{center}
\caption{Predictive performance as measured by root mean squared error.}
\label{fig:prediction-error}
\end{figure}
}

\textbf{Mean and Variance Estimates.}
We first consider a simple one-dimensional synthetic example to compare \pfdtc to VFE and SoR. 
\cref{fig:synthetic-fit} shows that, with 9 inducing points, \pfdtc and VFE
produce excellent, almost identical fits, while SoR performs
substantially worse at estimating both the posterior mean and standard deviation. 
\citet{Bauer:2016} found similarly poor performance for the fully independence training conditional (FITC) method of \citet{Snelson:2005}.

On more complex problems, we consider root mean squared error (RMSE) of posterior mean
and standard deviation estimates at all held-out test locations, as we vary the number of 
inducing points.
\cref{fig:mean-and-stdev-error} confirms that \pfdtc is competitive with VFE, validating the 
practicality of the \pfdname as an objective for approximate inference. 
VFE shows better performance for larger numbers of inducing points 
once the error dips below $10^{-3}$, which we suspect is related to numerical issues. 
SoR shows surprisingly good performance on standard deviation error but worse
mean estimation---particularly on abalone. 
As expected, subsampling performed poorly. 

\begin{figure}[H]
\begin{center}
\begin{subfigure}[b]{.45\textwidth} 
    \includegraphics[trim={0 0 0 0},clip,height=90pt]{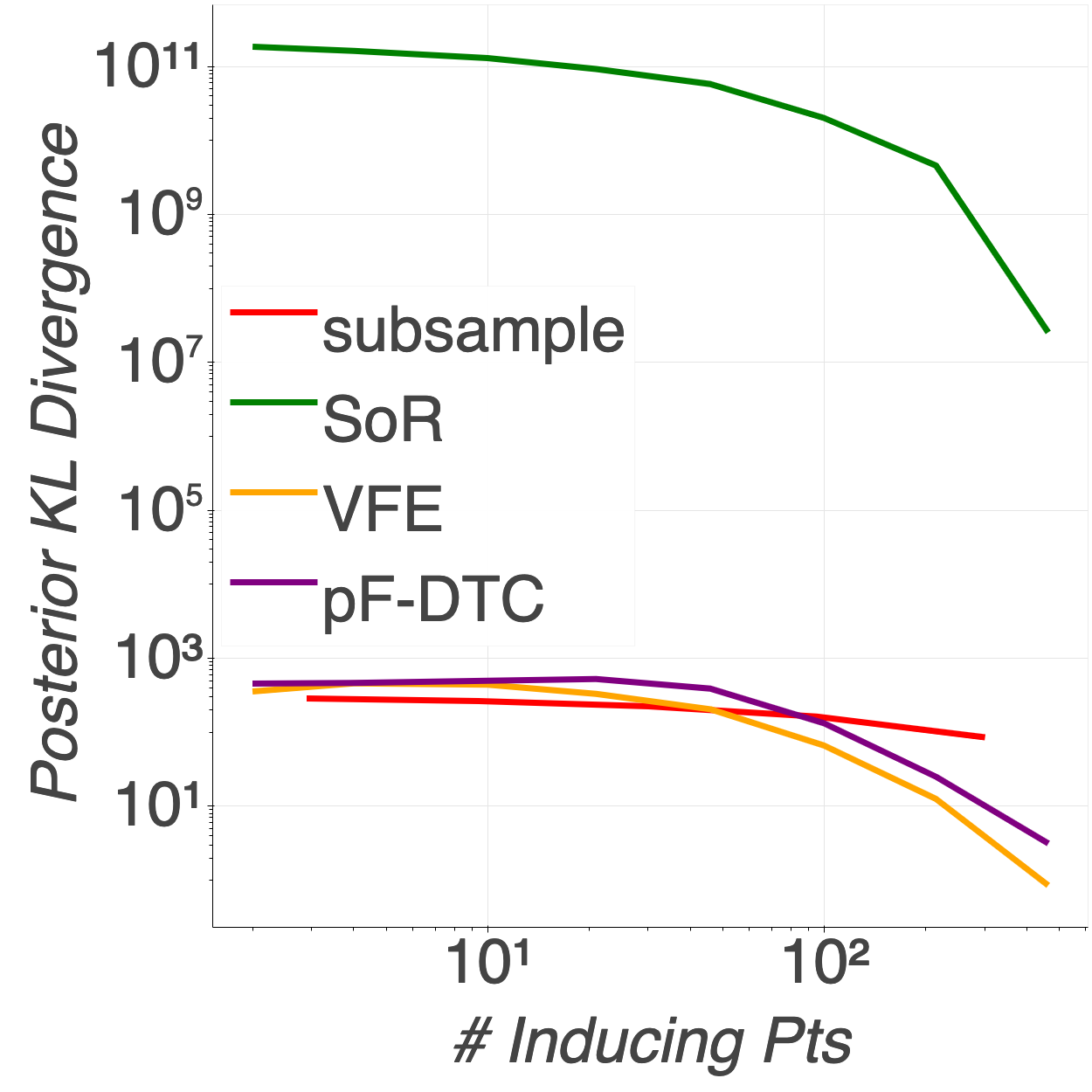} 
     \includegraphics[trim={0 0 0 0},clip,height=90pt]{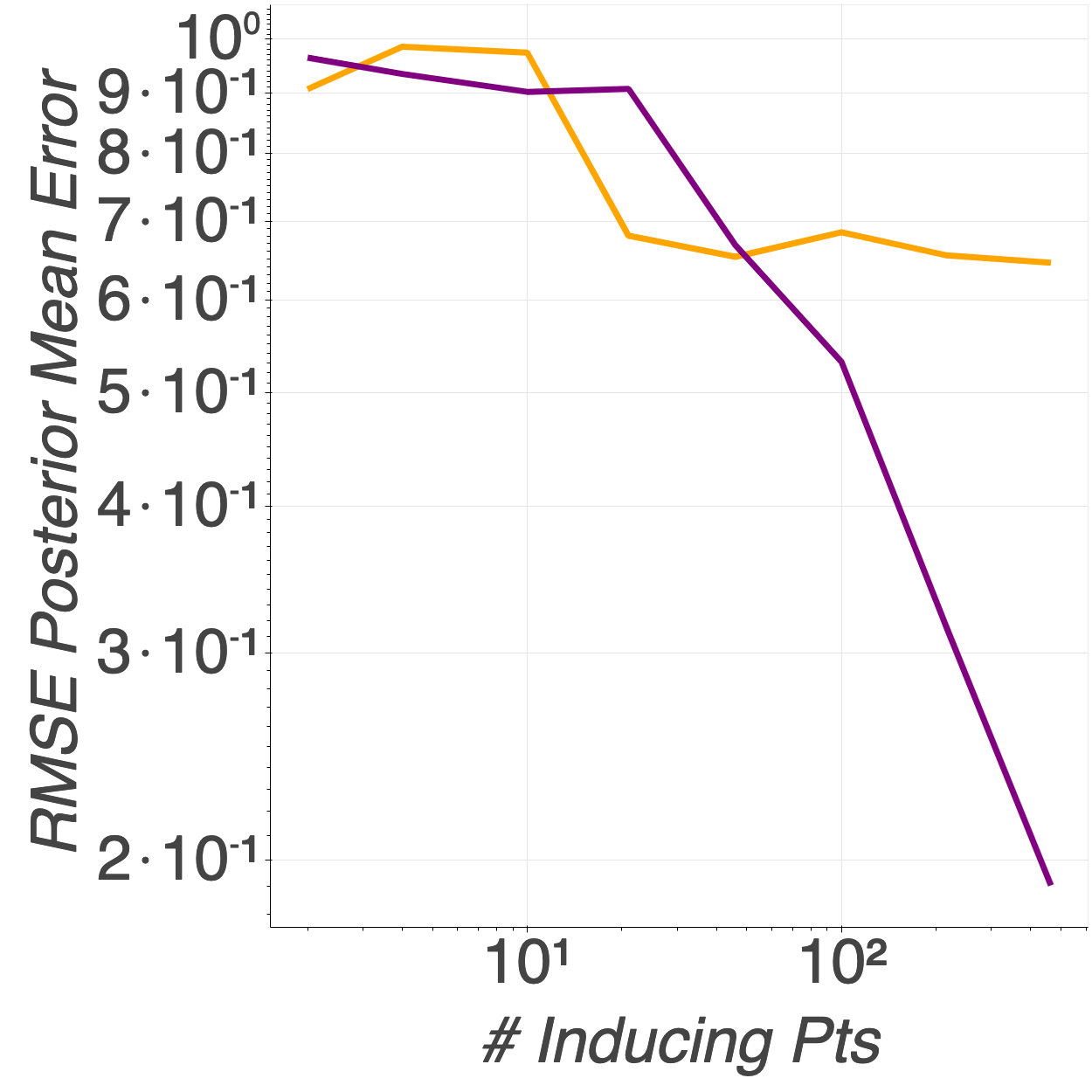} 
    \caption{airfoil}
    \label{fig:airfoil-KL-and-mean-best} 
\end{subfigure}  \\
\begin{subfigure}[b]{.45\textwidth} 
    \includegraphics[trim={0 0 0 0},clip,height=90pt]{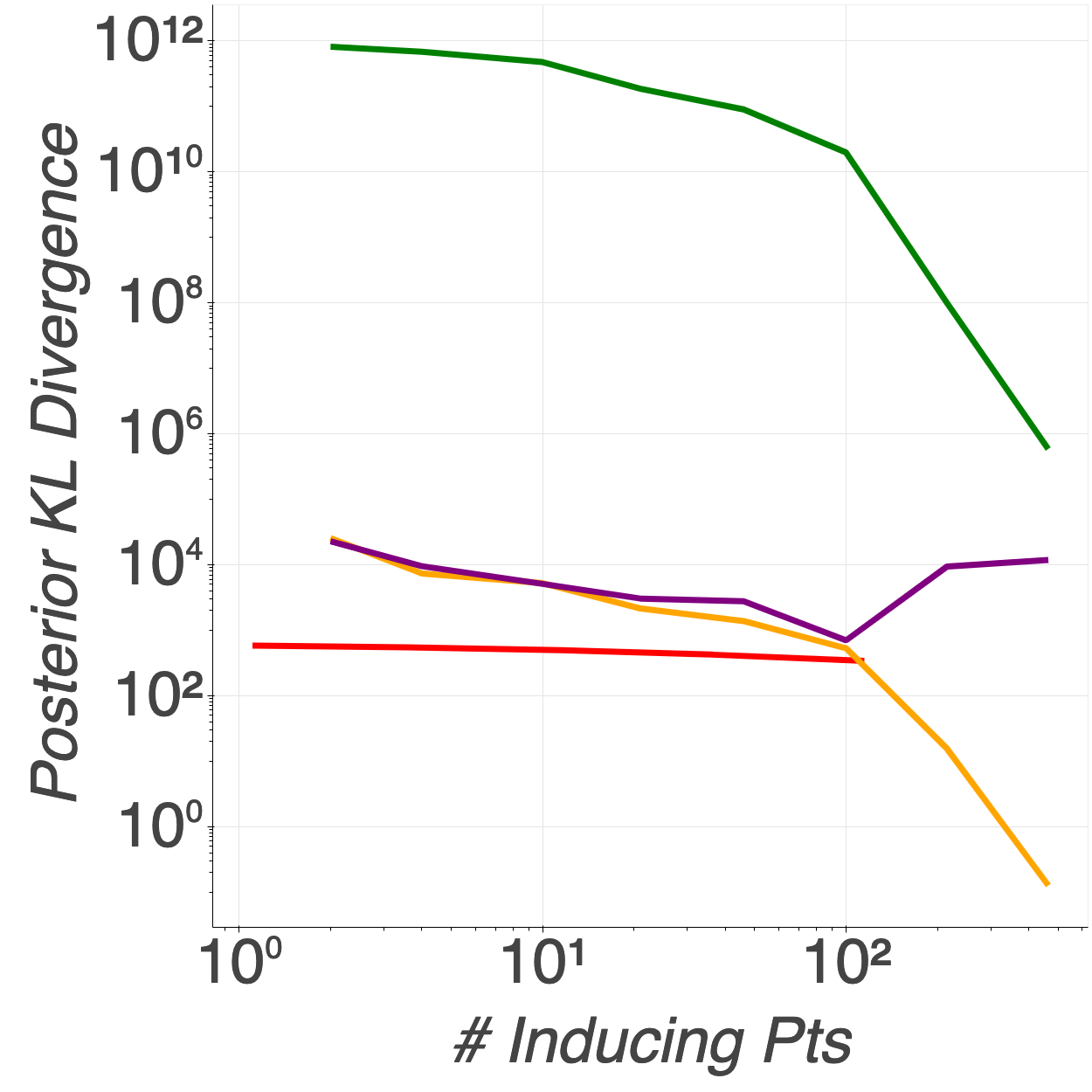}
    \includegraphics[trim={0 0 0 0},clip,height=90pt]{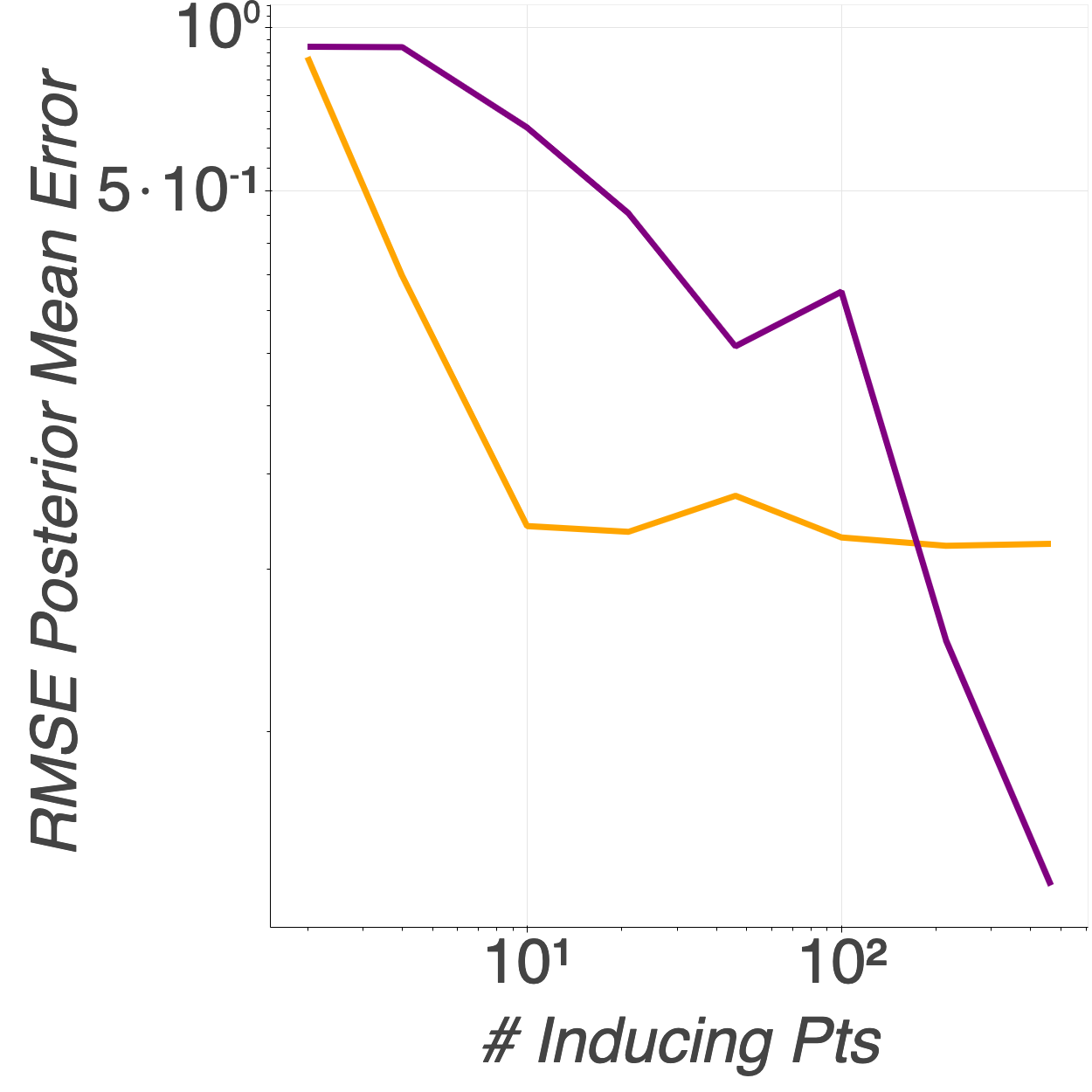}
    \caption{CCPP}
    \label{fig:ccpp-KL-and-mean-best}
\end{subfigure} 
\end{center}
\caption{KL divergences of the approximate posteriors and root mean squared error of the approximate
posteriors for the VFE and pF-DTC trials with the smallest objective values.}
\label{fig:KL-and-mean-best}
\vspace{-1em}
\end{figure}

\textbf{KL divergence and comparison of objective functions.}
In \cref{sec:wasserstein-and-fisher-distances} we argued that KL divergence is a suboptimal way to measure 
posterior approximation quality when the goal is good posterior mean and standard deviation estimates.
We here show that this issue arises in practice.
\cref{fig:KL-and-mean-best,fig:KL-and-mean-and-stdev-best-part-1,fig:KL-and-mean-and-stdev-best-part-2} show the KL divergences for all four approximations. 
We see that VFE sometimes yields small KL divergence but worse posterior mean and standard deviation estimates. Similarly, we see that \pfdtc can exhibit larger KL divergence values but very good posterior mean and standard deviation estimates. 

\textbf{Predictive Performance.} 
While our focus in this paper is on posterior mean and variance estimates,
we also used the held-out test data to check RMSE predictive performance. 
As shown in \cref{fig:prediction-error}, \pfdtc, SoR, and VFE performed quite similarly, 
with subsampling providing worse predictive accuracy. 

\textbf{Computation.} In over 80\% of our 60 experiments (10 experiments per dataset), \pfdtc used less computation than VFE.
On average, VFE was 4.5 times slower than \pfdtc.
Because they did not involve any optimization, SoR and subsample were orders of magnitude faster.

\section{Discussion} \label{sec:discussion}

In this paper we have developed an approach to scalable Gaussian process regression using a novel objective, the preconditioned Fisher divergence,
which we show bounds the 2-Wasserstein distance. 
We were motivated by the need to guarantee the finite-data accuracy of posterior mean and covariance function estimates.
Empirically we showed that using the \pfdname with a DTC likelihood approximation provides competitive mean and variance estimates with state-of-the-art approaches.

The current work demonstrates the feasibility of using the \pfdname, but there are many important directions for 
future work.
The first is to scale to larger datasets, e.g.\ using stochastic optimization techniques similar in spirit to SVI for GPs~\citep{Hensman:2013}.
A second direction is to use the \pfdname objective with other likelihood approximations.
Structured kernel approximations are of particular interest
since these are both fast to compute and can often better capture multi-scale structure~\citep{Wilson:2015,Gardner:2018}. 
Finally, it remains to efficiently utilize the preconditioned Fisher divergence for other likelihoods, 
such as in GP classification.

\begin{figure}[H]
\begin{center}
\begin{subfigure}[b]{.22\textwidth} 
    \includegraphics[trim={0 0 0 0},clip,height=100pt]{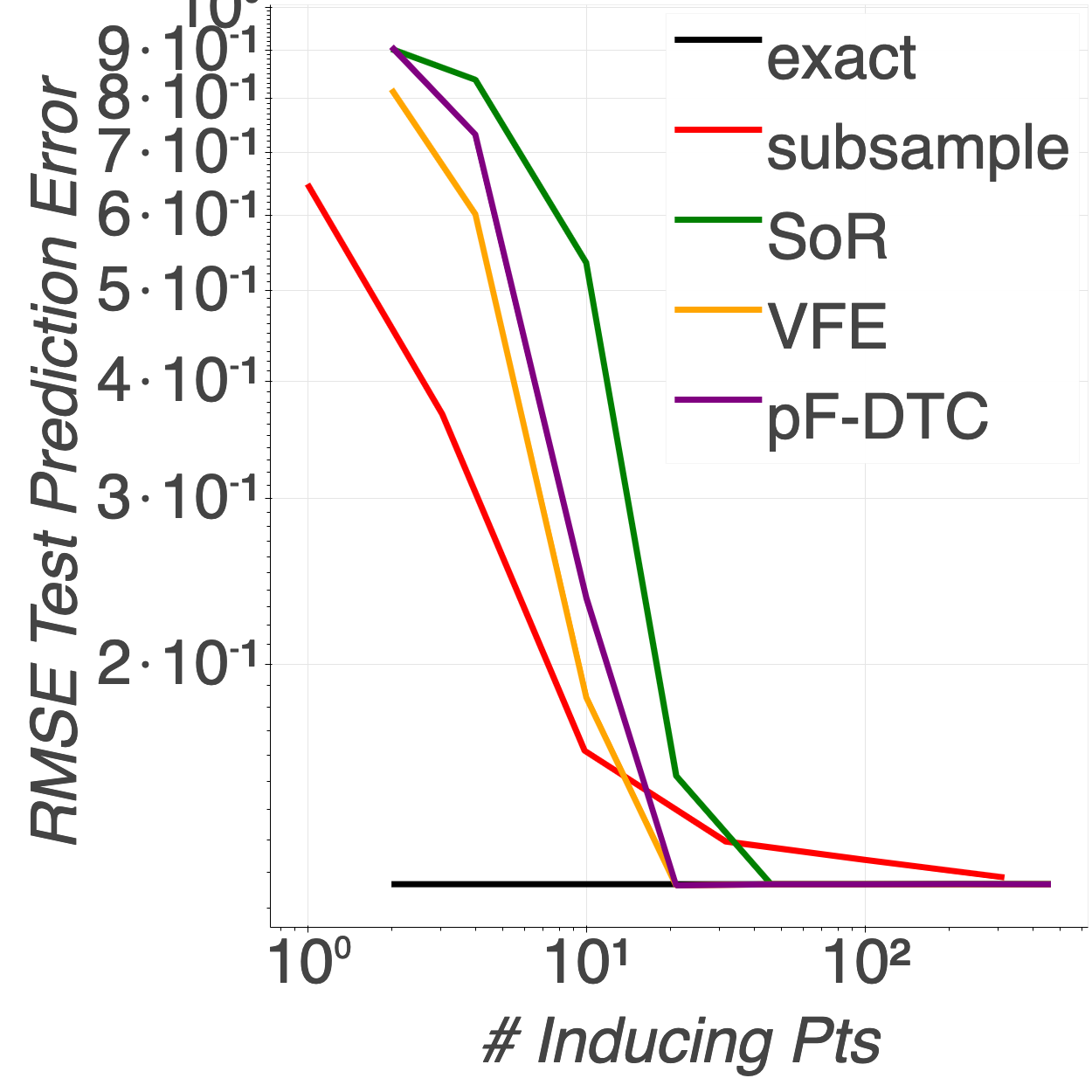}
    \caption{synthetic}
    \label{fig:prediction-synthetic}
\end{subfigure} 
\begin{subfigure}[b]{.22\textwidth} 
    \includegraphics[trim={0 0 0 0},clip,height=100pt]{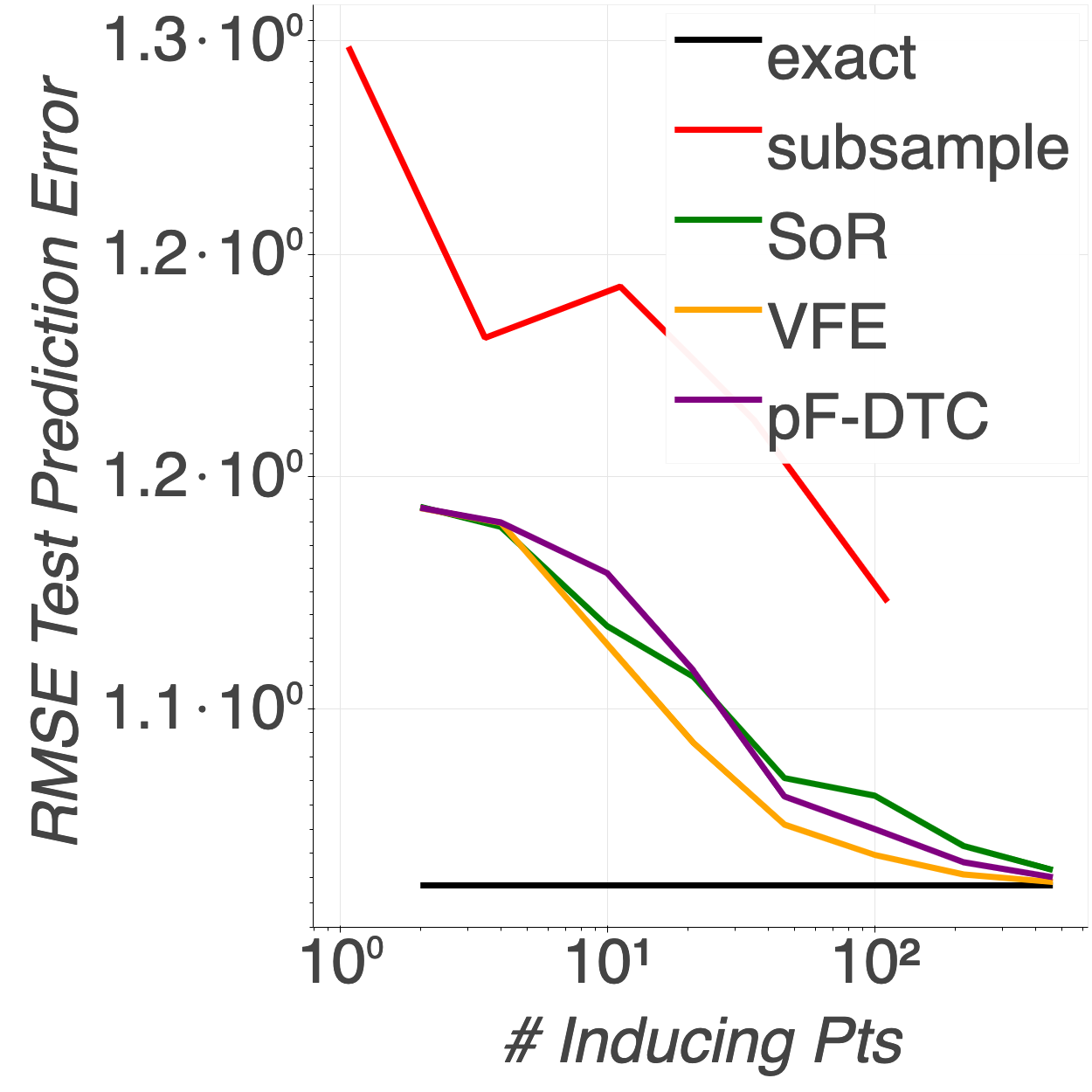} 
    \caption{delays10k}
    \label{fig:prediction-delays} 
\end{subfigure}  \\
\begin{subfigure}[b]{.22\textwidth} 
    \includegraphics[trim={0 0 0 0},clip,height=100pt]{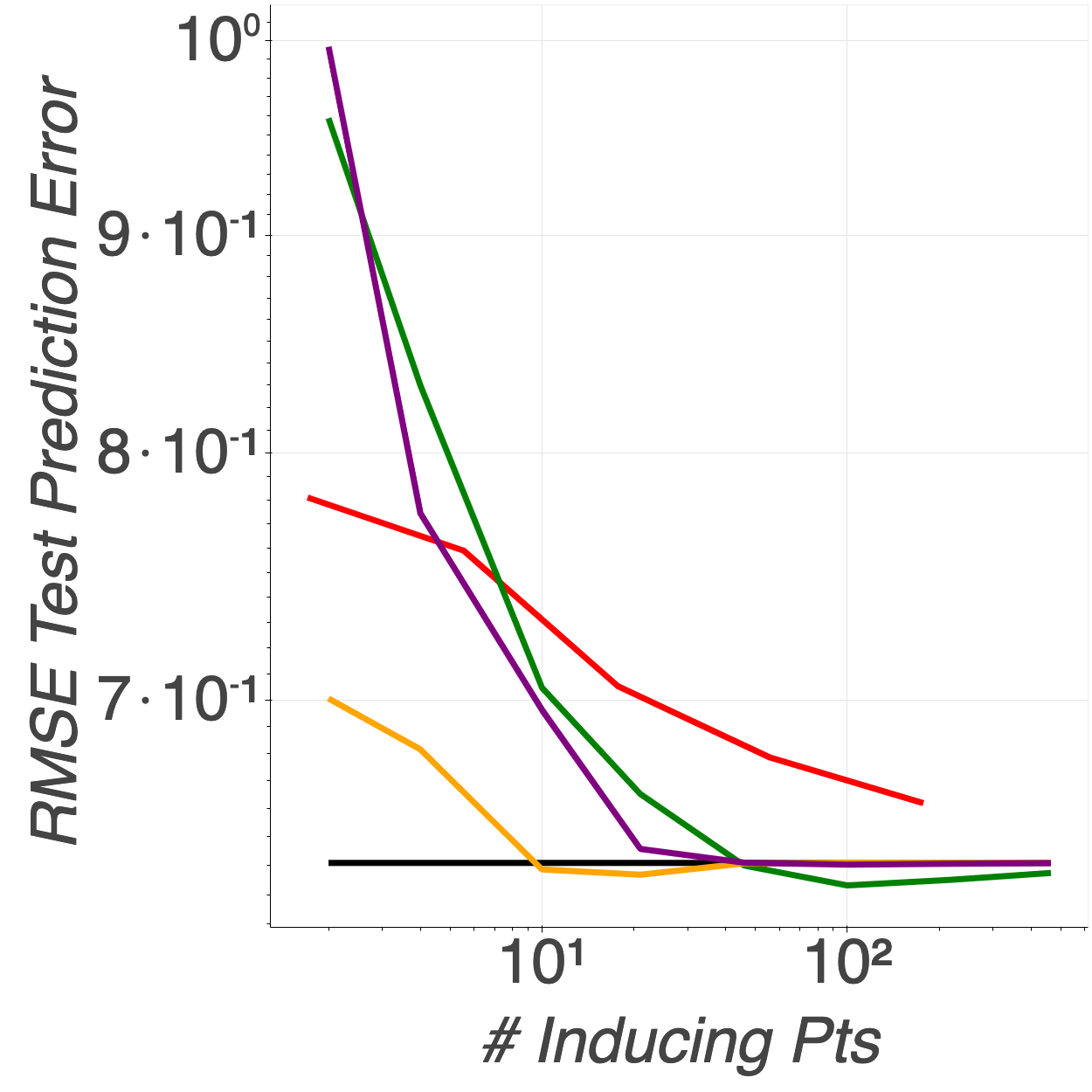}
    \caption{abalone}
    \label{fig:prediction-abalone}
\end{subfigure} 
\begin{subfigure}[b]{.22\textwidth} 
    \includegraphics[trim={0 0 0 0},clip,height=100pt]{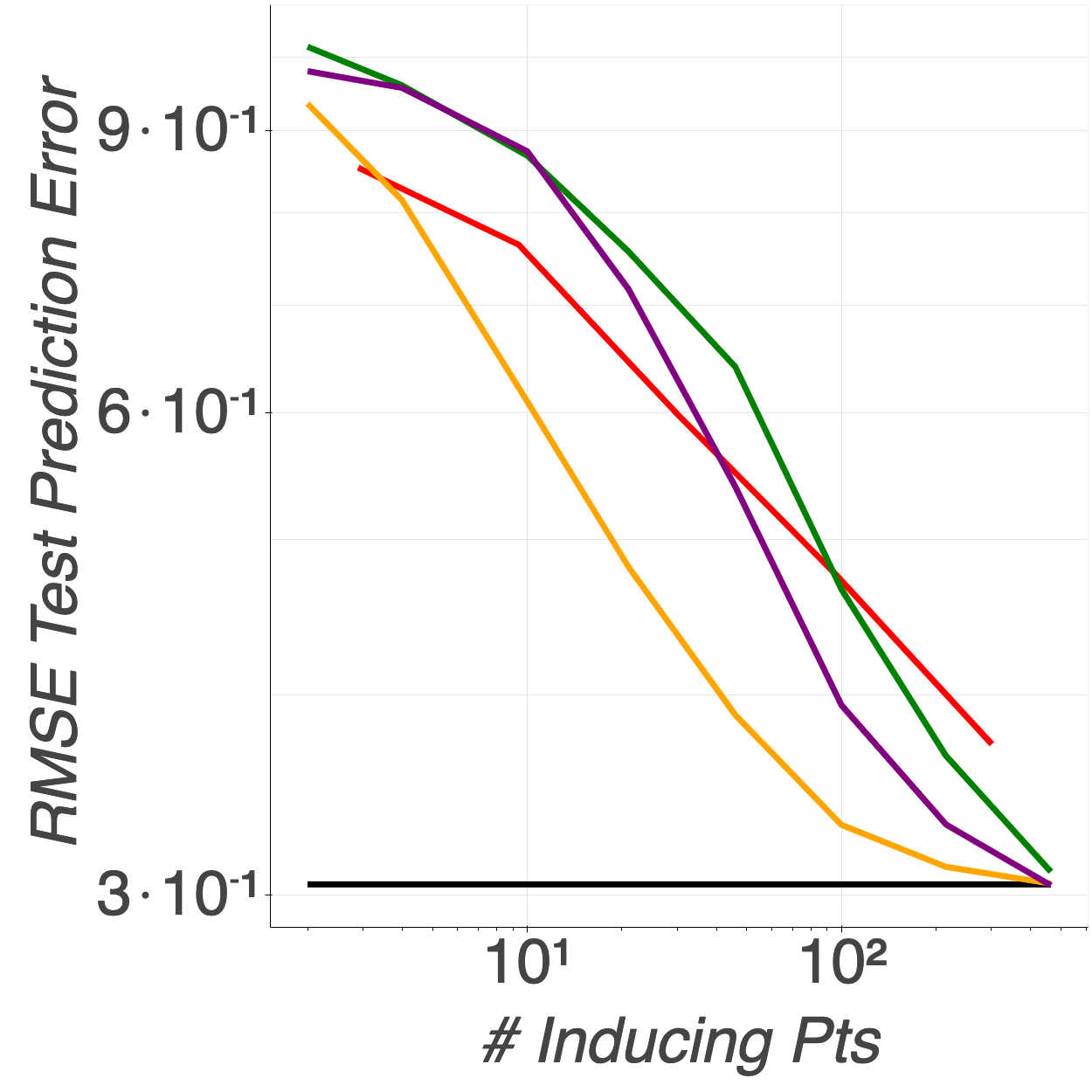} 
    \caption{airfoil}
    \label{fig:prediction-airfoil}
\end{subfigure} \\
\begin{subfigure}[b]{.22\textwidth} 
    \includegraphics[trim={0 0 0 0},clip,height=100pt]{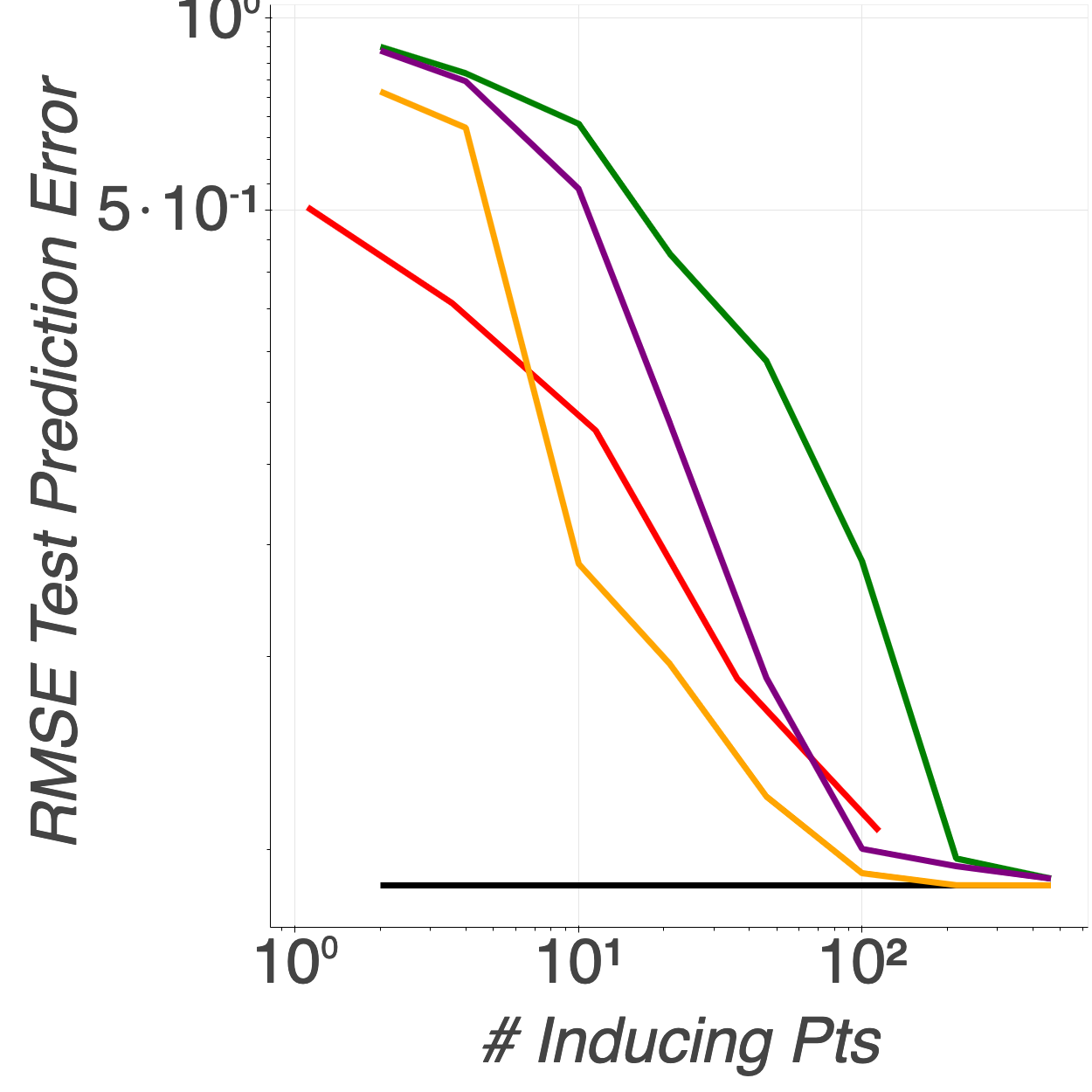}
    \caption{CCPP}
    \label{fig:prediction-ccpp}
\end{subfigure} 
\begin{subfigure}[b]{.22\textwidth} 
    \includegraphics[trim={0 0 0 0},clip,height=100pt]{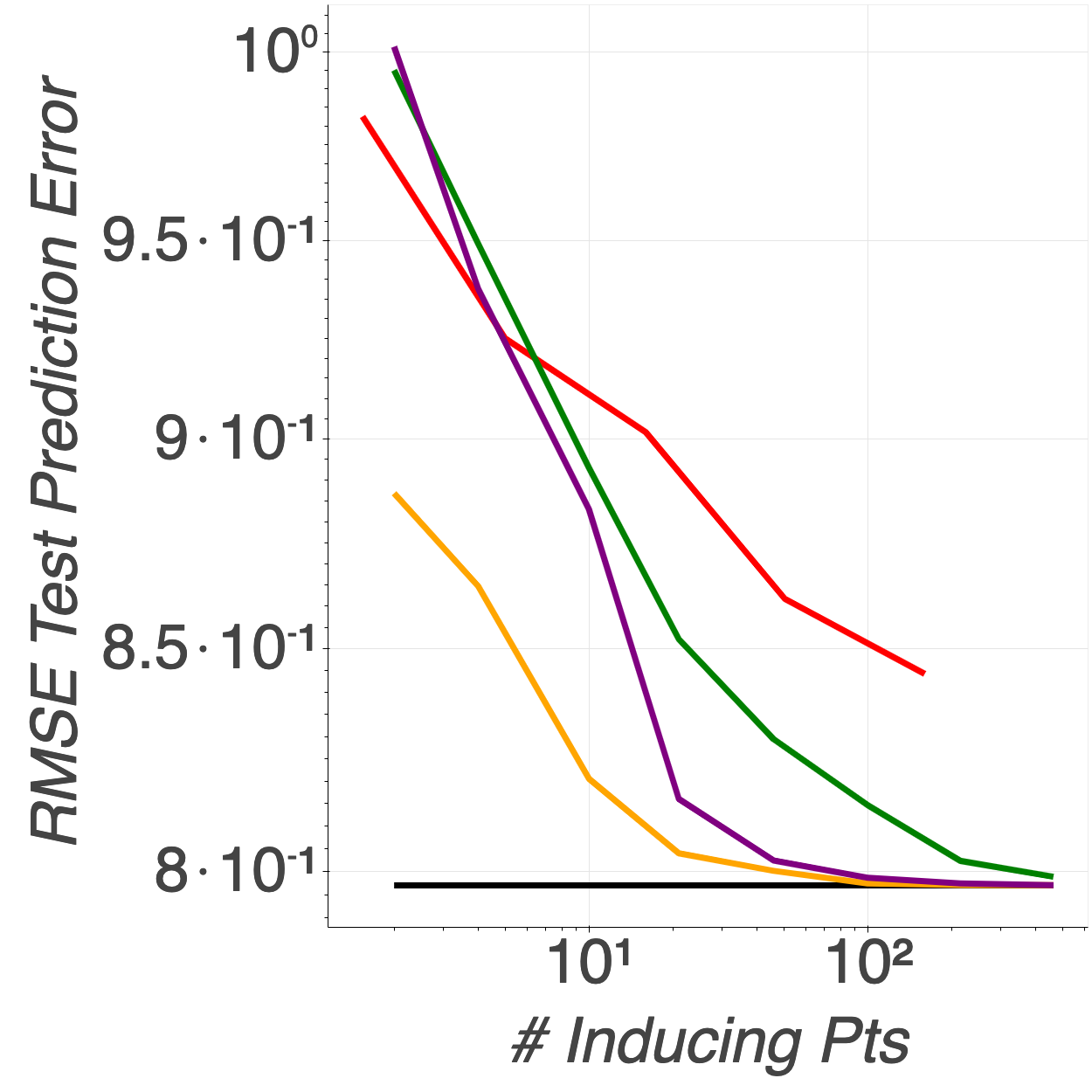} 
    \caption{wine}
    \label{fig:prediction-wine}
\end{subfigure} 
\end{center}
\caption{Predictive performance as measured by root mean squared error.}
\label{fig:prediction-error}
\vspace{-1em}
\end{figure}

\begin{figure}[H]
\begin{center}
\begin{subfigure}[b]{.45\textwidth} 
    \includegraphics[height=90pt]{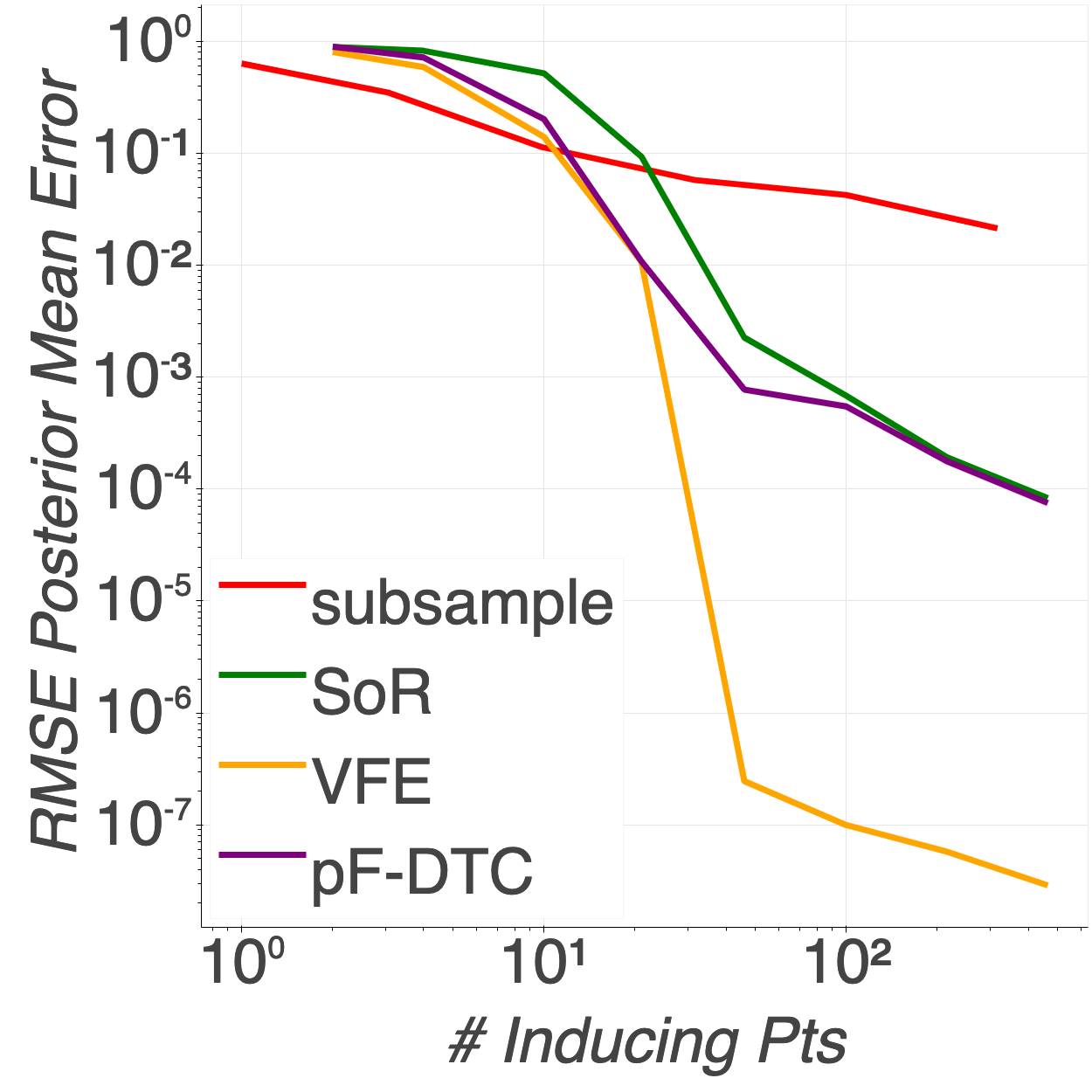} 
    \includegraphics[height=90pt]{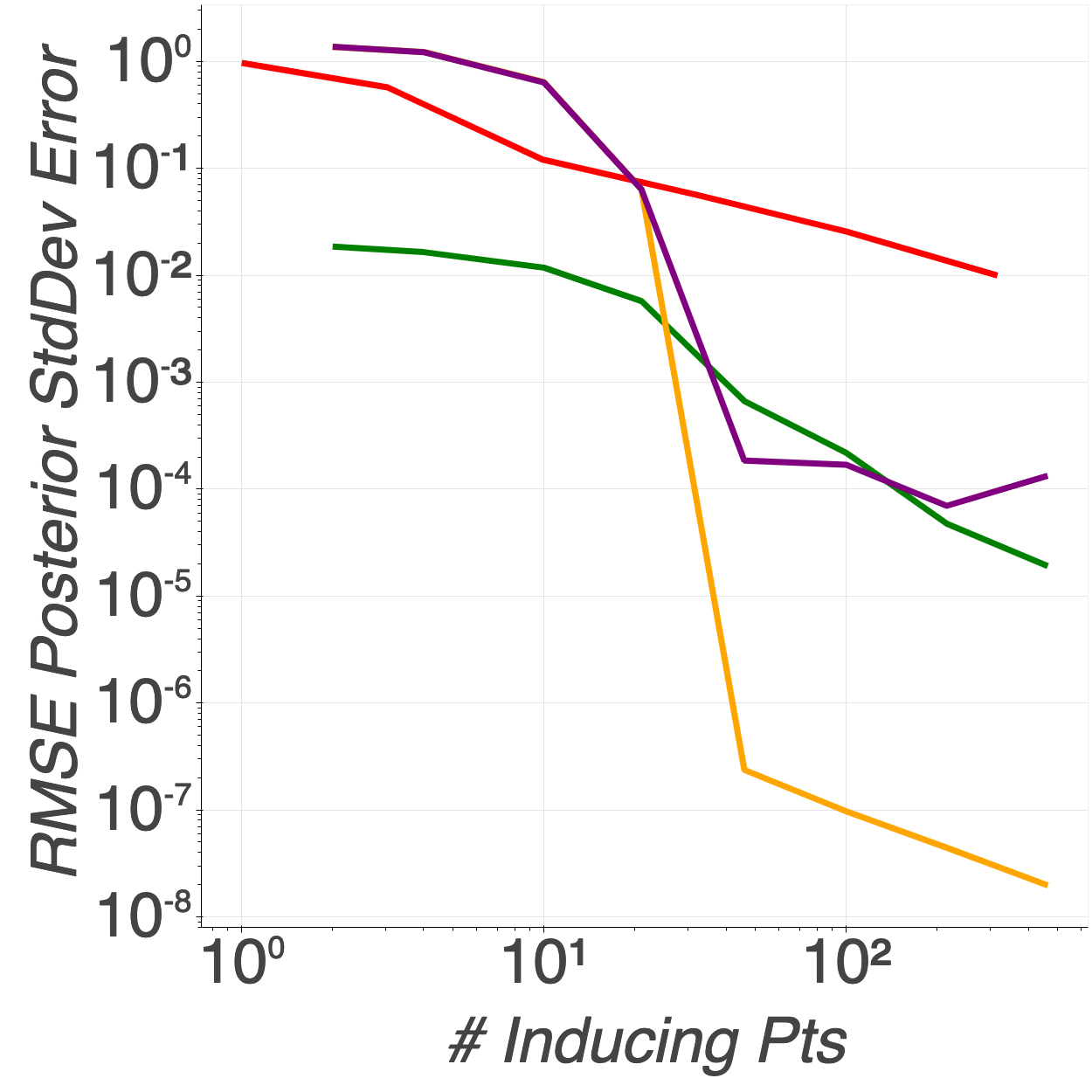}
    \caption{synthetic}
    \label{fig:synthetic}
\end{subfigure}  \\
\begin{subfigure}[b]{.45\textwidth} 
    \includegraphics[height=90pt]{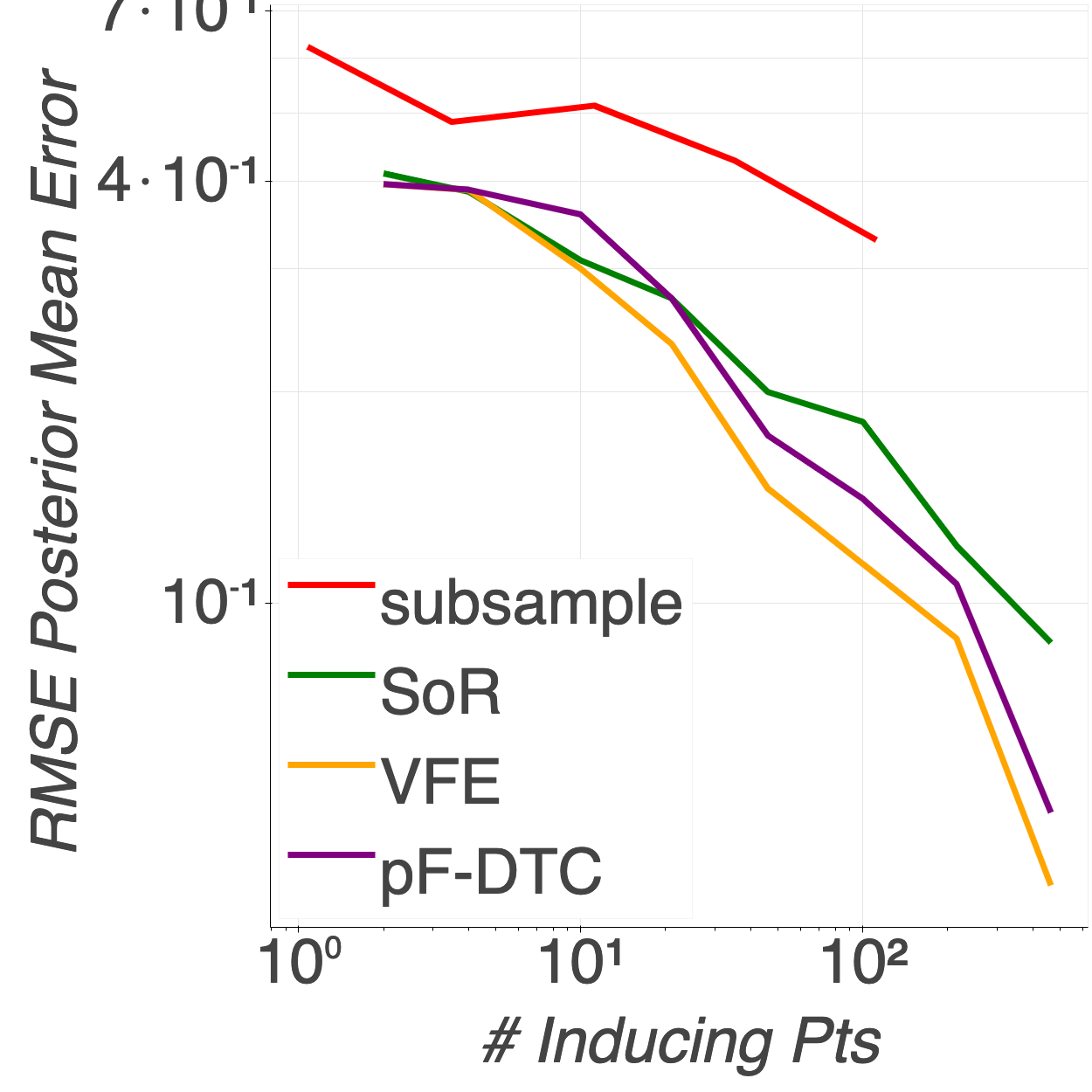} 
    \includegraphics[height=90pt]{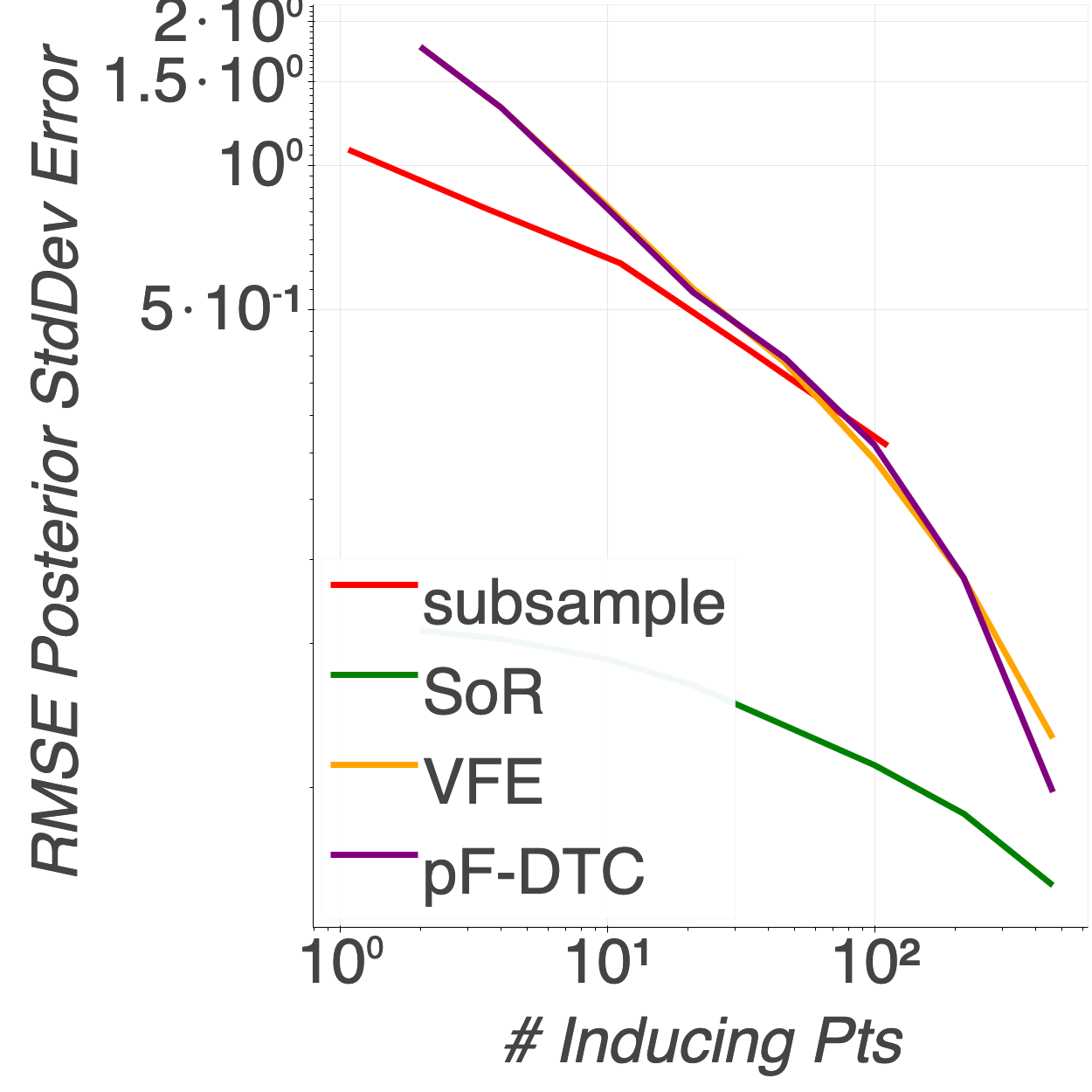}
    \caption{delays10k}
    \label{fig:delays}
\end{subfigure} \\
\begin{subfigure}[b]{.45\textwidth} 
    \includegraphics[height=90pt]{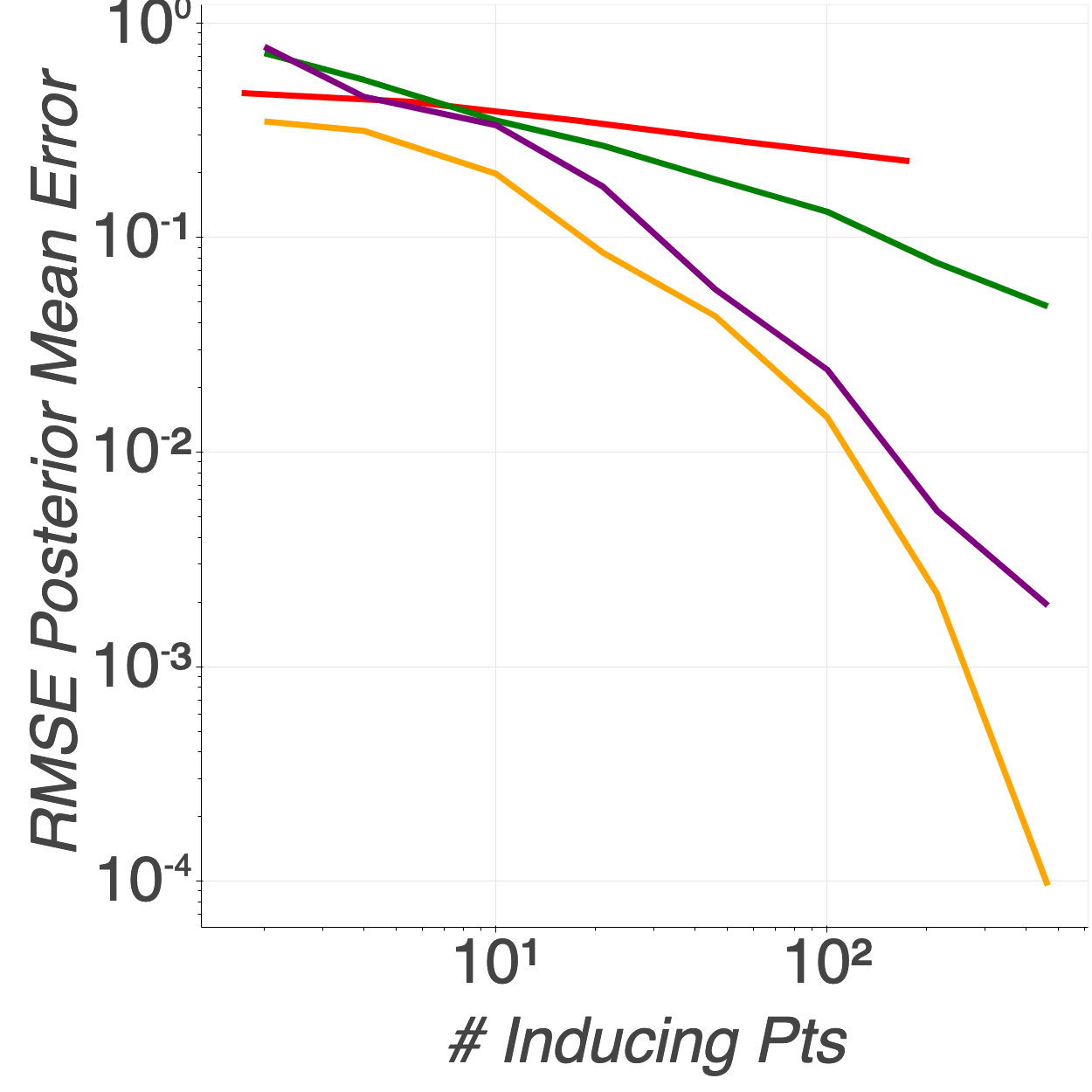} 
    \includegraphics[height=90pt]{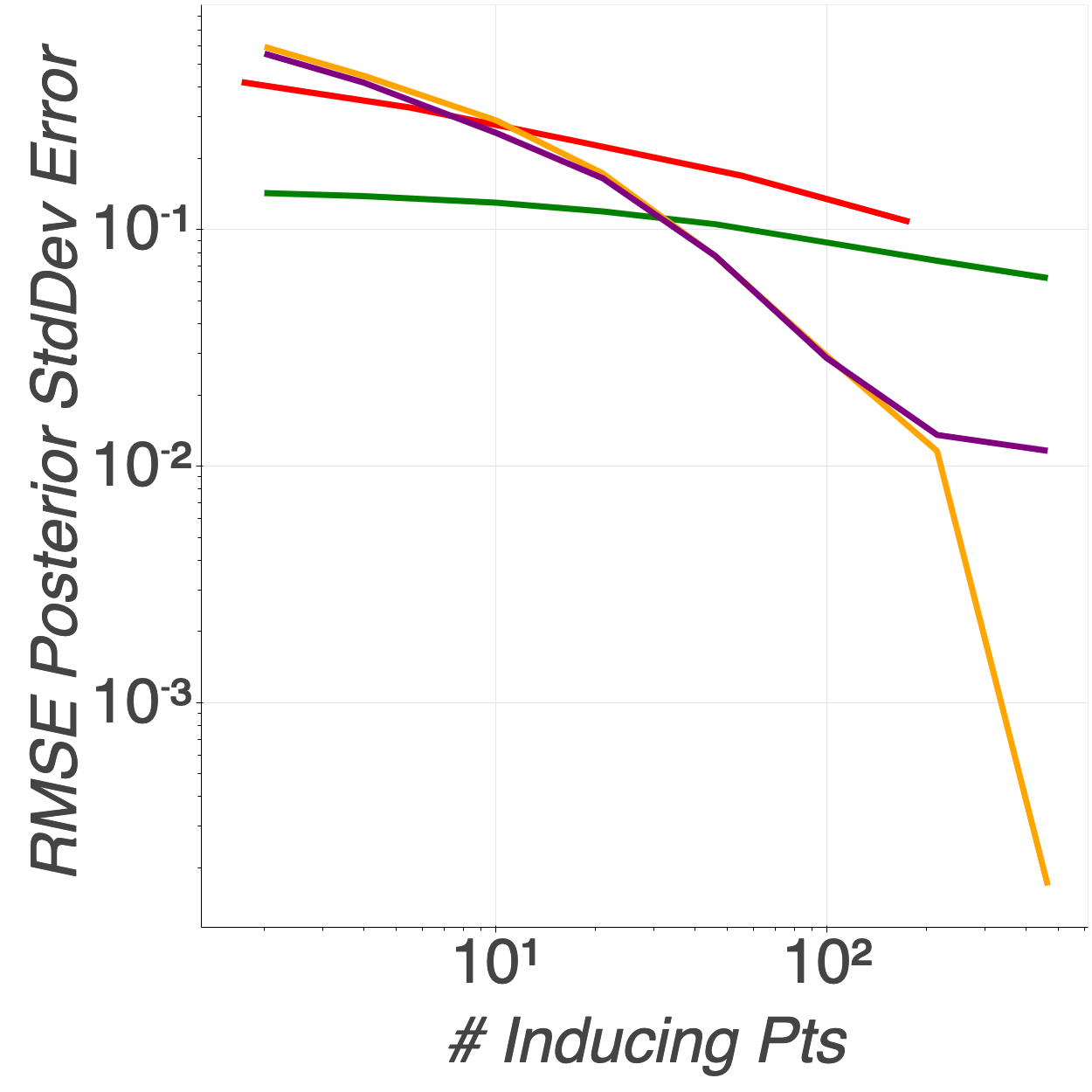}
    \caption{abalone}
    \label{fig:abalone}
\end{subfigure} \\
\begin{subfigure}[b]{.45\textwidth} 
    \includegraphics[height=90pt]{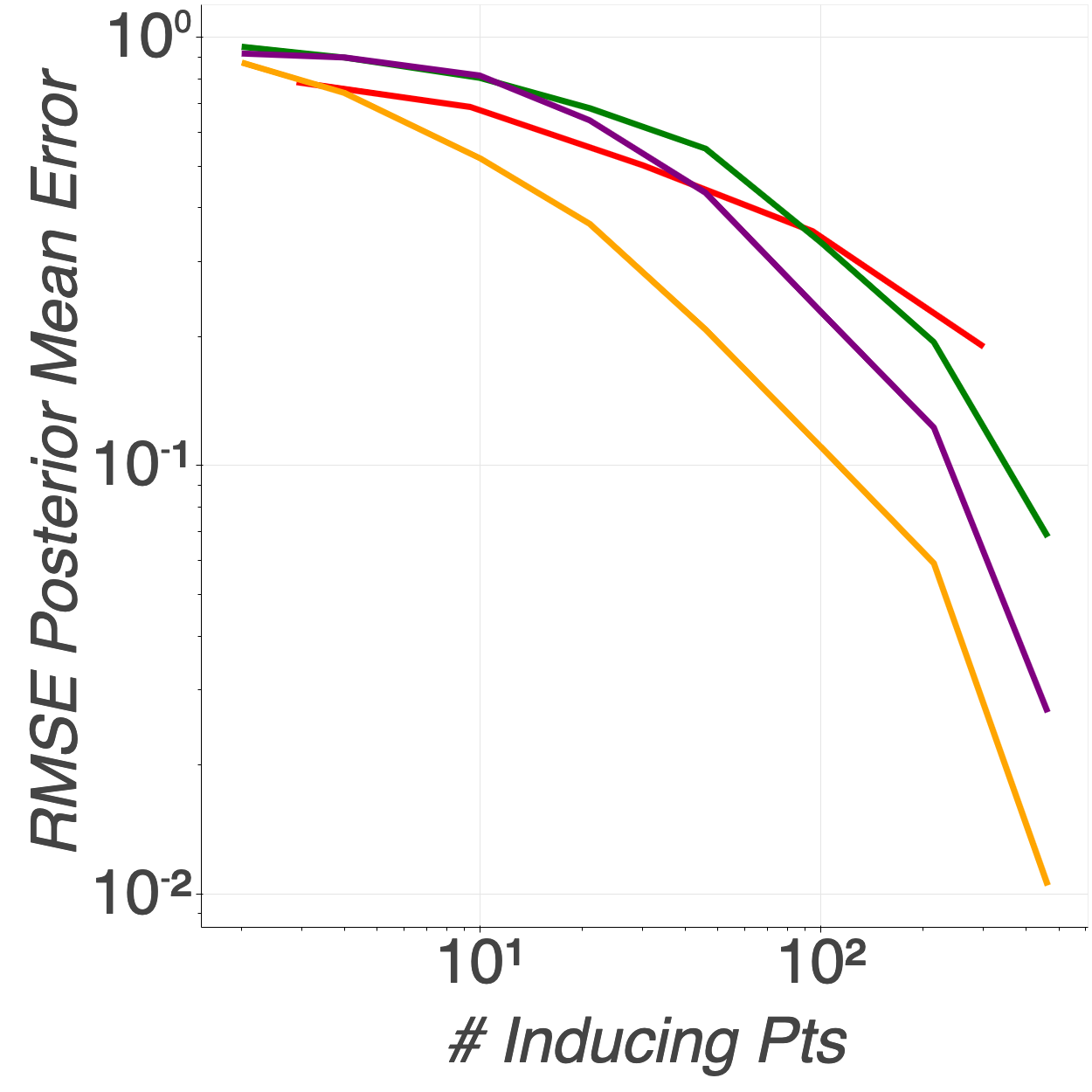} 
    \includegraphics[height=90pt]{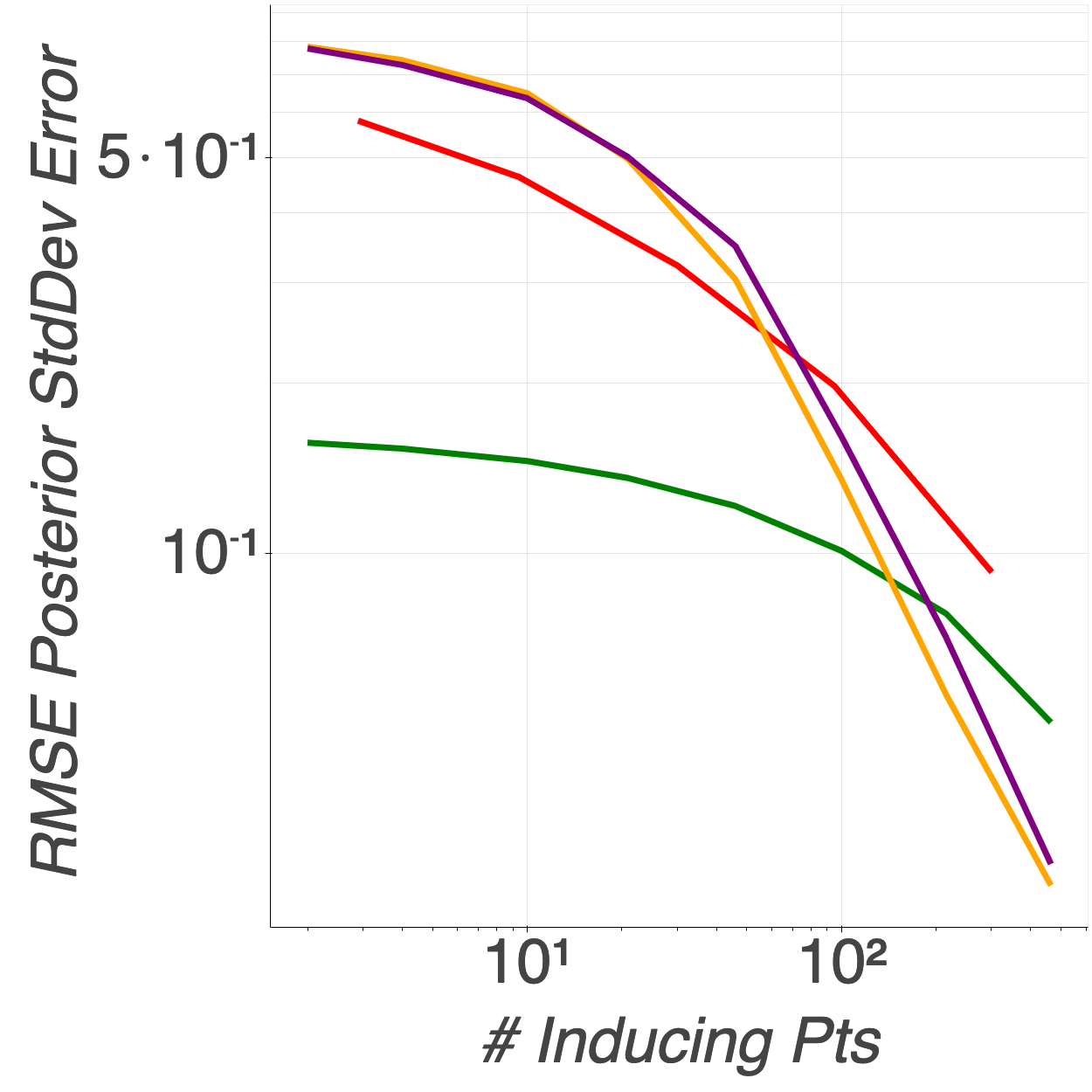}
    \caption{airfoil}
    \label{fig:airfoil}
\end{subfigure} \\
\begin{subfigure}[b]{.45\textwidth} 
    \includegraphics[height=90pt]{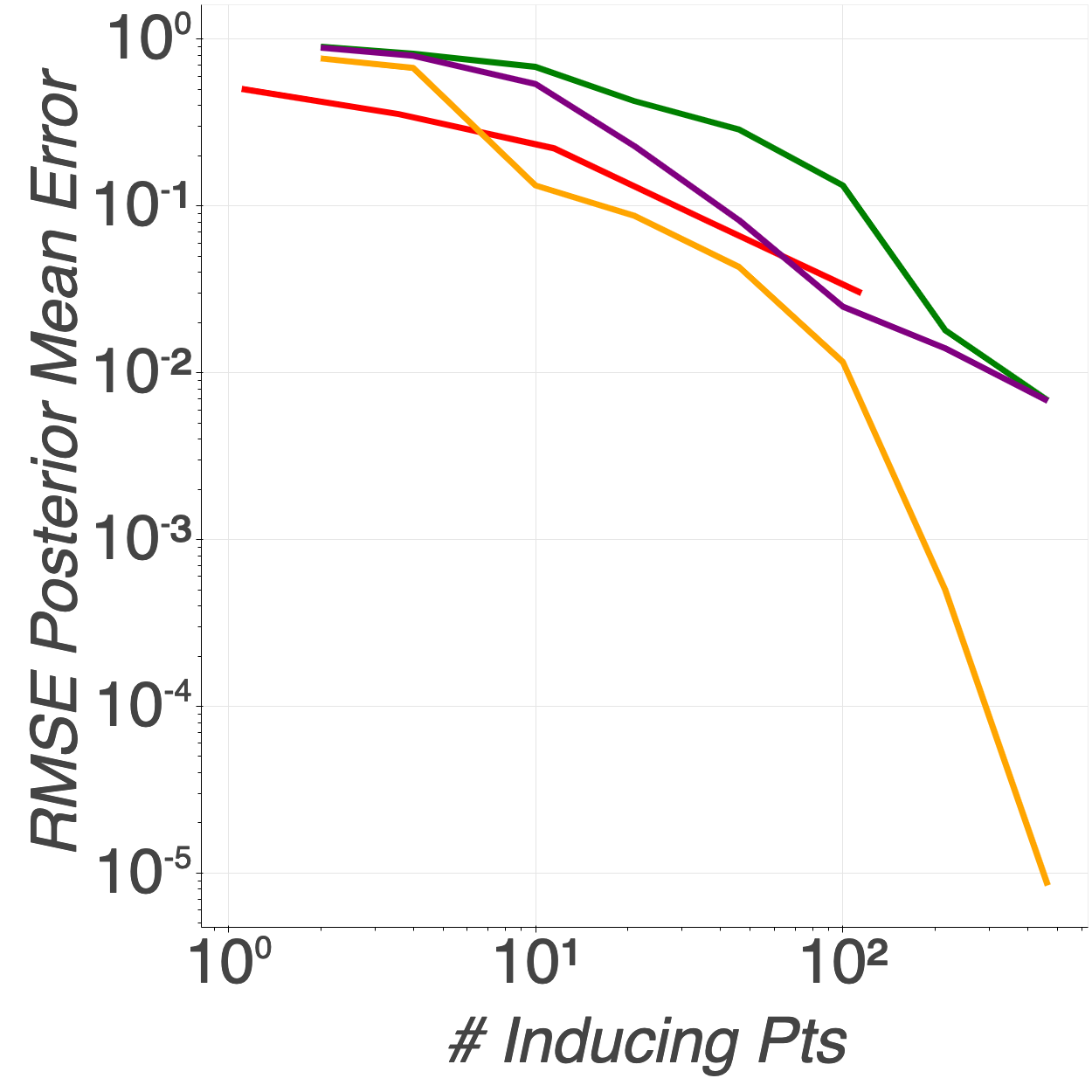} 
    \includegraphics[height=90pt]{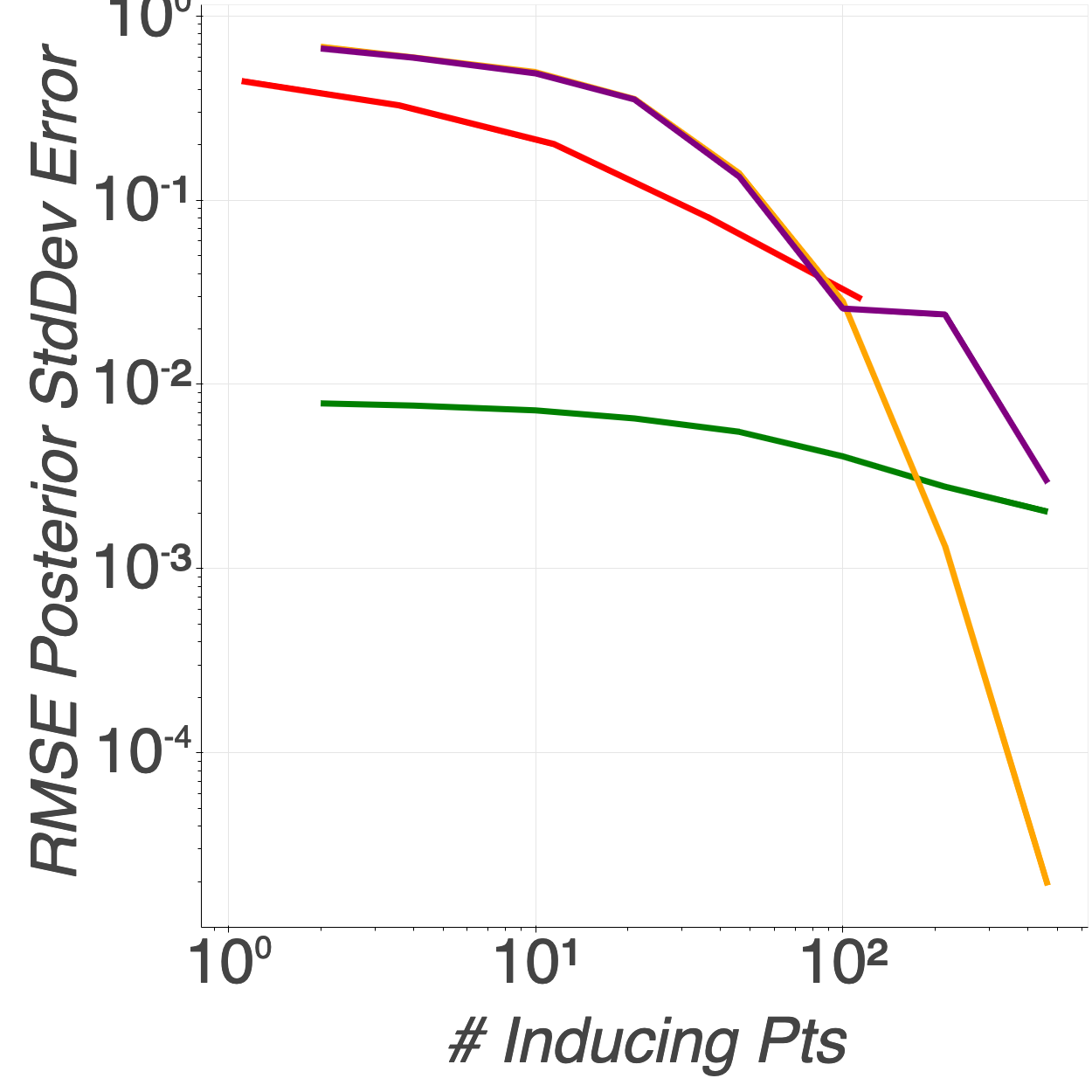}
    \caption{CCPP}
    \label{fig:CCPP}
\end{subfigure} \\
\begin{subfigure}[b]{.45\textwidth} 
    \includegraphics[height=90pt]{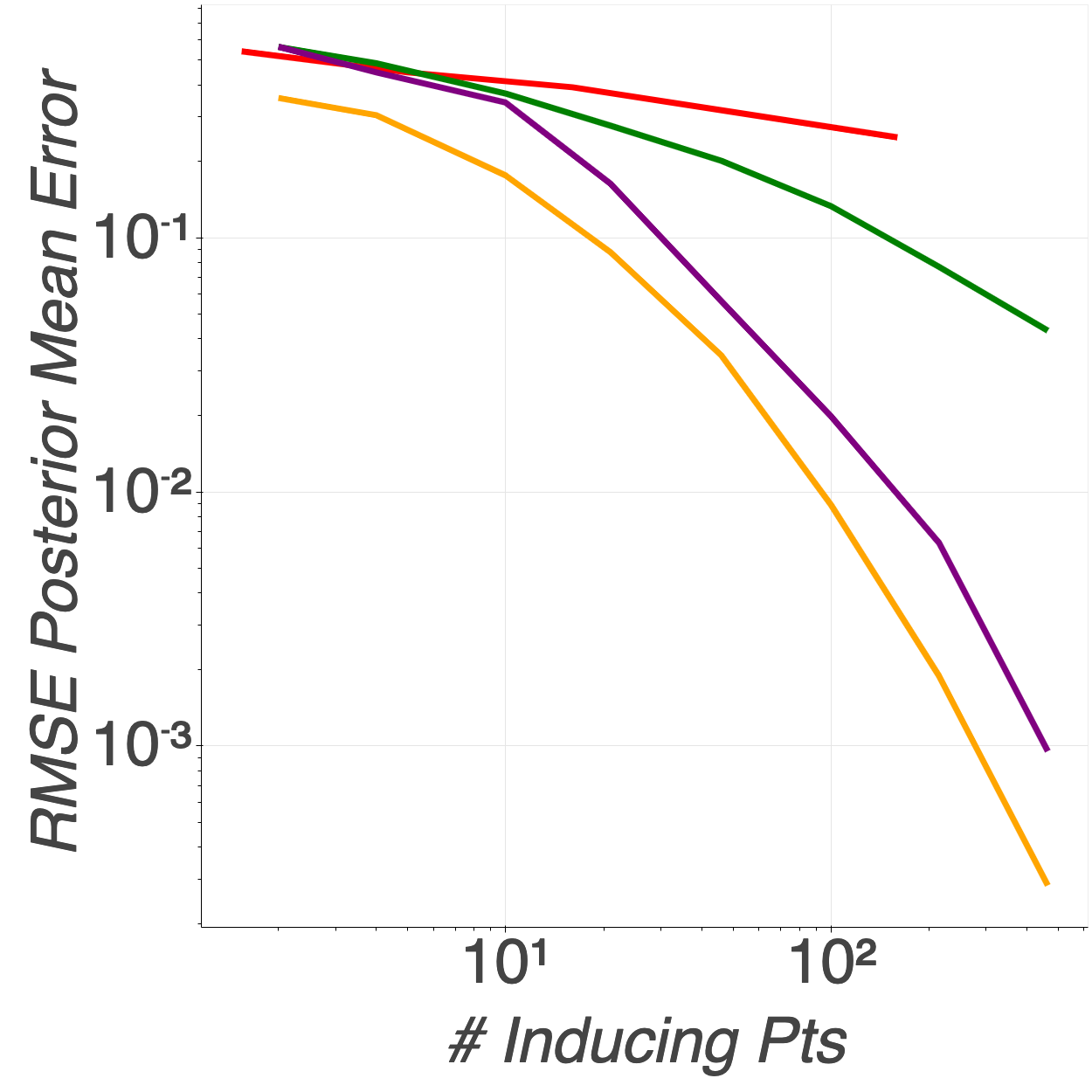} 
    \includegraphics[height=90pt]{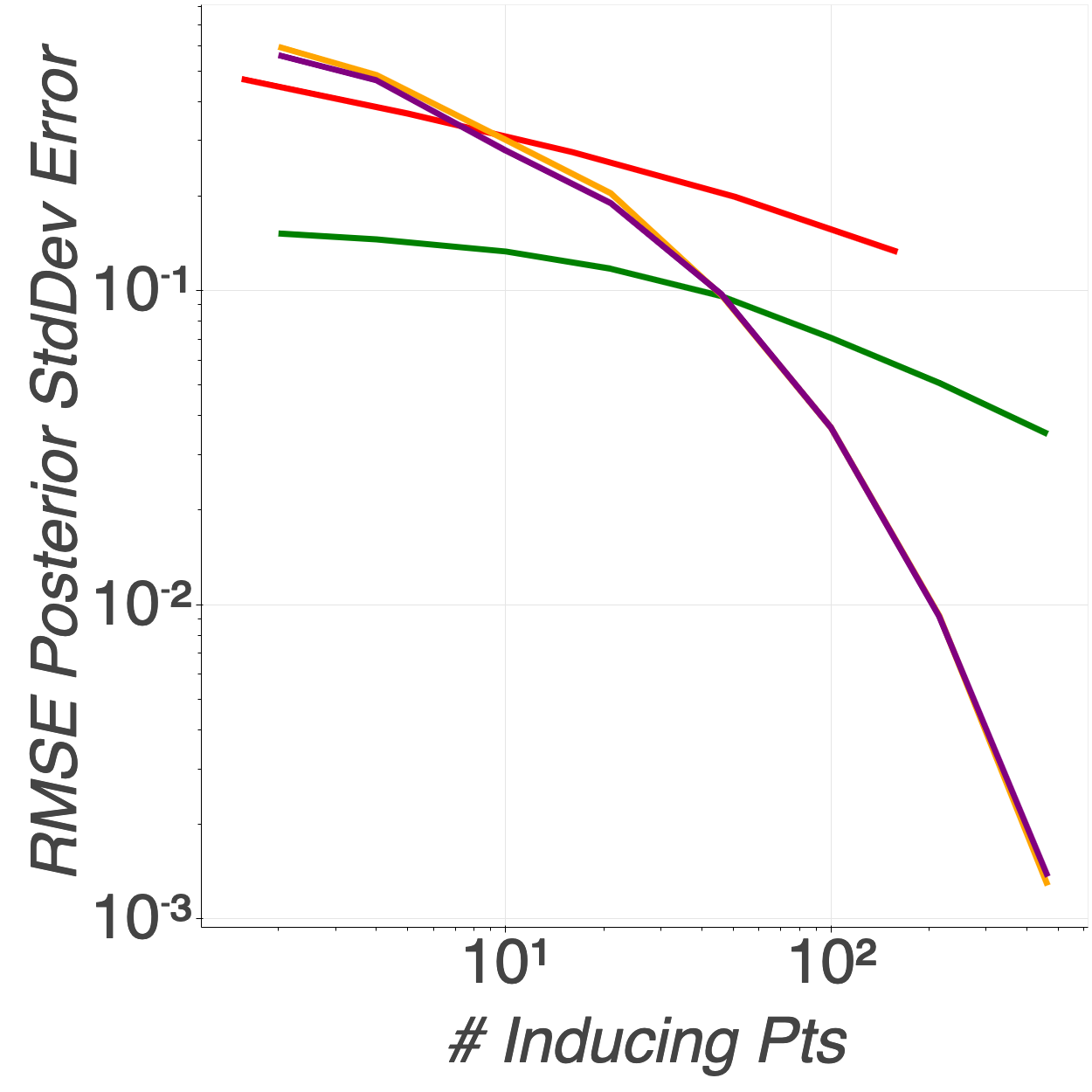}
    \caption{wine}
    \label{fig:wine}
\end{subfigure} 
\end{center}
\caption{Root mean squared error of the approximate posterior mean (left) and standard deviation (right) 
at the held-out test locations.}
\label{fig:mean-and-stdev-error}
\end{figure}%

\subsubsection*{Acknowledgments}
Thanks to Nicolo Fusi for helpful discussions. 
Thanks to Dan Simpson and Arthur Gretton for valuable feedback on an earlier draft of this paper. 
This research was supported in part by an NSF CAREER Award, an ARO
YIP Award, the Office of Naval Research, a Sloan Research Fellowship, IBM, and Amazon. 
M.\ Kasprzak was supported by an EPSRC studentship.

\bibliographystyle{eabbrvnat}
\bibliography{library}

\newpage

\onecolumn
\opt{arxiv}{\section*{Appendix}% !TEX root = scalable-GPs-AISTATS-2019.tex
% !TEX root = scalable-GPs-supplementary-material-NIPS-2018.tex

\appendix

\numberwithin{equation}{section}
\numberwithin{figure}{section}

\section{Experiments} \label{app:experiments}

%\begin{table*}[b]
\begin{center}
\begin{threeparttable}
\caption{Datasets used for experiments. All datasets from the UCI Machine Learning Repository\tnote{a}
except for synthetic and delays10k datasets.  \\ $K =$ number of datapoints used to construct $\auxdist$ (approximately 10\% of $N_{\text{train}}$) }
\label{tbl:datasets}
\begin{tabular*}{0.9\textwidth}{l|c|c|c|ccl|c|c|c|c}
Name & $N_{\text{train}}$  & $N_{\text{test}}$ & $d$ & $K$ && Name & $N_{\text{train}}$  & $N_{\text{test}}$ & $d$ & $K$  \\ 
%\hline
\cline{1-5}\cline{7-11}
synthetic & 1000 & 1000 & 1 & 100 & & abalone & 3177 & 1000 & 8 & 300  \\
%Power plant & & & 1000 && KEGG (directed) & & & 1000\\
delays10k\tnote{b} & 8000 & 2000 & 8 & 800 & & airfoil & 1103 & 400 & 5 & 100  \\
%kin8nm\tnote{c} & 6192 & 2000 &  8 & 600 \\ 
%Protein & & & & & &  Propulsion & & & &  \\
CCPP & 7568 & 2000 & 4 & 700 & & wine quality & 3898 & 1000 & 11 & 300 
\end{tabular*}
%\begin{tabular*}{0.9\textwidth}{l|c|c|c}
%Name & $N$ & $d$ & max $M$  \\ 
%\hline
%Synthetic & 1000 & 1 &  500  \\
%Power plant & & & 1000\\
%Propulsion & & & 1000 \\
%Protein & & & 1000 \\
%%\end{tabular*} \quad 
%%\begin{tabular}{l|c|c|c}
%%Name & $N$ & $d$ & max $M$  \\ 
%%\hline
%Abalone & & & 1000 \\
%KEGG (directed) & & & 1000 \\
%Kin\tnote{b} & & & 1000 \\ 
%Airline delays\tnote{c} & & & 1000 \\
%Pumadyn\tnote{d} &  & & 1000 \\
%\end{tabular*}
\begin{tablenotes}
\item[a] {\footnotesize\url{http://archive.ics.uci.edu/ml/index.php}}
\item[b] \footnotesize{\citet{Hensman:2013}}
%\item[ds] \footnotesize{\url{https://gaussianprocess.org/gpml/data/}}
%\item[c] \footnotesize{\url{https://www.cs.utoronto.ca/~delve/data/kin/desc.html}}
\end{tablenotes}
\end{threeparttable}
\end{center}
%\end{table*}

\begin{figure}[H]
\begin{center}
\begin{subfigure}[b]{.99\textwidth} 
 \centering\includegraphics[trim={0 0 0 0},clip,height=120pt]{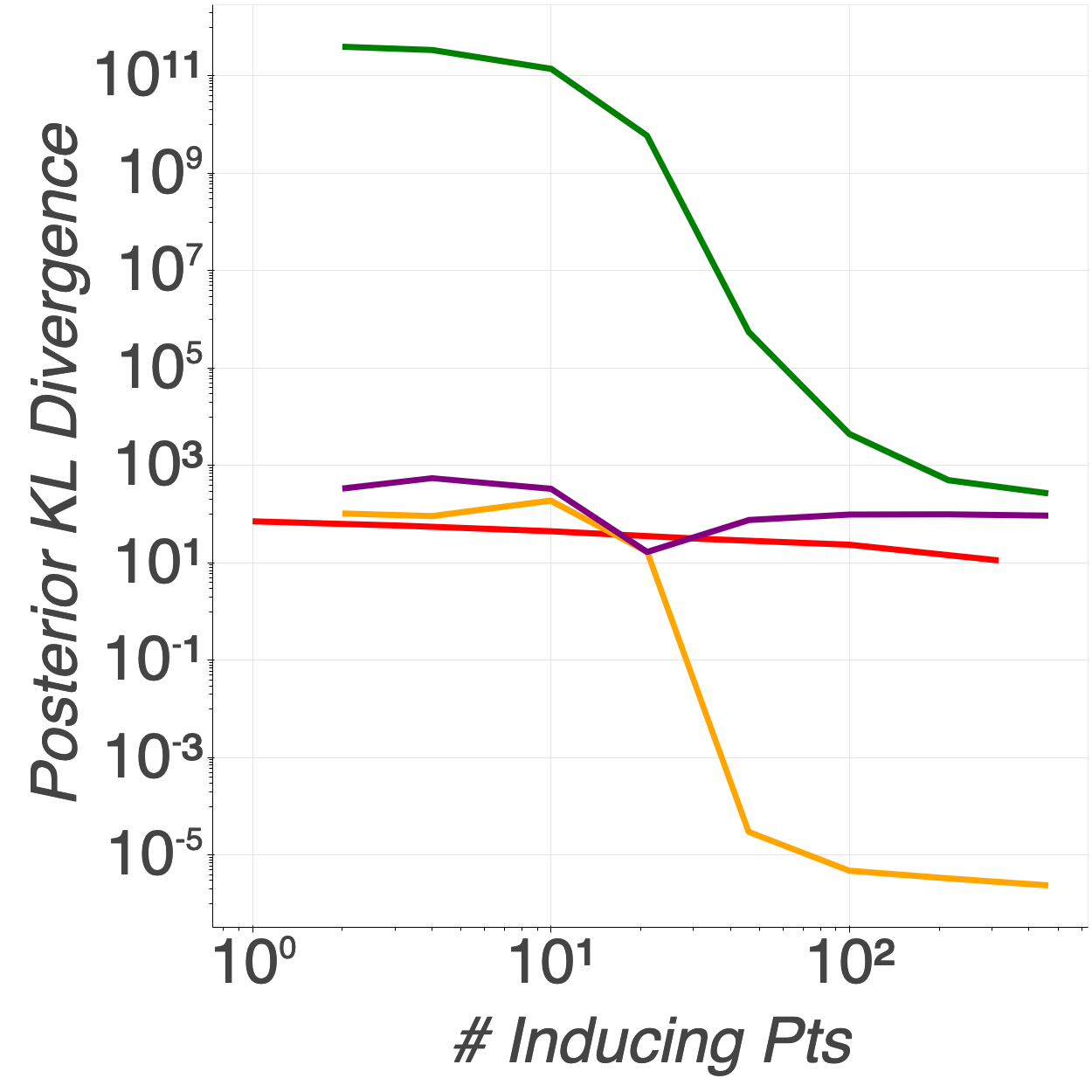}
    \includegraphics[trim={0 0 0 0},clip,height=120pt]{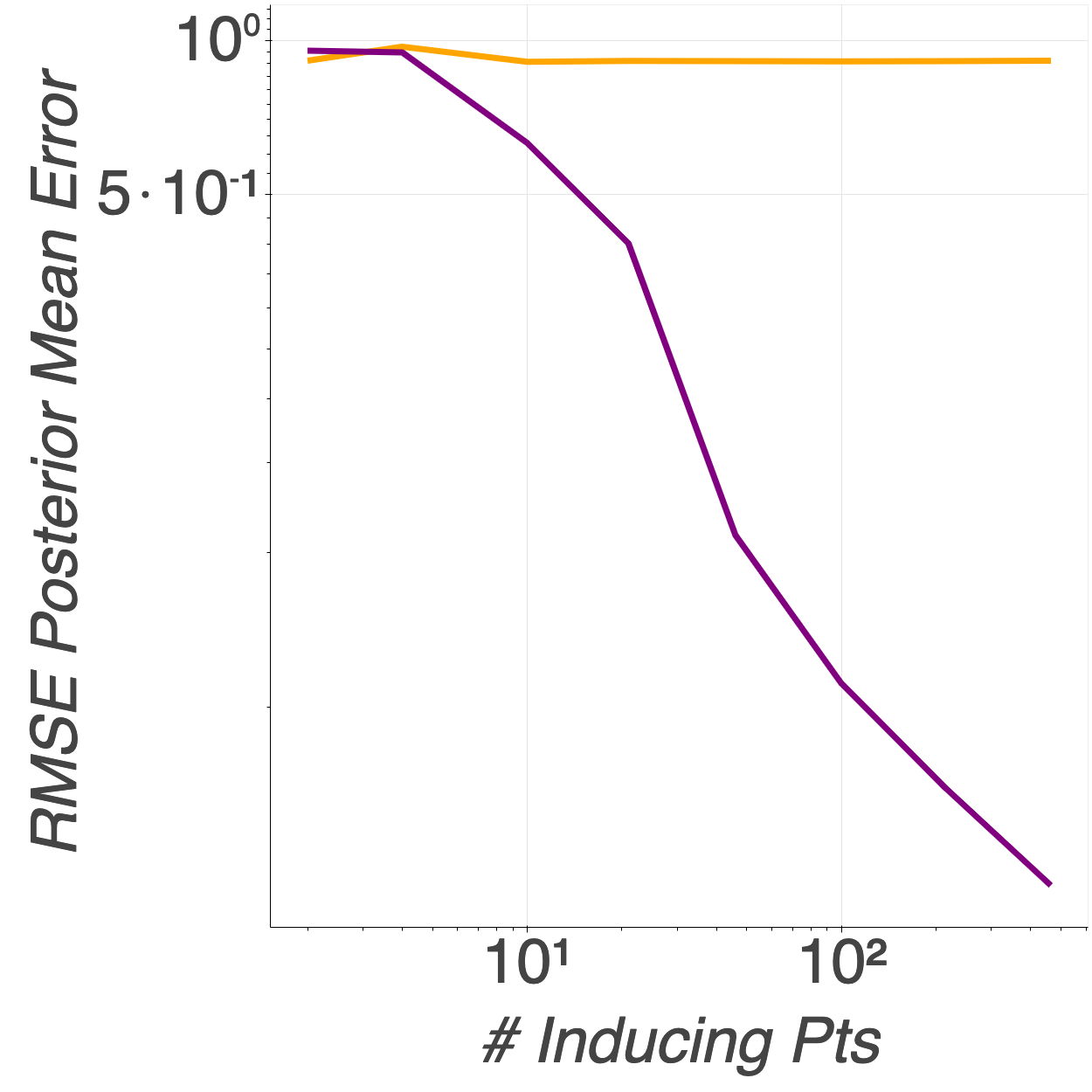}
    \includegraphics[trim={0 0 0 0},clip,height=120pt]{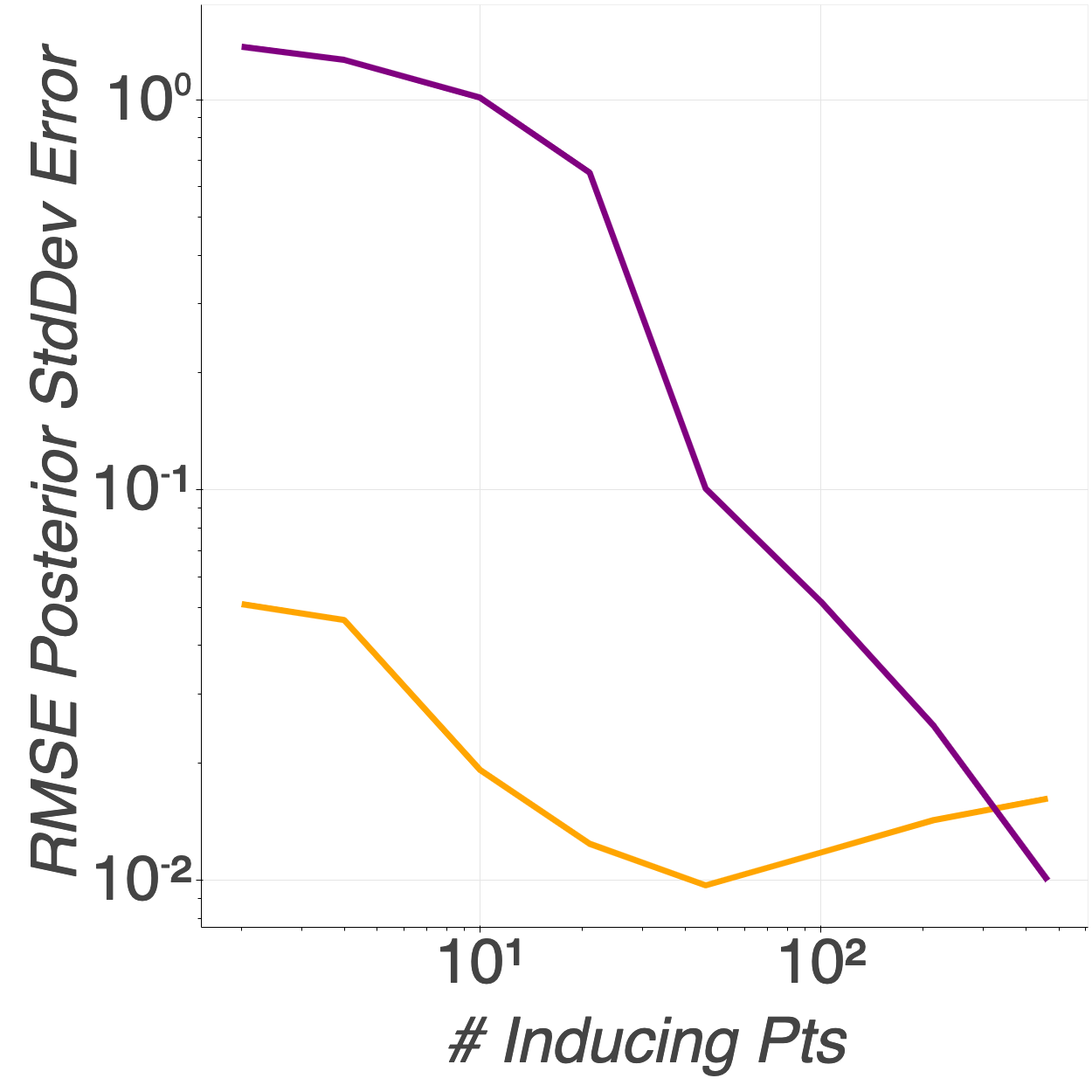}
    \caption{synthetic}
    \label{fig:synthetic-KL-and-mean-best} 
\end{subfigure} \\
\begin{subfigure}[b]{.99\textwidth} 
 \centering\includegraphics[trim={0 0 0 0},clip,height=120pt]{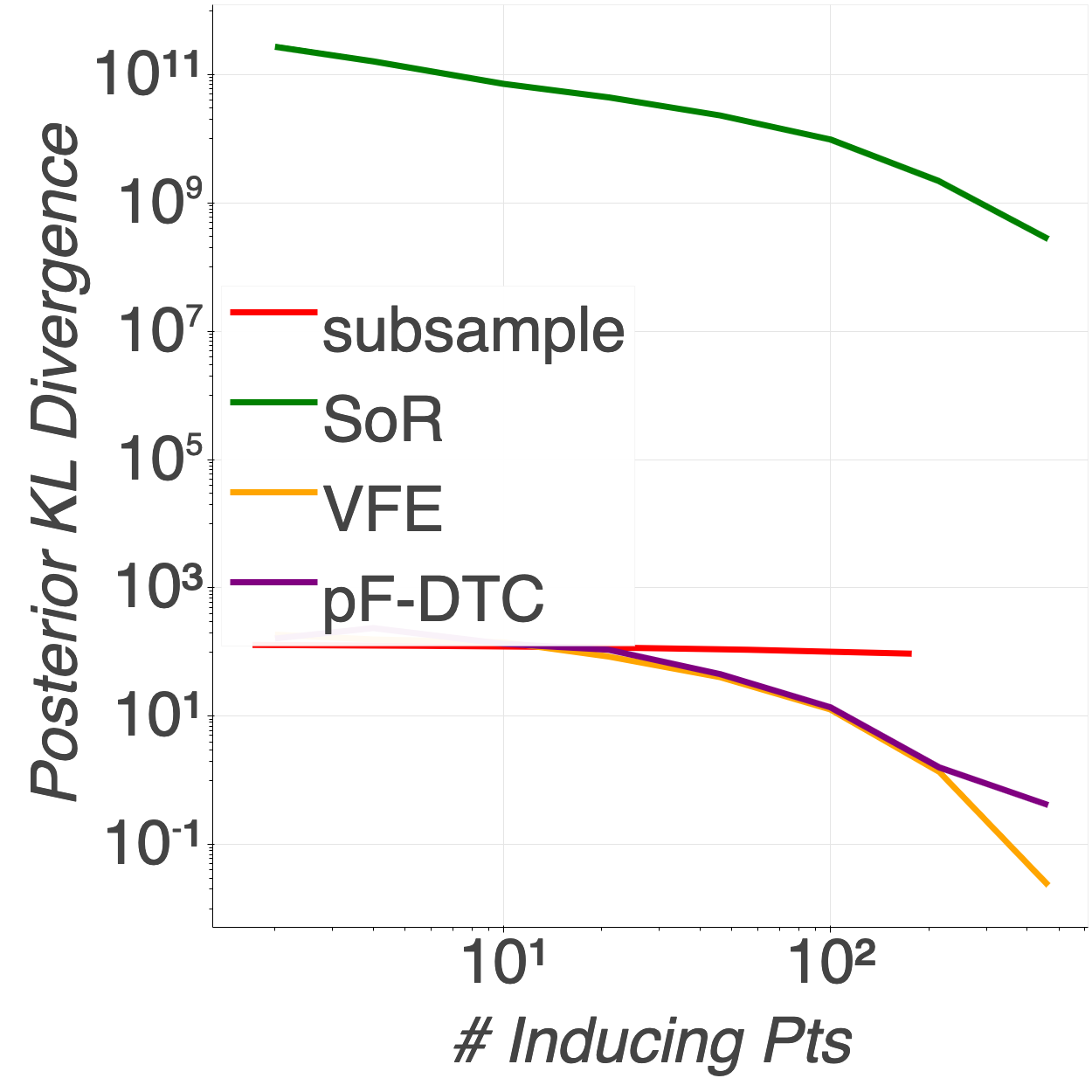}
    \includegraphics[trim={0 0 0 0},clip,height=120pt]{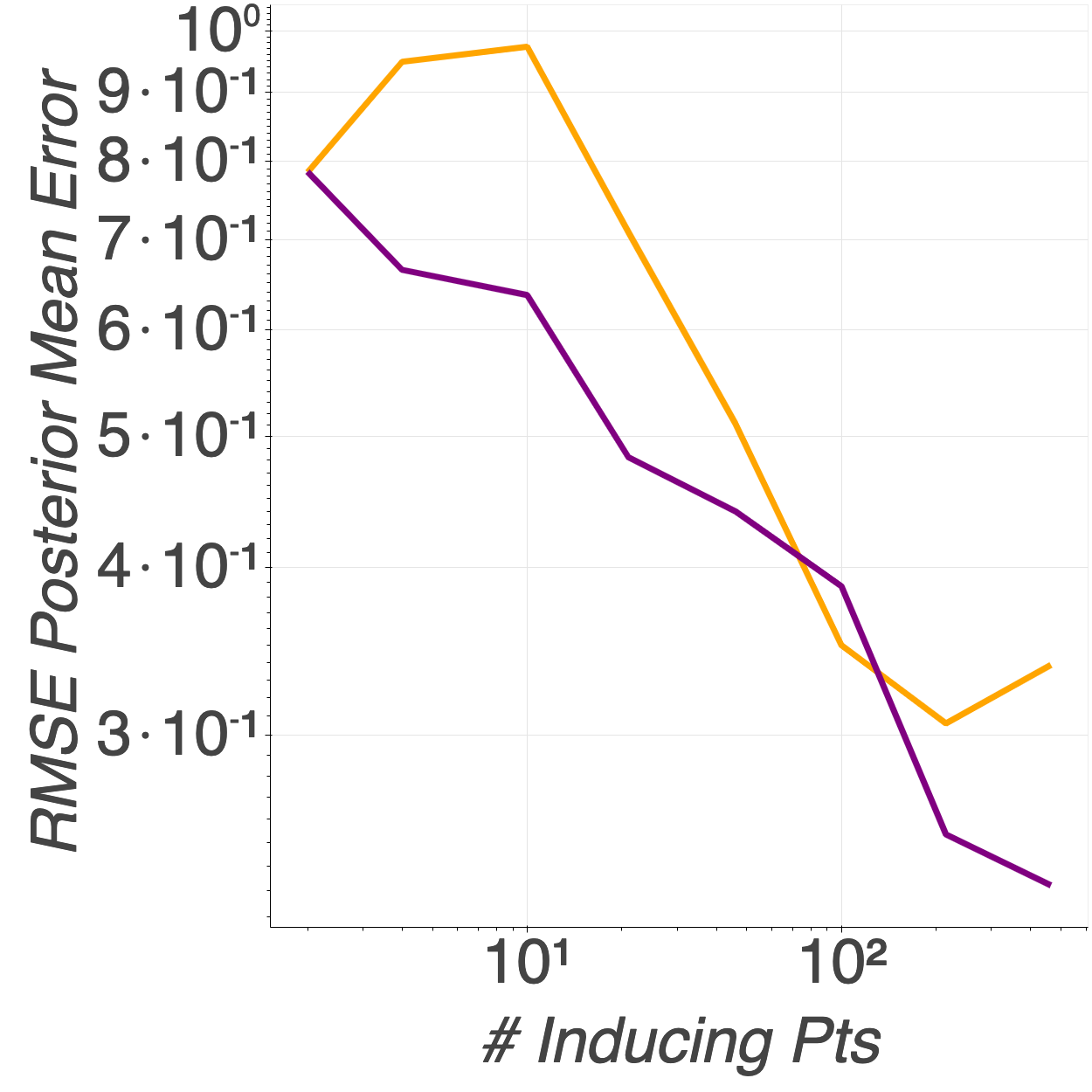}
    \includegraphics[trim={0 0 0 0},clip,height=120pt]{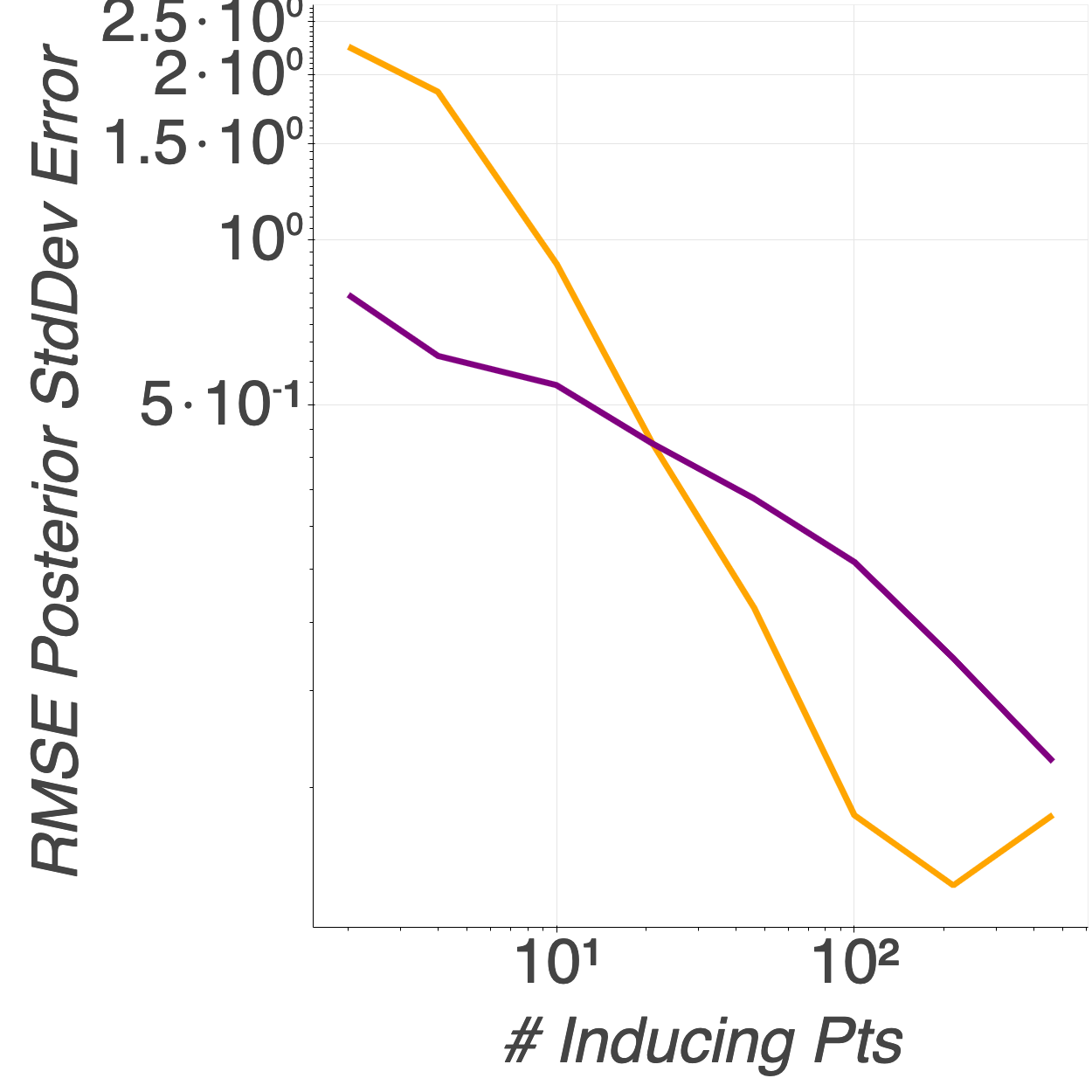}
    \caption{abalone}
    \label{fig:abalone-KL-and-mean-best} 
\end{subfigure} \\
\begin{subfigure}[b]{.99\textwidth} 
   \centering \includegraphics[trim={0 0 0 0},clip,height=120pt]{airfoil-KL-legend} 
     \includegraphics[trim={0 0 0 0},clip,height=120pt]{airfoil-mean-best} 
     \includegraphics[trim={0 0 0 0},clip,height=120pt]{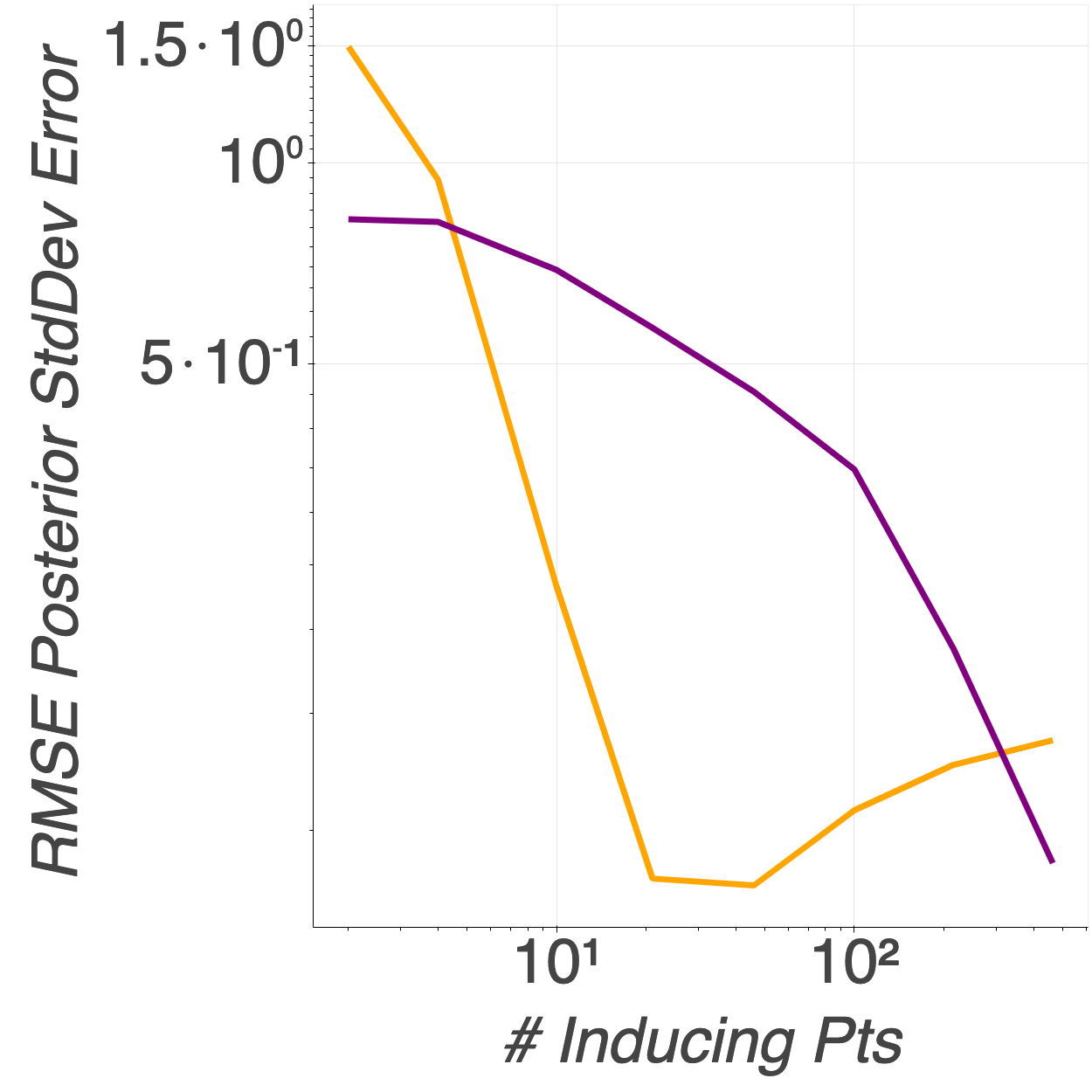} 
    \caption{airfoil}
    \label{fig:airfoil-KL-and-mean-best} 
\end{subfigure}
\end{center}
\caption{KL divergences of the approximate posteriors and root mean squared error of the approximate
posteriors for the VFE and pF-DTC trials with the smallest objective values.}
\label{fig:KL-and-mean-and-stdev-best-part-2}
\end{figure}

\begin{figure}[H]
\begin{center}
\begin{subfigure}[b]{.99\textwidth} 
 \centering\includegraphics[trim={0 0 0 0},clip,height=120pt]{ccpp-KL}
    \includegraphics[trim={0 0 0 0},clip,height=120pt]{ccpp-mean-best}
    \includegraphics[trim={0 0 0 0},clip,height=120pt]{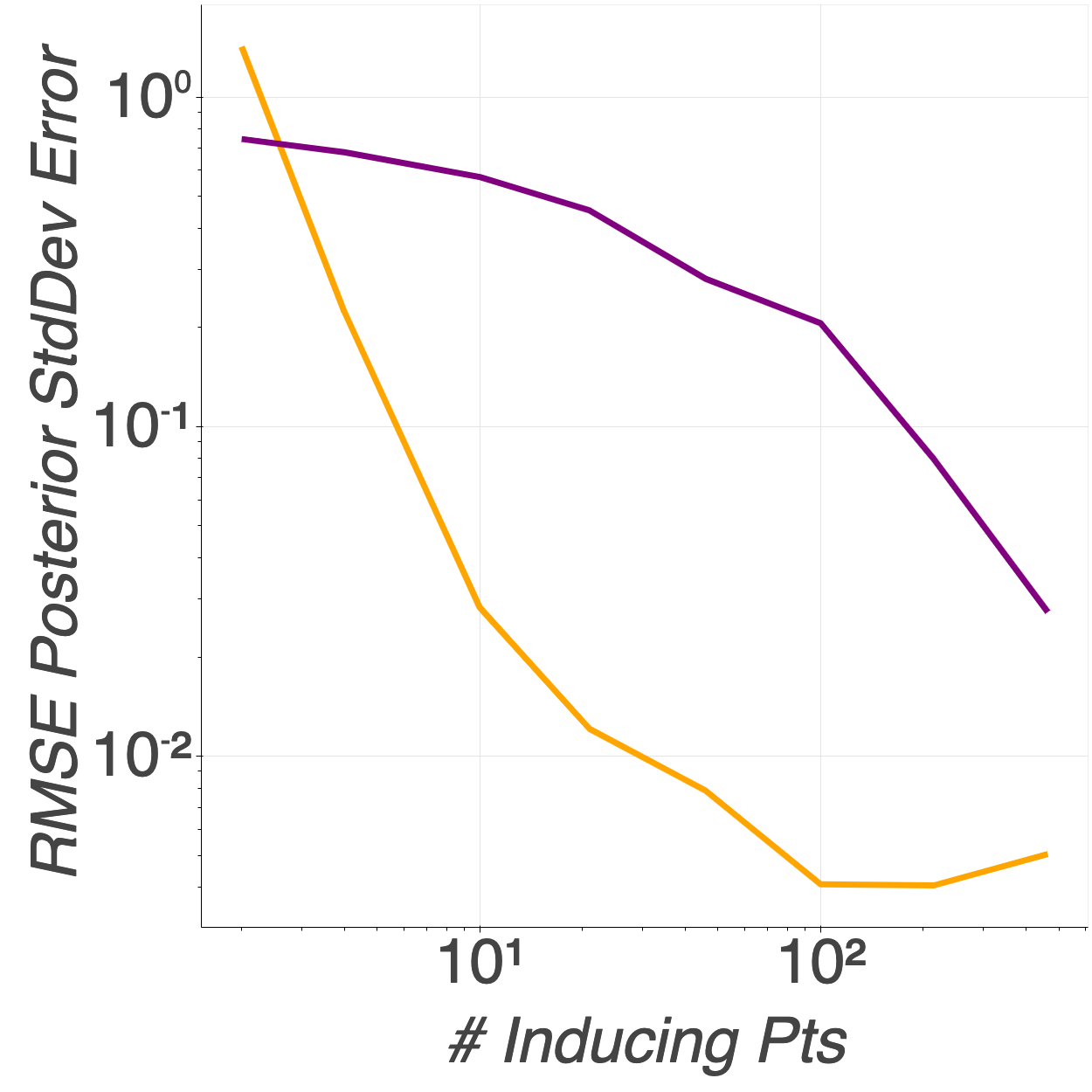}
    \caption{CCPP}
    \label{fig:ccpp-KL-and-mean-best}  
\end{subfigure} \\
\begin{subfigure}[b]{.99\textwidth} 
 \centering\includegraphics[trim={0 0 0 0},clip,height=120pt]{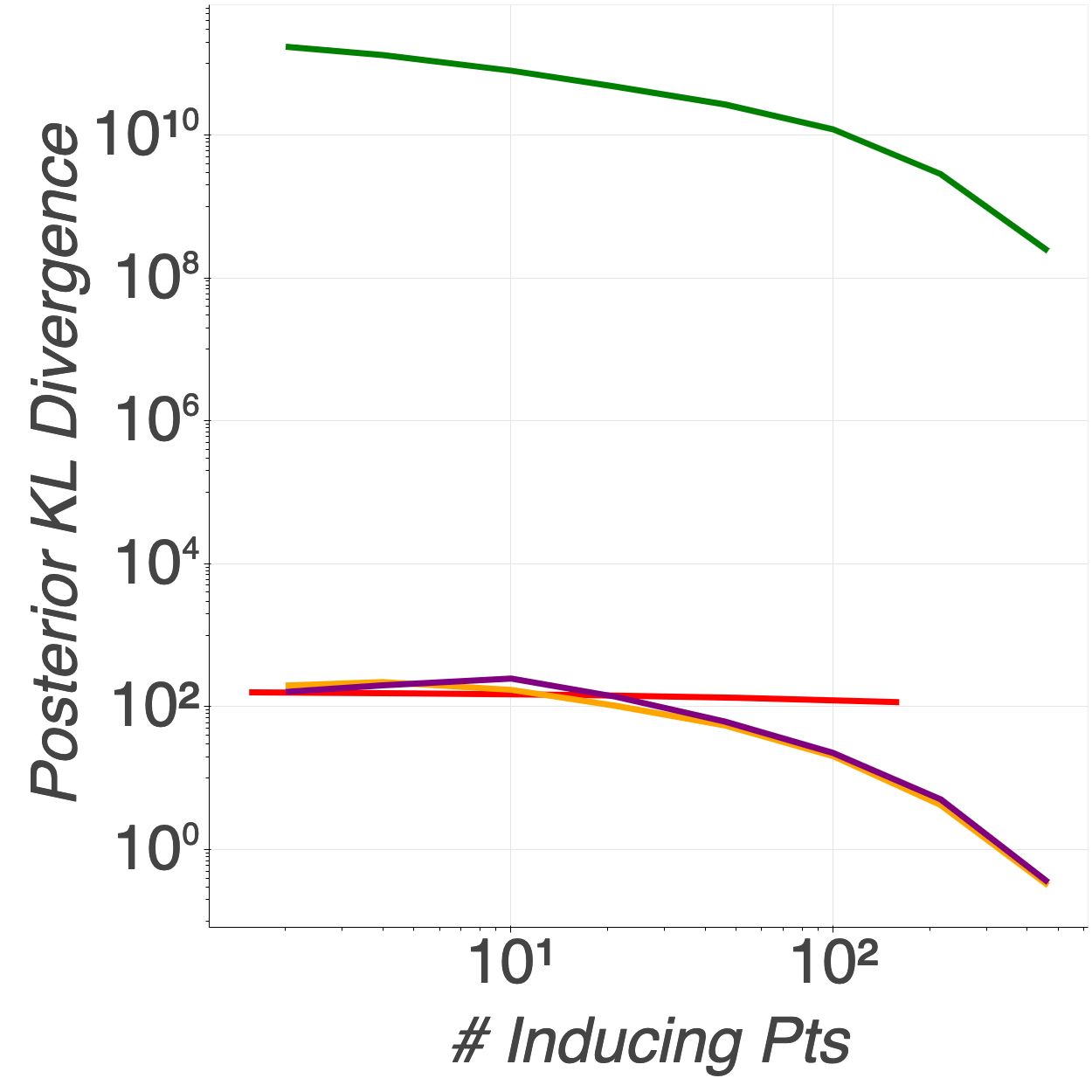}
    \includegraphics[trim={0 0 0 0},clip,height=120pt]{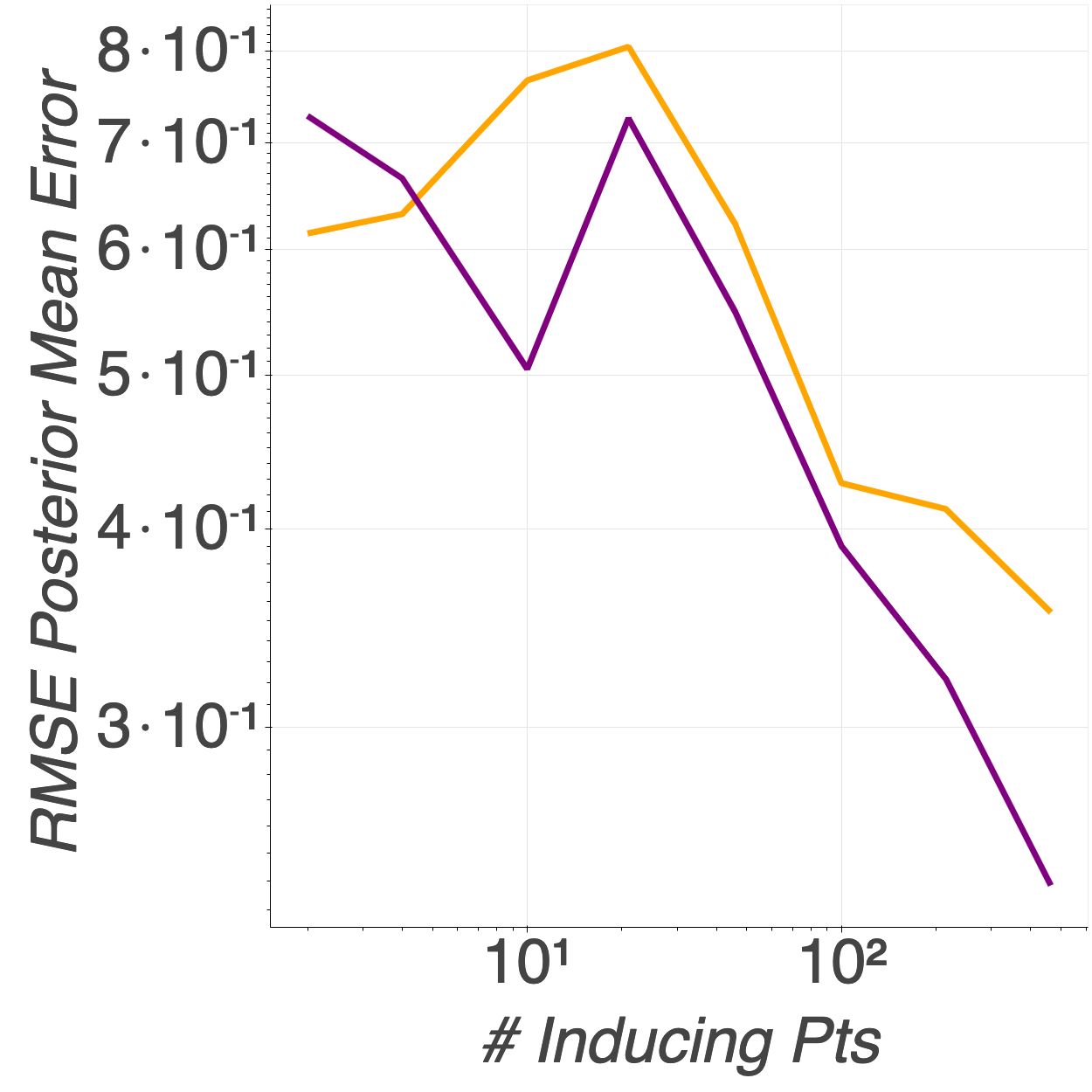}
    \includegraphics[trim={0 0 0 0},clip,height=120pt]{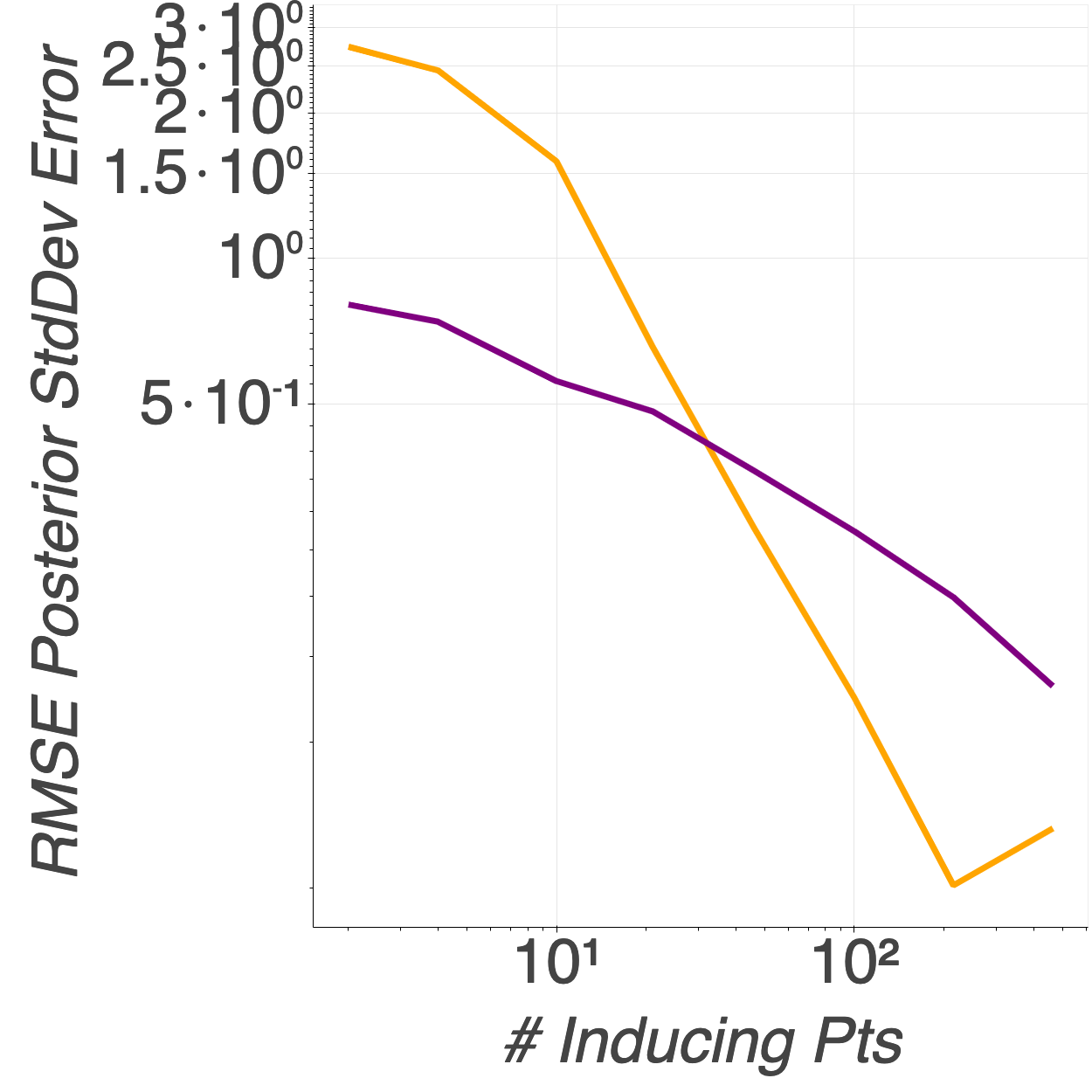}
    \caption{wine}
    \label{fig:wine-KL-and-mean-best} 
\end{subfigure}  \\
\begin{subfigure}[b]{.99\textwidth} 
 \centering\includegraphics[trim={0 0 0 0},clip,height=120pt]{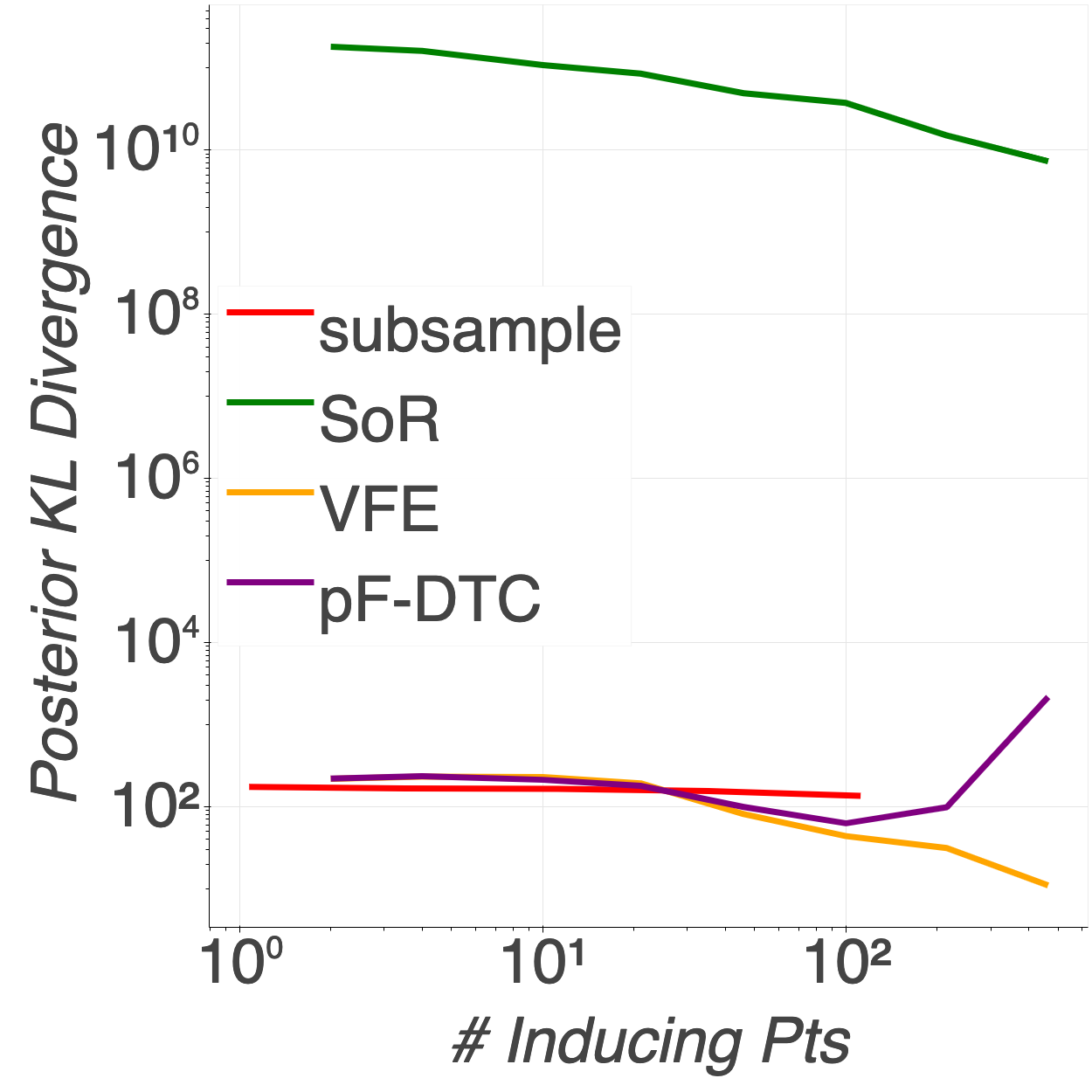}
    \includegraphics[trim={0 0 0 0},clip,height=120pt]{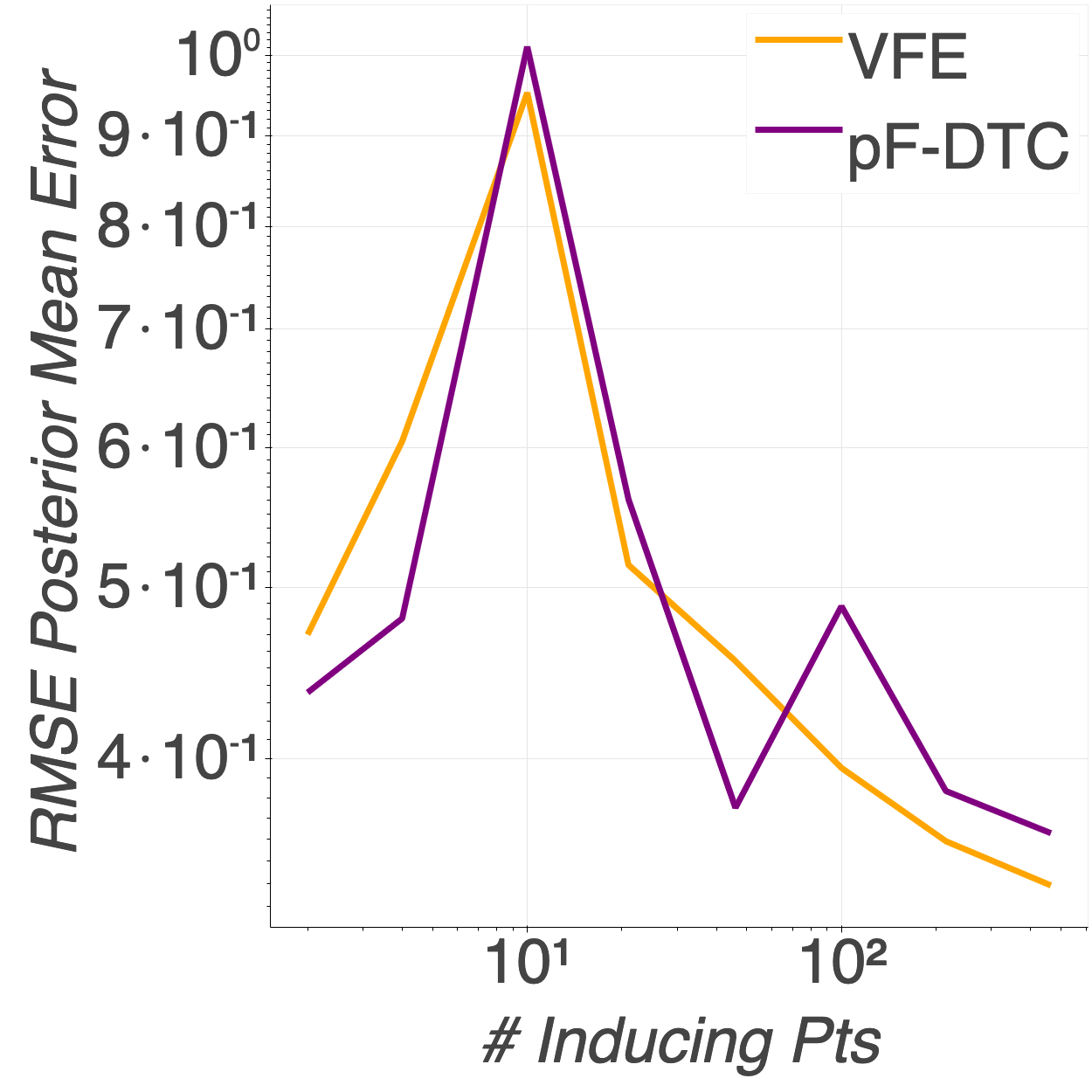}
    \includegraphics[trim={0 0 0 0},clip,height=120pt]{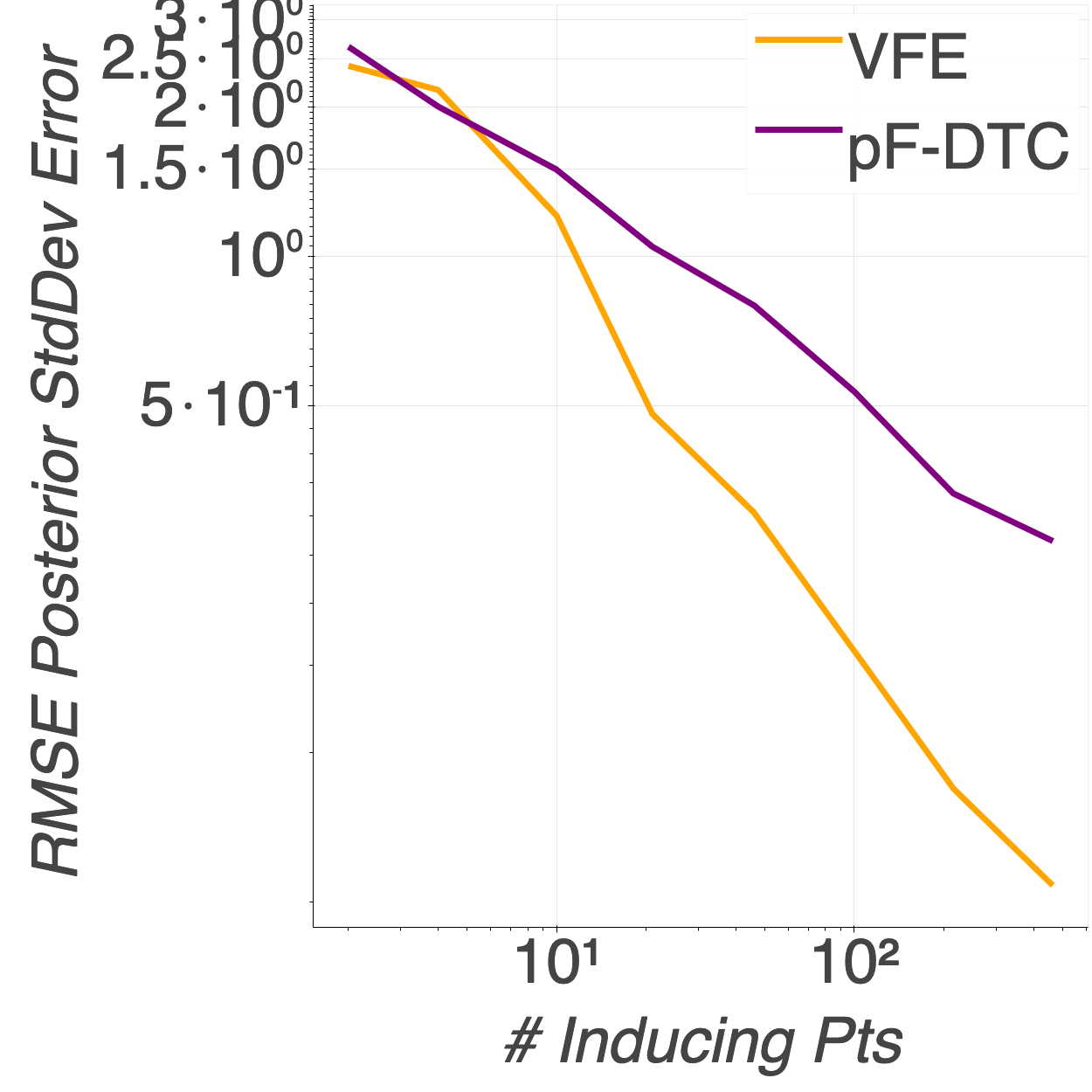}
    \caption{delays10k}
    \label{fig:delays10k-KL-and-mean-best} 
\end{subfigure}
\end{center}
\caption{KL divergences of the approximate posteriors and root mean squared error of the approximate
posteriors for the VFE and pF-DTC trials with the smallest objective values.}
\label{fig:KL-and-mean-and-stdev-best-part-1}
\end{figure}

\section{Proof of \cref{prop:KL-divergence-problems}}
\label{app:KL-prob-proof}

Choose the means and variances of $\mainmeas$ and $\estmeas$ such that $(\estmean - \mainmean)^{2} = \eststd^{2}\{\exp(2\delta) - 1\}$
and $\mainstd^{2} = \exp(2\delta)\eststd^{2}$.
We then have that 
\(
\lefteqn{\kl{\estmeas}{\mainmeas}} \\
&= 0{\cdot}5\{\eststd^{2}/\mainstd^{2} - 1 + \log(\mainstd^{2}/\eststd^{2}) + (\estmean - \mainmean)^{2}/\mainstd^{2}\} \\
&= 0{\cdot}5[\eststd^{2}/\{\exp(2\delta)\eststd^{2}\} - 1 + \log\{\exp(2\delta)\eststd^{2}/\eststd^{2}\} +  \eststd^{2}\{\exp(2\delta) - 1\}/\{\exp(2\delta)\eststd^{2}\}] \\
&= 0{\cdot}5[\exp(-2\delta) - 1 + \log\{\exp(2\delta)\} +  \{\exp(2\delta) - 1\} \exp(-2\delta)] \\
&= \delta.
\)

\section{Details of \cref{exa:fisher-distance-broken}}
\label{app:fisher-distance-broken}

%We take $\GPkernel(x,x') = e^{-(x - x')^{2}/2}$. 
%Therefore, 
We take $\rkhs$ to be the reproducing kernel Hilbert space with reproducing kernel $\RKHSkernel$. 
The posterior covariance functions for  $\mainmeas$ and $\estmeas$ are equal to 
\[
\postkernel(x, x') = e^{-(x-x')^{2}/2} - (1+\sigma^{2})^{-1}e^{-x^{2}/2-(x')^{2}/2} \label{eq:example-post-kernel}
\]
while their posterior means are, respectively, $\mainmean(x) = (1 + \sigma^{2})^{-1}e^{-x^{2}/2}t$ 
and $\estmean(x) = (1 + \sigma^{2})^{-1}e^{-x^{2}/2}\tilde{t}$
Define the induced kernel $\sqGPkernel(x, x') \defined  \inner{\GPkernel_{x}}{\GPkernel_{x'}}$.
Since their covariance operators are equal, the 2-Wasserstein distance between the $\mainmeas$ and $\estmeas$ is~\citep[Thm.~3.5]{Gelbrich:1990}
\[
\pwassSimple{2}{\mainmeas}{\estmeas}
&= \norm{\mainmean - \estmean} 
= \norm{k(0,\cdot)} (1 + \sigma^{2})^{-1}|t - \tilde{t}| \nonumber \\
&= \sqrt{\sqGPkernel(0, 0)}(1 + \sigma^{2})^{-1}|t - \tilde{t}|. \label{eq:example-2-Wasserstein}
\]

%Let $\loglik{n}(f) \defined -\frac{1}{2\sigma^{2}}(f(\xn{n}) - \yn{n})^{2}$ and note
%that
%\[
%\kernelop{\GPkernel}\Fderiv\loglik{n}(f)(\x) 
%&= - \sigma^{-2}(f(\xn{n}) - \yn{n})\rinner{\RKHSkernel(\xn{n}, \cdot)}{\GPkernel(\x, \cdot)} \\
%&= - \sigma^{-2}(f(\xn{n}) - \yn{n})\GPkernel(\x, \xn{n}).
%\)
%\fTBD{JHH: this should eventually be moved into a lemma elsewhere}Letting $\sqGPkernel(\xn{n}, \xn{m}) \defined  \inner{\GPkernel(\xn{n}, \cdot)}{\GPkernel(\xn{m}, \cdot)}$, we have
%\[
%\lefteqn{\EE_{f \dist \auxdist}[\inner{\kernelop{\GPkernel}\Fderiv\loglik{n}(f)}{\kernelop{\GPkernel}\Fderiv\loglik{m}(f)}]} \\
%&= \sigma^{-4} \inner{\GPkernel(\xn{n}, \cdot)}{\GPkernel(\xn{m}, \cdot)} \EE_{f \dist \auxdist}[(f(\xn{n}) - \yn{n})(f(\xn{m}) - \yn{m})] \\
%&= \sigma^{-4} \sqGPkernel(\xn{n}, \xn{m})[\auxkernel(\xn{n}, \xn{m}) + (\yn{n} - \auxmean(\xn{n}))(\yn{m} - \auxmean(\xn{m}))], \label{eq:fisher-inner-prod}
%\)
%Letting $\sqGPkernel(\xn{n}, \xn{m}) \defined  \inner{\GPkernel(\xn{n}, \cdot)}{\GPkernel(\xn{m}, \cdot)}$, we have
%\[
%\lefteqn{\EE_{f \dist \auxdist}[\inner{\kernelop{\GPkernel}\Fderiv\loglik{n}(f)}{\kernelop{\GPkernel}\Fderiv\loglik{m}(f)}]} \\
%&= \sigma^{-4} \inner{\GPkernel(\xn{n}, \cdot)}{\GPkernel(\xn{m}, \cdot)} \EE_{f \dist \auxdist}[(f(\xn{n}) - \yn{n})(f(\xn{m}) - \yn{m})] \\
%&= \sigma^{-4} \sqGPkernel(\xn{n}, \xn{m})[\auxkernel(\xn{n}, \xn{m}) + (\yn{n} - \auxmean(\xn{n}))(\yn{m} - \auxmean(\xn{m}))], \label{eq:fisher-inner-prod}
%\)
The log-likelihoods associated with $\mainmeas$ and $\estmeas$ are, respectively, $\loglik{}(f) \defined -\frac{1}{2\sigma^{2}}(f(0) - t)^{2}$
and $\tilde{\loglik{}}(f) \defined -\frac{1}{2\sigma^{2}}(f(0) - \tilde{t})^{2}$.
Using \cref{lem:loglik-fisher-inner-prod}, in the non-preconditioned case we have 
\[
\fd{\auxdist}{\mainmeas}{\estmeas}^{2}
&= \EE_{f \dist \auxdist}[\inner{\Fderiv\loglik{}}{\Fderiv\loglik{}} + \inner{\Fderiv\tilde{\loglik{}}}{\Fderiv\tilde{\loglik{}}}  - 2\inner{\Fderiv\loglik{}}{\Fderiv\tilde{\loglik{}}}] \nonumber \\
&= \sigma^{-4}\RKHSkernel(0, 0)[(t - \auxmean(0))^{2} + (\tilde{t} - \auxmean(0))^{2} - 2(t - \auxmean(0)) (\tilde{t} - \auxmean(0))] \nonumber \\
&= \sigma^{-4}\RKHSkernel(0, 0)(t - \tilde{t})^{2}. \label{eq:example-Fisher-distance}
\]
\cref{eq:example-2-Wasserstein,eq:example-Fisher-distance} together show that $c = \sqrt{\RKHSkernel(0, 0)/\sqGPkernel(0, 0)}$.

The preconditioned case is almost identical to \cref{eq:example-Fisher-distance}.
Using \cref{lem:loglik-term-derivative,lem:GP-covariance-operator,eq:example-post-kernel}, 
for any $f \in \rkhs$,
\(
\kernelop{\estmeas}\Fderiv\loglik{}(f)
&= - (1+\sigma^{2})^{-1}(f(0) - t)\GPkernel(0, \cdot)
\)
and similarly for $\kernelop{\estmeas}\Fderiv\tilde{\loglik{}}(f)$.
Hence,
\[
\pfd{\nu}{\mainmeas}{\estmeas}
&=  \EE_{f \dist \auxdist}[\inner{\kernelop{\estmeas}\Fderiv\loglik{}}{\kernelop{\estmeas}\Fderiv\loglik{}} + \inner{\kernelop{\estmeas}\Fderiv\tilde{\loglik{}}}{\kernelop{\estmeas}\Fderiv\tilde{\loglik{}}}  - 2\inner{\kernelop{\estmeas}\Fderiv\loglik{}}{\kernelop{\estmeas}\Fderiv\tilde{\loglik{}}}] \nonumber \\
&= (1+\sigma^{2})^{-2}\sqGPkernel(0, 0)[(t - \auxmean(0))^{2} + (\tilde{t} - \auxmean(0))^{2} - 2(t - \auxmean(0)) (\tilde{t} - \auxmean(0))] \nonumber \\
&= (1+\sigma^{2})^{-2}\sqGPkernel(0, 0)(t - \tilde{t})^{2}. \label{eq:example-pFisher-divergence}
\]
\cref{eq:example-2-Wasserstein,eq:example-pFisher-divergence} together show that 
$
\pfd{\auxdist}{\mainmeas}{\estmeas}
%&= (1+\sigma^{2})^{-1}\sqGPkernel(0, 0)^{1/2}(t - \tilde{t})
= \pwassSimple{2}{\mainmeas}{\estmeas}.
$

\section{Proof of \cref{thm:pfd-controls-mean-and-variance}}
\label{app:proof-of-pfd-controls-mean-and-variance}

\cref{thm:pfd-controls-mean-and-variance} will follow almost immediately after we develop a number of preliminary results. 
For more details on infinite-dimensional SDEs and related ideas, we recommend \citet{Hairer:2005,Hairer:2007} 
and \citet{DaPrato:2014}.

The notation in this section differs slightly from the rest of the paper in order to follow the 
conventions of the stochastic processes literature.
Let $\Wiener{}$ denote a $\kernelop{}$-Wiener process~\cite[Definition 4.2]{DaPrato:2014}, 
where $\kernelop{} : \rkhs \to \rkhs$ is the linear, self-adjoint, positive semi-definite, trace-class operator.
%on a Hilbert space $\rkhs$ endowed with inner product $\inner{\cdot}{\cdot}$ and norm $\norm{\cdot}$. 
Let $\mu \in \rkhs$ and let $\drift, \adrift : \rkhs \to \reals$
and consider the following infinite-dimensional stochastic differential equations (SDEs) in $\rkhs$ :
\[
\dXt{t} &= (\GPmean-\Xt{t})\dt+\drift(\Xt{t})\dt+\sqrt{2}\dWt{t} \label{sde1} \\
\dYt{t} &= (\GPmean-\Yt{t})\dt+\adrift(\Yt{t})\dt+\sqrt{2}\dWt{t}. \label{sde2}
\]

We will need the constructions from the following lemma, the proof of which is deferred to \cref{sec:proof-of-joint-sde}.
\bnlem \label{lem:joint-sde}
Let $\tilde{\rkhs} \defined \rkhs \oplus \rkhs$, the direct sum of $\rkhs$ with itself, for which the inner product is given by 
\(
\innerarg{\tilde{\rkhs}}{(x_1,x_2)}{(y_1,y_2)}=\inner{x_1}{y_1}+\inner{x_2}{y_2}.
\)
Define the self-adjoint operator $\tilde{\kernelop{}}: \tilde{\rkhs} \to \tilde{\rkhs}$ given by $(x,y) \mapsto (\kernelop{}(x+y),\kernelop{}(x+y))$.
Then \cref{sde1,sde2} can be written on a common probability space as  
\[
\dee (\Xt{t},\Yt{t})=(\GPmean,\GPmean)\dt-(\Xt{t},\Yt{t})\dt+(\drift(\Xt{t}),\adrift(\Yt{t}))\dt+\sqrt{2}\dee (\Wiener{t},\Wiener{t}) \label{sde_joint}
\]
or
\[
(\Xt{t},\Yt{t}) = \int_0^t(\GPmean,\GPmean)\dee s - \int_0^t(\Xt{s},\Yt{s})\dee s+\int_0^t(\drift(\Xt{s}),\adrift(\Yt{s}))\dee s + \sqrt{2} (\Wiener{t},\Wiener{t}),
\]
where $ t\mapsto (\Xt{t},\Yt{t})$ is a process on $\tilde{\rkhs}$ and $ t\mapsto (\Wiener{t},\Wiener{t})$ 
is a $\tilde{\kernelop{}}$-Wiener process on $\tilde{\rkhs}$.
\enlem

%For any two measures $\eta, \nu \in \probspace$, let $\couplings{\eta}{\nu}$ denote the space of all couplings between $\eta$ and $\nu$ 
%(i.e.\ the laws of random variables $(X,Y)$ such that $X\dist\eta$ and $Y\dist\nu$).
%The 2-Wasserstein distance between $\eta$ and $\nu$ is defined by
%\[
%\pwassSimple{2}{\eta}{\nu}
%\defined \inf_{\gamma\in\couplings{\eta}{\nu}}\left(\int\norm{x-y}^2\gamma(\dee x,\dee y)\right)^{1/2}.
%\)
Let $\probspace$ denote the space of Borel measures on $\rkhs$. 
Recall that for any $\eta \in \probspace$, the $\normarg{\eta}{\cdot}$-norm acting on functions $A:\rkhs\to\rkhs$ is defined by
\(
\normarg{\eta}{A}:=\left(\int\norm{A(x)}^2\eta(\dee x)\right)^{1/2}.
\)

\bnthm\label{thm:2-Wasserstein-bound-simple}
Assume that \cref{sde_joint} has a unique stationary law with the marginal stationary
laws of \cref{sde1,sde2} given by $\estmeas$ and $\mainmeas$ respectively. 
Suppose that for $X \dist \estmeas$ and $Y\dist \mainmeas$, 
$\EE\norm{X}^2<\infty$ and $\EE\norm{Y}^2<\infty$. 
Suppose that for some $\alpha>0$, $\drift$ satisfies the one-sided Lipschitz condition
\[
\inner{\drift(x)-\drift(y)}{x-y}\leq (-\alpha+1)\norm{x-y}^2\text{ for all }x,y\in\rkhs.
\]
Then
\[
\pwassSimple{2}{\mainmeas}{\estmeas} \leq \alpha^{-1}\staticnormarg{\mainmeas}{\drift-\adrift}. \label{eq:2-wasserstein-sde-error}
\]
\enthm
We defer the proof to \cref{app:proof-of-2-Wasserstein-bound-simple}.

\bnprop \label{prop:wasserstein-error}
If the hypotheses of \cref{thm:2-Wasserstein-bound-simple} hold,
then for any distribution $\nu \in \probspace$ such that $\nu \abscont \mainmeas$, 
\[
\pwassSimple{2}{\mainmeas}{\estmeas} 
\le  \alpha^{-1}  \infnorm{\der{\mainmeas}{\nu}}^{1/2}\staticnormarg{\nu}{\drift - \adrift} \label{eq:2-wasserstein-error}
\]
\enprop
\bprf
Using \Holder's inequality, we have
\(
\staticnormarg{\mainmeas}{\drift - \adrift}^{2}
&= \int \staticnorm{\drift(x) - \adrift(x)}^{2} \mainmeas(\dee x) \\
&= \int  \der{\mainmeas}{\nu}(x)  \staticnorm{\drift(x) - \adrift(x)}^{2}\nu(\dee x) \\
&\le  \infnorm{\der{\mainmeas}{\nu}} \int \staticnorm{\drift(x) - \adrift(x)}^{2} \nu(\dee x) .
\)
\cref{eq:2-wasserstein-error} follows by plugging the previous display into \cref{eq:2-wasserstein-sde-error}.
\eprf

\bnprop \label{prop:rkhs-Wasserstein-mean-cov-bounds}
If $\eta, \nu \in \probspace$, $\pwassSimple{2}{\eta}{\nu} \le \veps$ and $\rkhs = \rkhsarg{\RKHSkernel}$, then for all $\x \in \covspace$,
\(
|\mu_{\eta}(\x) - \mu_{\nu}(\x)| &\le \RKHSkernel(\x, \x)^{1/2}\veps \\
|\GPkernel_{\eta}(\x, \x)^{1/2} - \GPkernel_{\nu}(\x, \x)^{1/2}|  
	&\le \sqrt{6}\RKHSkernel(\x, \x)^{1/2}\veps \\
|\GPkernel_{\eta}(\x, \x) - \GPkernel_{\nu}(\x, \x)|    
	&\le 3\,\RKHSkernel(\x, \x)^{1/2}\min(\GPkernel_{\eta}(\x, \x), \GPkernel_{\nu}(\x, \x))^{1/2}\veps  + 6\,\RKHSkernel(\x, \x)\veps^{2}.
\)      
\enprop
We defer the proof to \cref{app:proof-of-rkhs-Wasserstein-mean-cov-bounds}.

The result will follow by taking $\kernelop{} = \kernelop{\estmeas}$.
With this choice of $\kernelop{}$, $\drift = 0$ and $\mu = \mu_{\estmeas}$, so $\drift$ satisfies the one-sided Lipschitz condition with $\alpha=1$. 
The remaining hypotheses of \cref{thm:2-Wasserstein-bound-simple} hold by construction, 
so \cref{thm:pfd-controls-mean-and-variance} follows by applying \cref{prop:wasserstein-error,prop:rkhs-Wasserstein-mean-cov-bounds}.

\subsection{Proof of \cref{lem:joint-sde}} 
\label{sec:proof-of-joint-sde}

We first check that the process $t\mapsto(\Wiener{t},\Wiener{t})$ satisfies the definition of a Wiener process. 
It starts from $0$, has continuous trajectories and independent increments. Furthermore,
\(
\mathcal{L}\left((\Wiener{t},\Wiener{t})-(\Wiener{s},\Wiener{s})\right)=\mathcal{N}(0,(t-s)\tilde{\kernelop{}}).
\)
To see that for $t\geq s$ the variance of $(W_t,W_t)-(W_s,W_s)$ in $\tilde{\rkhs}$ is indeed equal to $(t-s)\tilde{\kernelop{}}$, note that, for any $(x_1,x_2),(y_1,y_2)\in\tilde{\rkhs}$
\(
\lefteqn{\EE\left[\innerarg{\tilde{\rkhs}}{(x_1,x_2)}{(\Wiener{t},\Wiener{t})-(\Wiener{s},\Wiener{s})}\innerarg{\tilde{\rkhs}}{(y_1,y_2)}{(\Wiener{t},\Wiener{t})-(\Wiener{s},\Wiener{s})}\right]} \\
&=\EE\left[\left(\inner{x_1}{\Wiener{t}-\Wiener{s}}+\inner{x_2}{\Wiener{t}-\Wiener{s}}\right)\left(\inner{y_1}{\Wiener{t}-\Wiener{s}}+\inner{y_2}{\Wiener{t}-\Wiener{s}}\right)\right]\\
&=\inner{(t-s)\kernelop{}x_1}{y_1}+\inner{(t-s)\kernelop{}x_1}{y_2}+\inner{(t-s)\kernelop{}x_2}{y_1}+\inner{(t-s)\kernelop{}x_2}{y_2}\\
&=\inner{(t-s)\kernelop{}(x_1+x_2)}{y_1}+\inner{(t-s)\kernelop{}(x_1+x_2)}{y_2}\\
&=\innerarg{\tilde{\rkhs}}{(t-s)\tilde{\kernelop{}}(x_1,x_2)}{(y_1,y_2)}.
\)
Given that $\kernelop{}$ is self-adjoint, it follows that $\tilde{\kernelop{}}$ is self-adjoint as well:
\(
\innerarg{\tilde{\rkhs}}{\tilde{\kernelop{}}(x_1,x_2)}{(y_1,y_2)}
&=\inner{\kernelop{}(x_1+x_2)}{y_1+y_2}\\
&=\inner{x_1+x_2}{\kernelop{}(y_1+y_2)}\\
&=\innerarg{\tilde{\rkhs}}{(x_1,x_2)}{\tilde{\kernelop{}}(y_1,y_2)}.
\)

\subsection{Proof of \cref{thm:2-Wasserstein-bound-simple}}
\label{app:proof-of-2-Wasserstein-bound-simple}

We begin by quoting the \Ito formula we will be using (see \citet{DaPrato:2014} for complete details):
\bnthm[\Ito formula, {\citet[Theorem 4.32]{DaPrato:2014}}]\label{ito}
Let $H$ and $U$ be two Hilbert spaces and $\Wiener{}$ be a $Q$-Wiener process for a symmetric non-negative operator $Q\in L(U)$. Let $U_0=Q^{1/2}(U)$ and let $L_2(U_0,H)$ be the space of all Hilbert-Schmidt operators from $U_0$ to $H$. Assume that $\Phi$ is an $L_2(U_0,H)$-valued process stochastically integrable in $[0,T]$, $\varphi$ is an $H$-valued predictable process Bochner integrable on $[0,T]$ almost surely, and $X(0)$ a $H$-valued random variable. Then the following process:
\(
\Xt{t}=\Xt{0}+\int_0^t\varphi(s)\dee s+\int_0^t\Phi(s)\dWt{s},\quad t\in[0,T]
\)
is well defined. Assume that a function $F:[0,T]\times H\to\reals$ and its partial derivatives $F_t,F_x,F_{xx}$ are uniformly continuous on bounded subsets of $[0,T]\times H$. Under these conditions, almost surely, for all $t\in[0,T]$:
\(
F(t,\Xt{t}) & =F(0,\Xt{0})+\int_0^t\left<F_x(s,\Xt{s}),\Phi(s)\dWt{t}\right>+\int_0^tF_t(s,\Xt{s})\dee s\\
&\phantom{=~}+\int_0^t\left<F_x(s,\Xt{s}),\varphi(s)\right>\dee s\\
&\phantom{=~}+\int_0^t\frac{1}{2}\tr\left[F_{xx}(s,\Xt{s})(\Phi(s)Q^{1/2})(\Phi(s)Q^{1/2})^*\right]\dee s.
\)
\enthm

Let $F:[0,\infty)\times\tilde{\rkhs}\to\reals$ be given by $F(t;x,y)=e^{2\alpha t}\norm{x-y}^2$. Then the \Frechet derivative of $F$ with respect to the space parameters is given by
\begin{equation}\label{frec1}
F_{(x,y)}(t;x,y)[(h_1,h_2)]=2e^{2\alpha t}\inner{x-y}{h_1-h_2}.
\end{equation}
\cref{frec1} holds because
\(
\lefteqn{\frac{\left|\norm{x+h_1-y-h_2}^2-\norm{x-y}^2-2\inner{x-y}{h_1-h_2}\right|}{\sqrt{\norm{h_1}^2+\norm{h_2}^2}}} \\
&=\frac{\norm{h_1-h_2}^2}{\sqrt{\norm{h_1}^2+\norm{h_2}^2}}\\
&\leq 2\sqrt{\norm{h_1}^2+\norm{h_2}^2}\xrightarrow{\norm{h_1},\norm{h_2}\to 0} 0.
\)
Furthermore, the second \Frechet derivative with respect to the space parameters is
\begin{equation*}
F_{(x,y),(x,y)}[(h_1,h_2),(h_3,h_4)]=2e^{2\alpha t}\inner{h_3-h_4}{h_1-h_2}.
\end{equation*}

Note that $\tilde{\kernelop{}}^{1/2}(x,y)=\frac{\sqrt{2}}{2} \left(\kernelop{}^{1/2}(x+y),\kernelop{}^{1/2}(x+y)\right)$.
Using the one-sided Lipschitz condition and the Cauchy-Schwarz inequality, we obtain
\begin{align}
\lefteqn{\inner{\drift(\Xt{t})-\adrift(\Yt{t})}{\Xt{t}-\Yt{t}}} \nonumber\\
&=\inner{\drift(\Xt{t})-\drift(\Yt{t})}{\Xt{t}-\Yt{t}}+\inner{\drift(\Yt{t})-\adrift(\Yt{t})}{\Xt{t}-\Yt{t}}\nonumber\\
&\leq (-\alpha+1)\norm{\Xt{t}-\Yt{t}}^2+\staticnorm{\drift(\Yt{t})-\adrift(\Yt{t})}\norm{\Xt{t}-\Yt{t}}.\label{25}
\end{align}

We will assume that we start the process $t \mapsto (\Xt{t},\Yt{t})$ at joint stationarity (with $\Xt{0}\dist \eta$ and $\Yt{0}\dist \nu$).
By the \Ito formula given by Theorem \ref{ito}, applied to the process described by \cref{sde_joint} 
and function $F$ (so that $\varphi(t)=(\drift(\Xt{t}),\adrift(\Yt{t}))-(\Xt{t},\Yt{t})$ in Theorem \ref{ito}):
\begin{align*}
e^{2\alpha t}\norm{\Xt{t}-\Yt{t}}^2
&=\norm{\Xt{0}-\Yt{0}}^2+\int_0^t2\sqrt{2}e^{2\alpha s}\inner{X_s-Y_s}{\dWt{s}-\dWt{s}}\\
&\phantom{=~}+\int_0^t2\alpha e^{2\alpha s}\norm{\Xt{s}-\Yt{s}}^2\dee s\nonumber\\
&\phantom{=~}+\int_0^t 2e^{2\alpha s}\inner{\Xt{s}-\Yt{s}}{\drift(\Xt{s})-\Xt{s}-\adrift(\Yt{s})+\Yt{s}}\dee s\nonumber\\
&\phantom{=~}+\int_0^te^{2\alpha s}\tr\left[(x,y)\mapsto\left(\kernelop{}(x+y)-\kernelop{}(x+y),\kernelop{}(x+y)-\kernelop{}(x+y)\right)\right]\dee s\nonumber\\
&=\norm{\Xt{0}-\Yt{0}}^2+\int_0^t2\alpha e^{2\alpha s}\norm{\Xt{s}-\Yt{s}}^2\dee s\nonumber\\
&\phantom{=~}+\int_0^t 2e^{2\alpha s}\inner{\Xt{s}-\Yt{s}}{\drift(\Xt{s})-\Xt{s}-\adrift(\Yt{s})+\Yt{s}}\dee s\nonumber.
\end{align*}

Taking expectations on both sides (with respect to everything that is random and at the fixed time $t$), multiplying by $e^{-2\alpha t}$ and applying \cref{25}

\begin{align}
\lefteqn{\EE\norm{\Xt{t}-\Yt{t}}^2} \nonumber\\
&\leq e^{-2\alpha t}\EE\norm{\Xt{0}-\Yt{0}}^2+\EE\left[\int_0^t2e^{2\alpha (s-t)}\staticnorm{\drift(\Yt{s})-\adrift(\Yt{s})}\norm{\Xt{s}-\Yt{s}}\dee s\right] \nonumber\\
&\leq e^{-2\alpha t}\EE\norm{\Xt{0}-\Yt{0}}^2\nonumber\\
&\phantom{=~}+\left(\int_0^t 2e^{2\alpha (s-t)}\EE\staticnorm{\drift(\Yt{s})-\adrift(\Yt{s})}^2\dee s\right)^{1/2}\left(\int_0^t 2e^{2\alpha (s-t)}\EE\norm{\Xt{s}-\Yt{s}}^2\dee s\right)^{1/2}\label{14}\\
&= e^{-2\alpha t}\EE\norm{\Xt{0}-\Yt{0}}^2\nonumber\\
&\phantom{=~} +\left(\alpha^{-1/2}(1-e^{-2\alpha t})^{1/2}\staticnormarg{\nu}{\drift-\adrift}\right)\left(\alpha^{-1/2}(1-e^{-2\alpha t})^{1/2}\left(\EE\norm{\Xt{t}-\Yt{t}}^2\right)^{1/2}\right)\label{2.11}\\
&= e^{-2\alpha t}\EE\norm{\Xt{0}-\Yt{0}}^2+\alpha^{-1}(1-e^{-2\alpha t})\staticnormarg{\nu}{\drift-\adrift}\left(\EE\norm{\Xt{t}-\Yt{t}}^2\right)^{1/2}\nonumber,
\end{align}
where \cref{14} follows by the Cauchy-Schwarz inequality and \cref{2.11} follows from the assumption that 
we start the process $t \mapsto (\Xt{t},\Yt{t})$ at joint stationarity.

Now, dividing by $\left(\EE\|\Xt{t}-\Yt{t}\|^2\right)^{1/2}$, taking $t\to\infty$ and 
noting that the process $t\mapsto (\Xt{t},\Yt{t})$ remains at joint stationarity, we obtain the result.
%\[
%\pwassSimple{2}{\eta}{\nu} \leq \alpha^{-1}\staticnormarg{\eta}{\drift-\adrift}.
%\)

\subsection{Proof of \cref{prop:rkhs-Wasserstein-mean-cov-bounds}}
\label{app:proof-of-rkhs-Wasserstein-mean-cov-bounds}

Let $f \dist \eta$ and $g \dist \nu$ and define $\bar\GPkernel_{\nu}(\x, \x') \defined \EE[g(\x)g(\x')]$. 
By Cauchy-Schwarz and Jensen's inequalities,
\(
|\mu_{\eta}(\x) - \mu_{\nu}(\x)|
&= |\EE[f(\x) - g(\x)]| 
= |\EE[\inner{f - g}{\RKHSkernel_{\x}}]| \\
&\le \EE[\norm{f - g} \norm{\RKHSkernel_{\x}}] 
\le \RKHSkernel(\x, \x)^{1/2}\EE[\norm{f - g}^{2}]^{1/2} \\
&\le \RKHSkernel(\x, \x)^{1/2}  \veps. 
\)

Without loss of generality we can assume $\mu_{\eta} = 0$, since if not
then we consider the random variables $\tf \defined f - \mu_{\eta}$ and $\tg \defined g - \mu_{\eta}$ instead. 
It follows from the Cauchy-Schwarz inequality that 
\(
|\GPkernel_{\eta}(\x, \x) - \bar\GPkernel_{\nu}(\x, \x)|
&= |\EE[f(\x)^{2} - g(\x)^{2}]| \\
&= \EE[(f(\x) - g(\x))(f(\x) + g(\x))] \\
&\le \sqrt{\EE[(f(\x) - g(\x))^{2}]} \sqrt{\EE[(f(\x) + g(\x))^{2}]} \\
&\le  \RKHSkernel(\x, \x)^{1/2} \veps \sqrt{2\EE[f(\x)^{2} + g(\x)^{2}]} \\
&\le \sqrt{2}\,\RKHSkernel(\x, \x)^{1/2}\veps (\GPkernel_{\eta}(\x, \x)^{1/2} + \bar\GPkernel_{\nu}(\x, \x)^{1/2}) \\
|\GPkernel_{\eta}(\x, \x)^{1/2} - \bar\GPkernel_{\nu}(\x, \x)^{1/2}|
&\le \sqrt{2}\,\RKHSkernel(\x, \x)^{1/2}\veps.
\)
Also, 
\(
\bar\GPkernel_{\nu}(\x, \x)^{1/2} 
&\le \sqrt{\GPkernel_{\nu}(\x, \x) + \mu_{\nu}(\x)^{2}} 
\le \GPkernel_{\nu}(\x, \x)^{1/2} + \RKHSkernel(\x, \x)^{1/2}  \veps.
\)
We now have that 
\(
|\GPkernel_{\eta}(\x, \x) - \GPkernel_{\nu}(\x, \x)|
&= |\GPkernel_{\eta}(\x, \x) - \bar\GPkernel_{\nu}(\x, \x) + \mu_{\nu}(\x)^{2}| \\
&\le |\GPkernel_{\eta}(\x, \x) - \bar\GPkernel_{\nu}(\x, \x)| + \mu_{\nu}(\x)^{2} \\
&\le \sqrt{2}\,\RKHSkernel(\x, \x)^{1/2}\veps (\GPkernel_{\eta}(\x, \x)^{1/2} + \bar\GPkernel_{\nu}(\x, \x)^{1/2})  + \RKHSkernel(\x, \x)\veps^{2} \\
&\le \sqrt{2}\,\RKHSkernel(\x, \x)^{1/2}\veps  (\GPkernel_{\eta}(\x, \x)^{1/2} + \GPkernel_{\nu}(\x, \x)^{1/2})  + (1 + \sqrt{2}) \RKHSkernel(\x, \x)\veps^{2}\\
|\GPkernel_{\eta}(\x, \x)^{1/2} - \GPkernel_{\nu}(\x, \x)^{1/2}| 
&\le \sqrt{2}\,\RKHSkernel(\x, \x)^{1/2}\veps  + \frac{(1 + \sqrt{2}) \RKHSkernel(\x, \x)\veps^{2}}{\GPkernel_{\eta}(\x, \x)^{1/2} + \GPkernel_{\nu}(\x, \x)^{1/2}}.
\)
Let $a \defined \frac{1 + \sqrt{3 + 2 \sqrt{2}}}{\sqrt{2}}\RKHSkernel(\x, \x)^{1/2}\veps$. 
If $\max(\GPkernel_{\eta}(\x, \x)^{1/2}, \GPkernel_{\nu}(\x, \x)^{1/2}) \le a$, 
then clearly $|\GPkernel_{\eta}(\x, \x)^{1/2} - \GPkernel_{\nu}(\x, \x)^{1/2}| \le  a$. 
Otherwise we have
\(
|\GPkernel_{\eta}(\x, \x)^{1/2} - \GPkernel_{\nu}(\x, \x)^{1/2}| 
&\le \sqrt{2}\,\RKHSkernel(\x, \x)^{1/2}\veps  + \frac{(1 + \sqrt{2}) \RKHSkernel(\x, \x)\veps^{2}}{a}
= a.
\)
Hence we conclude unconditionally that 
\(
|\GPkernel_{\eta}(\x, \x)^{1/2} - \GPkernel_{\nu}(\x, \x)^{1/2}| 
&\le  \frac{1 + \sqrt{3 + 2 \sqrt{2}}}{\sqrt{2}}\RKHSkernel(\x, \x)^{1/2}\veps 
< \sqrt{6}\RKHSkernel(\x, \x)^{1/2}\veps.
\) 

Thus, we also have that 
\(
|\GPkernel_{\eta}(\x, \x) - \GPkernel_{\nu}(\x, \x)|
&\le \sqrt{2}\,\RKHSkernel(\x, \x)^{1/2}\veps  (\GPkernel_{\eta}(\x, \x)^{1/2} + \GPkernel_{\nu}(\x, \x)^{1/2})  + (1 + \sqrt{2}) \RKHSkernel(\x, \x)\veps^{2} \\
&< \sqrt{2}\,\RKHSkernel(\x, \x)^{1/2}\veps  (2\GPkernel_{\eta}(\x, \x)^{1/2} + \sqrt{6}\RKHSkernel(\x, \x)^{1/2}\veps)  + (1 + \sqrt{2}) \RKHSkernel(\x, \x)\veps^{2} \\
&= 2\sqrt{2}\,\RKHSkernel(\x, \x)^{1/2}\GPkernel_{\eta}(\x, \x)^{1/2}\veps  + (1 + \sqrt{2} + \sqrt{12}) \RKHSkernel(\x, \x)\veps^{2} \\
&< 3\,\RKHSkernel(\x, \x)^{1/2}\GPkernel_{\eta}(\x, \x)^{1/2}\veps  + 6\,\RKHSkernel(\x, \x)\veps^{2}.
\)
The final inequality follows from Jensen's inequality (which implies that the 1-Wasserstein distance lower bound the 2-Wasserstein distance) and \citep[Rmk.~6.5]{Villani:2009}. 

\section{Proof of \cref{prop:use-k-as-r}}
\label{sec:proof-of-use-k-as-r}
We first write $\GPkernel$ in terms of the orthonormal basis of $\rkhsarg{\GPkernel}$:
\(
\GPkernel(\x, \x') = \textstyle{\sum_{j \ge 1}} \basis{j}(\x)\basis{j}(\x').
\)
Define 
\(
\RKHSkernel(\x, \x') \defined \textstyle{\sum_{j \ge 1}} \keval{j}\basis{j}(\x)\basis{j}(\x').
\)
If $\sum_{j \ge 1}\keval{j}^{-1} < \infty$ then $\RKHSkernel$ dominates $\GPkernel$. So 
given inputs $\xall = (\x_n)_{n=1}^N$, and defining $a_{nm,j} \defined \basis{j}(\x_n)\basis{j}(\x_m)$,
to show the existence of the required kernel $\RKHSkernel$ we need to show there exists a solution to 
\(
\forall (n,m) \in [N]^2, \quad \left|\sum_{j\geq1}\keval{j} a_{nm,j} - \sum_{j\geq1}a_{nm,j}\right| \leq \epsilon,\quad \sum_{j\geq1}\keval{j}^{-1} < \infty, \quad\text{and}\quad \forall j\in\nats, \lambda_j\geq 0.
\)
By assumption on the pointwise decay of orthonormal basis elements, for all $(n,m)\in[N]^2$, $\left|a_{nm,j}\right| = o(j^{-2})$. Define $a_j \defined \max_{(n,m)\in[N]^2} |a_{nm,j}|$. Therefore 
$\sqrt{a_j} = o(j^{-1})$, $\sum_{j\geq 1} \sqrt{a_j} < \infty$, and there exists a $J>0$ such that
\(
\forall j > J, \sqrt{a_j} < 1 \quad\text{and}\quad\sum_{j\geq J} \sqrt{a_j} < \epsilon. 
\)
Setting $\lambda_j = 1$ for each $j\in1, \dots, J$ and $\lambda_j = 1+\sqrt{a_j}^{-1}$ for $j > J$, we have that
for any $(n,m)\in[N]^2$,
\(
\left|\sum_{j\geq1}\keval{j} a_{nm,j} - \sum_{j\geq1}a_{nm,j}\right|  = \left|\sum_{j\geq J}\frac{a_{nm,j}}{\sqrt{a_j}}\right| \leq \sum_{j\geq J}\sqrt{a_j} < \epsilon.
\)
Finally since $\sqrt{a_j} = o(j^{-1})$, $\lambda_j = \omega(j)$, and so $\lambda_j^{-1} = o(j^{-1})$ yielding $\sum_{j\geq 1} \lambda_j^{-1} < \infty$.

\section{Proof of \cref{prop:pf-for-DTC}}
\label{app:proof-of-pf-for-DTC}

Let $\loglik{n}(f) \defined -\frac{1}{2\sigma^{2}}(f(\xn{n}) - \yn{n})^{2}$ denote the log-likelihood of the $n$th observation 
and recall that $\rkhs = \rkhsarg{\RKHSkernel}$. 

\bnlem \label{lem:loglik-term-derivative}
For any $f \in \rkhs$,
\(
\Fderiv\loglik{n}(f)
&= - \sigma^{-2}(f(\xn{n}) - \yn{n})\RKHSkernel_{\xn{n}}.
\)
\enlem
\bprf
For $g \in \rkhs$, 
\(
\lefteqn{|\loglik{n}(f + g) - \loglik{n}(f) + \inner{\sigma^{-2}(f(\xn{n}) - \yn{n})\RKHSkernel(\xn{n}, \cdot)}{g}|} \\
&= \left|-\frac{1}{2\sigma^{2}}(f(\xn{n}) + g(\xn{n}) - \yn{n})^{2} + \frac{1}{2\sigma^{2}}(f(\xn{n}) - \yn{n})^{2} + \sigma^{-2}(f(\xn{n}) - \yn{n})g(\xn{n})\right| \\
&\le \frac{1}{2\sigma^{2}}g(\xn{n})^{2}
= \frac{1}{2\sigma^{2}} \inner{\RKHSkernel(\xn{n}, \cdot)}{g}^{2}
\le \frac{\RKHSkernel(\xn{n},\xn{n})}{2\sigma^{2}} \norm{g}^{2}.
\)
\eprf

\bnlem \label{lem:loglik-derivatives}
For any $f \in \rkhs$,
\(
\Fderiv\loglik{}(f) = - \sigma^{-2}(f(\xall) - \yall)^{\top}\RKHSkernel_{\xall}
\)
and
\(
\Fderiv\indptloglik{}(f) 
= -\sigma^{-2}(\bar Q_{\xall\indpts}f(\indpts) - \yall)^{\top} \bar Q_{\xall\indpts} \RKHSkernel_{\indpts}.
\)
\enlem
\bprf
Both results follow directly from \cref{lem:loglik-term-derivative}.
\eprf
\bnlem \label{lem:loglik-fisher-inner-prod}
If $\nu = \distGP(\auxmean, \auxkernel)$, then 
\(
\EE_{f \dist \auxdist}[\inner{\Fderiv\loglik{n}(f)}{\Fderiv\loglik{m}(f)}]
&= \sigma^{-4} \RKHSkernel(\xn{n}, \xn{m})[\auxkernel(\xn{n}, \xn{m}) + (\yn{n} - \auxmean(\xn{n}))(\yn{m} - \auxmean(\xn{m}))]. %\label{eq:fisher-inner-prod}
\)
\enlem
\bprf
Using \cref{lem:loglik-term-derivative}, we have
\(
\EE_{f \dist \auxdist}[\inner{\Fderiv\loglik{n}(f)}{\Fderiv\loglik{m}(f)}]
&= \sigma^{-4} \inner{\RKHSkernel_{\xn{n}}}{\RKHSkernel_{\xn{m}}} \EE_{f \dist \auxdist}[(f(\xn{n}) - \yn{n})(f(\xn{m}) - \yn{m})] \\
&= \sigma^{-4} \RKHSkernel(\xn{n}, \xn{m})[\auxkernel(\xn{n}, \xn{m}) + (\yn{n} - \auxmean(\xn{n}))(\yn{m} - \auxmean(\xn{m}))].
\)
\eprf
\bnlem \label{lem:GP-covariance-operator} 
If $\eta = \distGP(0, \ell)$ then $(\kernelop{\eta}f)(\x) = \inner{f}{\ell_{\x}}$.
\enlem
\bprf
Since $(\kernelop{\eta}\RKHSkernel_{\x'}) = \inner{\RKHSkernel_{\x'}}{\ell_{\cdot}} = \ell_{\x'}$, for $f \dist \eta$,
\(
\inner{\RKHSkernel_{\x}}{\kernelop{\eta}\RKHSkernel_{\x'}} 
= \inner{\RKHSkernel_{\x}}{\ell_{\x'}} 
= \ell(\x, \x') 
= \cov(f(\x), f(\x')).
\)
\eprf
\bnlem \label{lem:DTC-covariance-operator}
For the DTC log-likelihood approximation $\approxdist$, 
\(
(\kernelop{\approxdist}f)(\x) 
= (\kernelop{\priordist}f)(\x) 
  - \inner{f}{\GPkernel_{\indpts}}(\GPkernel_{\indpts\indpts}^{-1} - \tSigma)\GPkernel_{\indpts \x},
\)
where $\tSigma \defined (\GPkernel_{\indpts\indpts} + \sigma^{-2}\GPkernel_{\indpts\xall}\GPkernel_{\xall\indpts})^{-1}$.
\enlem 
\bprf
Since $\approxdist$ has covariance function
$\GPkernel(\x, \x') - Q_{\x\x'} +  \GPkernel_{\x\indpts}\tSigma\,\GPkernel_{\indpts \x}$~\citep{QuinoneroCandela:2005},
the result follows from \cref{lem:GP-covariance-operator}.
\eprf

It follows from \cref{lem:loglik-derivatives,lem:DTC-covariance-operator} that 
\(
\kernelop{\approxdist}\Fderiv\indptloglik{}(f) 
&= -\sigma^{-2}(\bar Q_{\xall\indpts}f(\indpts) - \yall)^{\top} K_{\xall\indpts} \tSigma \GPkernel_{\indpts} \\
\kernelop{\approxdist}\Fderiv\loglik{}(f)
&= - \sigma^{-2}(f(\xall) - \yall)^{\top}(\GPkernel_{\xall} -  \bar Q_{\xall\indpts}\GPkernel_{\indpts} + K_{\xall\indpts}\tSigma\GPkernel_{\indpts}).      %\label{eq:precond-deriv}
\)
We therefore have that 
\(
\lefteqn{-\sigma^{2}\kernelop{\approxdist}\Fderiv(\loglik{} - \indptloglik{})(f) } \\
&= (f(\xall) - \yall)^{\top}(\GPkernel_{\xall} -  \bar Q_{\xall\indpts}\GPkernel_{\indpts}) + (f(\xall) - \bar Q_{\xall\indpts}f(\indpts))^{\top} K_{\xall\indpts}\tSigma\GPkernel_{\indpts}
\)

Consider the limit $\RKHSkernel \to \GPkernel$, so $\sqGPkernel \to \GPkernel$. 
Then 
\(
\lefteqn{\sigma^{4}\staticnorm{\kernelop{\approxdist}\Fderiv(\loglik{} - \indptloglik{})(f)}^{2}} \\
\begin{split}
&= (f(\xall) - \yall)^{\top}(K_{\xall\xall} + \bar Q_{\xall\indpts}K_{\indpts\indpts}\bar Q_{\xall\indpts}^{\top} - 2K_{\xall\indpts}\bar Q_{\xall\indpts}^{\top}) (f(\xall) - \yall) \\
&\phantom{=} + (f(\xall) - \yall)^{\top}(K_{\xall\indpts}\tSigma K_{\indpts\xall} - \bar Q_{\xall\indpts}K_{\indpts\indpts} \tSigma K_{\indpts\xall})(f(\xall) - \bar Q_{\xall\indpts}f(\indpts)) \\
&\phantom{=} + (f(\xall) - \bar Q_{\xall\indpts}f(\indpts))^{\top} K_{\xall\indpts}\tSigma K_{\indpts\indpts} \tSigma K_{\indpts\xall}(f(\xall) - \bar Q_{\xall\indpts}f(\indpts)) 
\end{split} \\
\begin{split}
&= (f(\xall) - \yall)^{\top}(K_{\xall\xall} - Q_{\xall\xall}) (f(\xall) - \yall)  \\
&\phantom{=} +  (f(\xall) - \bar Q_{\xall\indpts}f(\indpts))^{\top} S_{\xall\xall}(f(\xall) - \bar Q_{\xall\indpts}f(\indpts)),
\end{split}
\)
where $S_{\xall\xall} \defined  K_{\xall\indpts}\tSigma K_{\indpts\indpts} \tSigma K_{\indpts\xall}$. 
Let $E_{\xall\xall} \defined K_{\xall\xall} - Q_{\xall\xall}$. 
Taking expectations we get
\(
\lefteqn{\EE_{\auxdist}[ (f(\xall) - \yall)^{\top}E_{\xall\xall}(f(\xall) - \yall)]} \\
&= \EE_{\auxdist}[(f(\xall) - \auxmean(\xall) + \auxmean(\xall) - \yall)^{\top}E_{\xall\xall}(f(\xall) - \auxmean(\xall) + \auxmean(\xall) - \yall)] \\
&= \tr(\hK_{\xall\xall}E_{\xall\xall}) + (\auxmean(\xall) - \yall)^{\top}E_{\xall\xall}(\auxmean(\xall) - \yall)
\)
and 
\(
\lefteqn{\EE_{\auxdist}[(f(\xall) - \bar Q_{\xall\indpts}f(\indpts))^{\top} S_{\xall\xall}(f(\xall) - \bar Q_{\xall\indpts}f(\indpts))]} \\
&= \EE_{\auxdist}[\statictwonorm{(f(\xall) - \auxmean(\xall) +  \bar Q_{\xall\indpts}\auxmean(\indpts) - \bar Q_{\xall\indpts}f(\indpts) + \auxmean(\xall) - \bar Q_{\xall\indpts} \auxmean(\indpts) )^{\top} S_{\xall\xall}^{1/2}}^{2}] \\
\begin{split}
&= \tr(\hK_{\xall\xall}S_{\xall\xall}) + \tr(\hK_{\indpts\indpts}\bar Q_{\xall\indpts}^{\top}S_{\xall\xall}\bar Q_{\xall\indpts}) -  2\tr(\hK_{\indpts\xall}S_{\xall\xall}\bar Q_{\xall\indpts}) \\
&\phantom{=} + (\auxmean(\xall) - \bar Q_{\xall\indpts} \auxmean(\indpts) )^{\top} S_{\xall\xall}(\auxmean(\xall) - \bar Q_{\xall\indpts} \auxmean(\indpts).
\end{split}
\)
Let $S_{\xall\indpts}' \defined  K_{\xall\indpts}\tSigma K_{\indpts\indpts} \tSigma$. 
Putting everything together, conclude that 
\[
\lefteqn{\sigma^{4}\staticnormarg{\auxdist}{\kernelop{\approxdist}\Fderiv(\loglik{} - \indptloglik{})}^{2}} \nonumber \\
\begin{split}
&= \tr((\hK_{\xall\xall} + (\auxmean(\xall) - \yall)(\auxmean(\xall) - \yall)^{\top})(K_{\xall\xall} - Q_{\xall\xall})) \\
&\phantom{=} + \tr(\hK_{\xall\xall}S_{\xall\xall}) + \tr(\hK_{\indpts\indpts}\bar Q_{\xall\indpts}^{\top}S_{\xall\xall}\bar Q_{\xall\indpts}) -  2\tr(\hK_{\indpts\xall}S_{\xall\xall}\bar Q_{\xall\indpts}) \\
&\phantom{=} + (\auxmean(\xall) - \bar Q_{\xall\indpts} \auxmean(\indpts) )^{\top} S_{\xall\xall}(\auxmean(\xall) - \bar Q_{\xall\indpts} \auxmean(\indpts)).
\end{split} \nonumber \\
\begin{split} 
&= -\tr(K_{\indpts\xall}(\hK_{\xall\xall} +  (\auxmean(\xall) - \yall)(\auxmean(\xall) - \yall)^{\top})\bar Q_{\xall\indpts}) \\
&\phantom{=} + \tr((K_{\indpts\xall}\hK_{\xall\xall} + K_{\indpts\xall}\bar Q_{\xall\indpts}\hK_{\indpts\indpts}\bar Q_{\xall\indpts}^{\top} - 2K_{\indpts\xall}\bar Q_{\xall\indpts}\hK_{\indpts\xall})S_{\xall\indpts}') \\
&\phantom{=} + (\auxmean(\xall) - \bar Q_{\xall\indpts} \auxmean(\indpts) )^{\top} S_{\xall\indpts}' K_{\indpts\xall}(\auxmean(\xall) - \bar Q_{\xall\indpts} \auxmean(\indpts)) + C(\xall).
\label{eq:efficiently-computable-pf-for-DTC}
\end{split} 
\]
It is clear from \cref{eq:efficiently-computable-pf-for-DTC} that all quantities can be computed while never instantiating a matrix larger than $N \times M$,
hence, up to the constant $C(\xall)$, the \pfdname can be computed in $O(NM^{2})$ time and $O(NM)$ space.

}

\end{document}